\documentclass[preprint,12pt,nonatbib,nopreprintline]{elsarticle}

\usepackage{amssymb}
\usepackage{amsmath}
\usepackage{graphicx}
\setkeys{Gin}{draft=false}
\usepackage{booktabs}
\usepackage{tabularx}
\usepackage{multirow}
\usepackage{pifont}
\usepackage{placeins}
\usepackage[expansion=false]{microtype}
\usepackage{tikz}
\usepackage{pdflscape}
\newenvironment{widepage}{\begin{landscape}\setlength{\linewidth}{700pt}\setlength{\columnwidth}{700pt}\setlength{\hsize}{700pt}\addtolength{\topmargin}{-83pt}\setlength{\tabcolsep}{3.5pt}}{\end{landscape}}
\newsavebox{\fitbox}\newlength{\fitw}
\let\origresizebox\resizebox
\renewcommand{\resizebox}[3]{\setlength{\fitw}{#1}\sbox{\fitbox}{#3}\ifdim\wd\fitbox>\fitw\typeout{FITSCALE table \thetable: \the\numexpr 100*\numexpr\fitw/100\relax/\numexpr\wd\fitbox/100\relax\relax\space percent}\origresizebox{\fitw}{#2}{\usebox{\fitbox}}\else\usebox{\fitbox}\fi}
\makeatletter\let\c@author\@undefined\let\theauthor\@undefined\makeatother
\usepackage[american]{babel}
\usepackage{csquotes}
\usepackage[style=apa,backend=biber,natbib=true,doi=true,url=true]{biblatex}
\DeclareLanguageMapping{american}{american-apa}
\usetikzlibrary{arrows.meta,positioning,fit,backgrounds,calc}

\begin{document}

\begin{frontmatter}

\title{BioSync: Transformer Fusion for a Multimodal Wearable Digital Biomarker}

\author[unr]{Seyed Mahmoud Sajjadi Mohammadabadi\corref{cor1}}
\ead{mahmoud.sajjadi@unr.edu}
\ead[url]{https://orcid.org/0009-0001-9629-9734}

\author[cfl]{Sean Montgomery}
\ead{sean@connectedfuturelabs.com}

\author[unr]{Alireza Tavakkoli}
\ead{tavakkol@unr.edu}

\cortext[cor1]{Corresponding author.}

\affiliation[unr]{organization={Department of Computer Science and Engineering, University of Nevada, Reno},
            addressline={1664 N. Virginia St.},
            city={Reno},
            postcode={89557},
            state={NV},
            country={USA}}

\affiliation[cfl]{organization={Connected Future Labs},
            city={Reno},
            state={NV},
            country={USA}}

\begin{abstract}
BioSync maps cardiac, neural, behavioral, and speech features to a continuous BioSync Index (BSI). The wide-and-deep design combines a linear branch whose hypothesis class contains feature concatenation with a small self-attention branch over modality tokens (one encoder block, embedding width 8, 1,119 parameters). Evaluation covers two synthetic cohorts informed by published literature (360 profiles each), seven further generators, and real wearable or physiological datasets (N=15 to 469). In single axis scenarios BioSync stayed within +0.001 to +0.011 AUC of concatenated logistic regression, and the attention branch added information where outcomes depended on interactions across modalities (+0.034 and +0.043 AUC over logistic regression at N=360 and N=1,440). Soft voting or a tuned nonlinear model was often as good or better. Attention weights remained near uniform and did not track predictive contribution (rank correlation with modality removal effects: $-0.32$ to $+0.32$) or sensor reliability. Variants described in the appendices address this: a quality-gated variant tracked sensor reliability when it varied across subjects (rank correlation 0.78), and a mixture variant had the smallest worst-case regret across 11 untouched scenarios (0.034 against 0.048 to 0.107 for six baseline families). The refined 1,159-parameter network needs 4.6 kB (1.3 kB with INT8 weights), 4 to 250 times less than tuned SVMs and tree ensembles. On real data BioSync matched logistic regression on WESAD. The refined variants exceeded logistic regression on a Parkinson's smartwatch cohort (N=469) and on stress-resilience wrist recordings (N=35), although a tuned MLP was close on PADS; all models were near chance on two small summary-feature cohorts (N=56 to 100). The BSI is a ranking score, not a calibrated probability. Results characterize computational behavior but do not establish biomarker validity.
\end{abstract}

\begin{highlights}
\item Wide-and-deep fusion of wearable heart, brain, activity and speech features
\item Attention recovers cross-modal interactions that linear fusion misses
\item Attention weights do not track contribution; a quality gate tracks sensor reliability
\item Smallest worst-case regret across 11 regimes; 1.3 kB footprint with INT8 weights
\item Five real cohorts (N=15--469): refined variants beat LR on PADS, HCI; WESAD parity
\end{highlights}


\end{frontmatter}

\section{Introduction}
\label{sec:intro}

Dementia affects more than 55 million people worldwide, with Alzheimer's disease (AD) accounting for most cases. Prevalence is expected to rise sharply as populations age \citep{livingston2020,alzfacts2023}. Clinical diagnosis relies on episodic neuropsychological examinations combined with cerebrospinal fluid assays, amyloid positron emission tomography, and structural magnetic resonance imaging. Although sensitive, these clinical procedures are invasive, costly, and restricted to clinics and laboratories. These constraints limit continuous or frequent longitudinal monitoring of mild cognitive impairment (MCI) during daily living \citep{jack2018,califf2018}. Consumer wearables, smartphones, and portable electroencephalography (EEG) headbands offer an alternative and can capture objective physiological, behavioral, and vocal parameters continuously outside hospital settings \citep{vasudevan2022,kourtis2019,piau2019,dunn2021}. Large scale consumer sensor trials have already demonstrated that wearable photoplethysmography (PPG) and smartwatches can detect paroxysmal cardiac abnormalities \citep{perez2019} and early infectious or systemic inflammatory onset \citep{quer2021}. However, clinical and analytical validation of digital health technologies for neurodegenerative and metabolic conditions remains nascent \citep{coravos2019,babrak2020}. In a systematic scoping review of 431 digital biomarker studies, the mean AUC was 0.887 for AD models using AI and 0.821 for MCI models; fewer than 3\% performed external prospective validation, and only 24 examined multimodal biomarkers \citep{qi2025}.

Individual wearable sensors observe separate aspects of systemic pathophysiology. Heart rate variability (HRV) derived from interbeat intervals quantifies autonomic nervous system equilibrium. Blunted parasympathetic withdrawal and vagal tone have been linked to early central autonomic network dysregulation in AD, although observed effect sizes and directions fluctuate across cohorts \citep{bateman2025,marcolini2026,liu2014,shaffer2017hrv}. Resting quantitative EEG (qEEG) provides neurophysiological markers of cortical slowing, notably elevated theta-to-alpha power ratios and collapsed spectral and Lempel-Ziv complexity, which reflect synaptic disconnection in amnestic MCI and can be acquired through wireless dry electrode headbands \citep{boudaya2024,katayama2023,klimesch1999eeg,nuwer1999qeeg}. Continuous wrist accelerometry captures rest and activity fragmentation, intradaily variability, and suppressed circadian amplitude, which correlate with biological senescence and neurocognitive vulnerability in epidemiological cohorts exceeding 80,000 individuals \citep{ijbnpa2025,cosinorage2024}. Brief spontaneous speech captures vocal acoustic degradation, reduced prosodic pitch variability, and semantic density loss. Pooled screening sensitivity and specificity approach 80\% for MCI and pre-frailty \citep{teh2022mci,cay2024speech,xu2025language}. Combining cardiac autonomic, neural, behavioral, and acoustic channels may recover diagnostic signals that no single modality provides \citep{li2023synergy,radu2018multimodal}.

Continuous ambulatory monitoring must meet hardware constraints that standard multimodal benchmarks often do not address. In real world environments, loose wristbands, arm movements, motion artifacts, dry electrode impedance shifts, battery drain, and nonwear events can corrupt or cause the loss of individual channels \citep{goldsack2020v3,bent2020,elgendi2012ppg}. A robust fusion framework must process asynchronous inputs that may be partially missing or corrupted by noise without large performance drops. Computational complexity must fit the resource budgets of wearable microcontrollers and edge gateways, because available static RAM (SRAM) is often constrained to tens of kilobytes and may not support temporal neural networks with millions of parameters \citep{warden2019tinyml,ray2021tinyml,lin2020mcunet,banbury2021micronets}. Patient privacy and commercial telemetry restrictions present another constraint: institutional clinics and device manufacturers often prohibit the centralized pooling of raw, high frequency biopotential waveforms \citep{can2021,ali2022fedsurvey}. An architecture for on device use must fuse compact summary features, quantify modality contributions for each patient and session, and remain computationally tractable at the edge.

Current multimodal digital biomarker pipelines use two common static approaches: early fusion, which concatenates flattened feature vectors before feeding a single classifier, and late fusion, which averages or votes over independently trained model outputs for each modality \citep{qi2025,li2023synergy,xu2025language}. Neither approach weights modalities separately for each session. While tensor fusion \citep{zadeh2017tensor} and attention across modalities \citep{tsai2019mult,jaegle2021perceiver,nagrani2021bottlenecks} have demonstrated reported gains over static concatenation in emotional computing and cardiovascular diagnosis \citep{crossattn2025,pacfnet2025,ctaf2026}, their application to wearable neurocognitive and metabolic screening under resource constraints remains underexplored \citep{boudaya2024}, particularly with limited on device memory and structured sensor failures.

BioSync addresses these challenges by projecting each sensing modality into a discrete token embedding and computing self-attention with multiple heads over the joint sequence. The resulting continuous score, termed the \emph{BioSync Index (BSI)}, is formalized as a composite digital susceptibility and monitoring biomarker under the FDA-NIH BEST framework \citep{best2016,califf2018,babrak2020}. BioSync includes a parallel linear bypass whose hypothesis space contains feature concatenation as a strict subset (Section~\ref{sec:widedeep}). Section~\ref{sec:theory} links the attention path to measurement theory for latent variables and multivariate information synergy. Figure~\ref{fig:overview} summarizes the workflow, from multimodal measurements through the BioSync fusion model to the BSI and the evaluations reported here.

\begin{figure}[t]
\centering
\resizebox{0.95\textwidth}{!}{%
\begin{tikzpicture}[
    node distance=0.8cm,
    box/.style={draw, rounded corners, minimum height=2.2cm, align=center, inner sep=6pt, font=\small},
    input/.style={box, fill=blue!7, text width=3.2cm},
    fusion/.style={box, fill=purple!8, text width=4.0cm},
    output/.style={box, fill=green!10, text width=3.2cm},
    test/.style={box, fill=orange!9, text width=3.6cm},
    arr/.style={-{Latex[length=2.5mm]}, thick}
]

\node[input] (inputs) {
    \textbf{Multimodal measurements}\\[2pt]
    Cognitive: HRV, EEG,\\ actigraphy, speech\\[2pt]
    Metabolic: cardiac-autonomic,\\ behavioral
};

\node[fusion, right=of inputs] (biosync) {
    \textbf{BioSync fusion}\\[2pt]
    Modality tokens and\\ multi-head self-attention\\
    $+$\\
    Linear concatenation path
};

\node[output, right=of biosync] (bsi) {
    \textbf{Model outputs}\\[2pt]
    Continuous BioSync Index\\ (BSI), a ranking score\\[2pt]
    Per-subject attention allocation
};

\node[test, right=of bsi] (evaluation) {
    \textbf{Evaluation}\\[2pt]
    Accuracy, F1-score, AUC\\
    Severity association\\
    Corruption robustness\\
    Ablations, branch contributions\\
    Calibration; real recordings
};

\draw[arr] (inputs) -- (biosync);
\draw[arr] (biosync) -- (bsi);
\draw[arr] (bsi) -- (evaluation);

\end{tikzpicture}%
}
\caption{High-level study overview. BioSync combines modality-specific inputs through an attention path and a parallel linear path, produces a continuous BSI score and a per-subject attention allocation (not a contribution or reliability measure), and is evaluated on discrimination, severity association, corruption robustness, ablations and branch contributions, calibration, and real recordings.}
\label{fig:overview}
\end{figure}

The study makes six contributions. It formalizes the BSI as a continuous biomarker for each subject, grounded in the FDA-NIH BEST and Digital Medicine Society (DiMe) V3 verification frameworks. It also adapts the established wide-and-deep design \citep{cheng2016widedeep} to modality tokens with a standard self-attention block, motivated by precision estimation for latent variables and multivariate information synergy. We do not claim a new learning algorithm. Instead, we provide a controlled characterization of when this combination differs from linear concatenation and when it does not. The architecture is evaluated against baselines for each modality, static concatenation, and pure attention across two synthetic cohorts informed by published literature ($N=360$ each). The evaluation includes hyperparameter sweeps and modality dropout perturbations. A second JAX implementation, written separately, evaluates high feature noise, Markovian artifact bursts, and a synthetic cohort whose outcome depends only on interactions (Section~\ref{sec:extra}). Additional analyses compare seven further synthetic generators, quantify the contribution of each branch and the faithfulness of the attention weights, report calibration and stability of the BSI, and evaluate the model and tuned baselines on three real wearable or physiological datasets with reference outcomes (Section~\ref{sec:review}). The study also reports the memory and computational profile for on device use (847--1,119 parameters, 3.4--4.5~kB SRAM, Section~\ref{sec:deployment}).

The cognitive decline cohort evaluates a configuration with four modality tokens ($M=4$: HRV, dry EEG, wrist actigraphy, speech acoustics). The metabolic autonomic cohort reproduces the public wearable activity monitor schema of the NIH Bridge2AI AI-READI program ($M=2$: Garmin Vivosmart~5 cardiac autonomic and behavioral tokens) \citep{aireadi2024}, guided by clinical diabetic autonomic neuropathy literature \citep{vinik2003diabetic,spallone2011can}. Because both cohorts are generated from models parameterized from published literature without human participant records, absolute classification metrics quantify architectural mechanics rather than prospective clinical efficacy.

\section{Related Work}
\label{sec:related}

\subsection{Digital biomarkers for Alzheimer's disease and MCI}
The joint FDA-NIH BEST (Biomarkers, EndpointS, and other Tools) resource formalizes seven distinct biomarker classes: susceptibility/risk, diagnostic, monitoring, prognostic, predictive, pharmacodynamic/response, and safety \citep{best2016,califf2018}. Capturing physiological or behavioral measures through digital health technologies transforms them into \emph{digital biomarkers} \citep{vasudevan2022,babrak2020}. To implement translational evaluation, the Digital Medicine Society established the modular V3 framework: technical verification (verifying sensor hardware precision), analytical validation (confirming algorithm measurement consistency against gold standards), and clinical validation (demonstrating that the biomarker identifies or predicts a defined clinical endpoint in the target population) \citep{goldsack2020v3,manta2020,coravos2019}. BioSync is positioned at the intersection of technical verification and analytical exploration, and is intended to support subsequent clinical trials (Section~\ref{sec:limitations}).

Digital health technologies for cognitive assessment span diverse sensor modalities. \citet{kourtis2019} reviewed mobile and wearable sensing modalities for AD, noting that motor gait and acoustic features provide scalable adjuncts to traditional cognitive batteries. \citet{piau2019} surveyed unobtrusive home based continuous technologies, emphasizing that ecologically valid longitudinal metrics capture subtle functional declines that snapshot clinic assessments fail to detect. \citet{qi2025} systematically reviewed 431 digital biomarker studies and 86 AI models and found that 94.4\% investigated single modality pipelines; only 24 studies attempted multimodal fusion, and reports of calibration or external cohort validation were scarce. In an acoustic study, \citet{xu2025language} reported 91.8\% screening accuracy using hard voting late fusion across linguistic markers. Hard voting assigns uniform modality contributions across all participants regardless of signal quality.

\subsection{Cardiac autonomic, neural, and behavioral digital signals}
Autonomic nervous system function is indexed noninvasively through time and frequency domain HRV metrics \citep{shaffer2017hrv}. The central autonomic network links prefrontal and limbic cortices to autonomic brainstem nuclei, and its degradation occurs early in Alzheimer's pathogenesis \citep{liu2014,bateman2025}. However, reported associations remain conflicting: \citet{bateman2025} observed distinct task induced HRV shifts during cognitive testing, whereas \citet{marcolini2026} identified minimal independent predictive power for resting HRV in predicting brain amyloidosis. These conflicting findings support using HRV as a complementary component in a multisensor ensemble rather than as an isolated screening metric.

Resting quantitative EEG (qEEG) provides high temporal resolution correlates of cortical integrity \citep{nuwer1999qeeg}. Cognitive slowing in amnestic MCI typically manifests as power redistribution from higher frequencies to slower bands, producing elevated theta-to-alpha ratios and diminished spectral and Lempel-Ziv complexity \citep{klimesch1999eeg,katayama2023}. \citet{boudaya2024} established that joint qEEG and HRV feature sets significantly outperform either modality alone for automated MCI classification, confirming the diagnostic value of cross system neural and cardiac coupling.

Wearable wrist accelerometers quantify continuous circadian rest and activity patterns. Alterations in sleep efficiency, intradaily variability, and circadian rhythmicity have been validated as sensitive markers of biological aging and cognitive deterioration across large biobanks ($>80,000$ participants) \citep{cosinorage2024,ijbnpa2025}. In metabolic conditions such as type 2 diabetes (T2D), cardiac autonomic neuropathy (CAN) causes progressive vagal denervation, leading to resting tachycardia, diminished HRV, and impaired circadian blood pressure and activity modulation \citep{vinik2003diabetic,spallone2011can,sletten2012compass}. These physiological phenomena form the clinical basis for the metabolic autonomic token grouping examined in Section~\ref{sec:features}.

\subsection{Multimodal fusion architectures}
Multimodal machine learning architectures are conventionally grouped into early, intermediate, and late fusion \citep{baltrusaitis2019multimodal,qi2025}. Early fusion directly concatenates input features before model inference, permitting linear models to capture linear combinations across all inputs but failing to model complex interactions across modalities without explicit polynomial feature expansion. Late fusion trains independent models for each sensor and combines predictions through hard voting or soft score averaging, precluding any feature synergy at the representation level \citep{li2023synergy,xu2025language}. Intermediate fusion models representations jointly: \citet{zadeh2017tensor} introduced Tensor Fusion Networks to capture unimodal, bimodal, and trimodal interactions via outer products, though at significant parameter expansion.

Transformers with attention across modalities have surpassed static fusion across diverse biomedical domains. \citet{tsai2019mult} proposed Multimodal Transformers (MulT) with directional attention across modalities for unaligned sequential streams. \citet{jaegle2021perceiver} introduced the Perceiver architecture, using latent query arrays to map heterogeneous inputs to shared latent bottlenecks with few dimensions, while \citet{nagrani2021bottlenecks} showed that compact attention bottlenecks restrict interactions across modalities to designated tokens, reducing quadratic self-attention costs. In wearable and physiological sensing, cross-attention networks have reported higher accuracy in emotion recognition from EEG and peripheral sensors \citep{crossattn2025}, dual ECG-phonocardiogram cardiovascular screening \citep{pacfnet2025}, and self-supervised cross-temporal alignment \citep{ctaf2026}. Similarly, kernel discriminant correlation analysis has been applied in autism screening \citep{wadhera2024}, and explainable electrophysiological attention has been used for sleep staging \citep{ellis2021} and coma prognosis \citep{amiri2023}. However, existing attention architectures typically contain millions of parameters and ignore the microwatt power and less than 10~kB SRAM constraints of wearable microcontrollers. BioSync applies a lightweight Wide-and-Deep token transformer to edge physiological computing.

\subsection{Synthetic data for multimodal physiological modeling}
\label{sec:synthrelated}
Synthetic physiological data are produced in three broad ways. Mechanistic generators simulate a waveform from a physiological model, for example the dynamical electrocardiogram model of \citet{mcsharry2003ecgsyn}, and signal simulation toolboxes such as NeuroKit2 \citep{makowski2021neurokit} synthesize ECG, electrodermal, and other signals. Generators at the population level, such as Synthea \citep{walonoski2017synthea} simulate patient trajectories and health records from modules for disease progression. Statistical generators draw summary features from a specified joint distribution. A simulation study should state its data generating mechanism, the estimand, and the performance measures in advance \citep{morris2019simulation}. Synthetic health data are only as informative as their fidelity to real data, and their reported utility should be checked against real data where possible \citep{chen2021synthetic,gonzales2023synthetic}. Recovery of a latent factor structure depends on sample size and on how strongly the indicators load on the factors \citep{maccallum1999sample}. The generator used in this paper belongs to the third group: it is an independently designed simulation in which each feature is a function of a latent severity variable plus noise, seeded with effect directions from published literature, and it does not adapt an established generator. Mechanistic generators and generators at the record level address questions about raw signals and records, but they do not give the controlled feature structure across modalities that is needed to test a fusion rule. A summary feature simulator makes its assumptions explicit and lets the structure be varied (one latent factor or several, interactions, reliability that varies across subjects). Its limitation is that realism is assumed rather than validated, so conclusions are conditional on the generator. We therefore list every generator parameter (Section~\ref{app:params}), test the conclusions across several generators (Section~\ref{app:generators}), and add evaluations on real recordings (Section~\ref{app:realdata}).

\section{Theoretical Rationale}
\label{sec:theory}

A fused multimodal signal can contain information that is absent from any single modality. Two complementary theoretical arguments motivate this claim: latent variable precision weighting and multivariate information synergy.

\subsection{A latent variable measurement model}
Consider an unobserved latent physiological state $z \in \mathbb{R}$, such as the degree of central autonomic network disruption associated with cognitive decline or the degree of cardiac autonomic neuropathy associated with a metabolic complication. No wearable sensor measures $z$ directly; each modality $m \in \{1,\ldots,M\}$ instead provides a noisy, partial projection,
\begin{equation}
x_m = g_m(z) + \varepsilon_m, \qquad \varepsilon_m \sim \mathcal{N}(0, \sigma_m^2),
\end{equation}
where $g_m$ is a potentially nonlinear measurement function specific to each modality and $\sigma_m^2$ denotes the modality's observation noise variance. In continuous ambulatory monitoring, observation noise varies systematically across subjects and recording sessions due to physical interference (e.g., dry electrode motion artifacts during movement or loose optical sensor contact) \citep{bent2020}. From the perspectives of classical measurement theory and structural equations, each modality estimates $z$ consistently but with different estimation efficiencies. Assuming the measurement errors $\varepsilon_m$ are conditionally independent given $z$, maximum likelihood estimation combines unbiased estimates at the modality level $\hat{z}_m$ through a linear combination weighted by precision, or equivalently by inverse variance,
\begin{equation}
\hat{z} = \frac{\sum_{m=1}^M \sigma_m^{-2}\, \hat{z}_m}{\sum_{m=1}^M \sigma_m^{-2}}, \qquad \mathrm{Var}(\hat{z}) = \left(\sum_{m=1}^M \sigma_m^{-2}\right)^{-1},
\end{equation}
which achieves lower variance than any single unimodal estimator $\hat{z}_m$. Kalman filter state estimation and meta-analysis weighted by precision share this principle. In BioSync (Section~\ref{sec:transformer}), the attention weights $\boldsymbol{\alpha}$, normalized by softmax, act as an analogue of precision weights that vary with the input, $\sigma_m^{-2} / \sum_j \sigma_j^{-2}$: the network assigns greater representational weight to modalities with higher signal clarity for each input instance. Although learned attention weights should not be conflated with calibrated statistical precision (Section~\ref{sec:extra}), this measurement framework provides a principled motivation for weighting that depends on the input.

\subsection{Information theoretic complementarity and synergy}
While the measurement model justifies multimodal integration when sensors provide independent noisy measurements of a shared scalar cause, multivariate information theory addresses a distinct phenomenon: information that exists only in joint configurations of modalities.

Formally, Partial Information Decomposition (PID) decomposes the total mutual information $I(Y; X_1, \ldots, X_M)$ that a set of modalities jointly provides regarding an outcome $Y$ \citep{williams2010pid,lizier2018information}. For a bivariate modality pair $\{X_1, X_2\}$, the total mutual information has a unique partition into four nonnegative components:
\begin{equation}
I(Y; X_1, X_2) = \mathrm{Red}(Y; X_1, X_2) + \mathrm{Uniq}(Y; X_1) + \mathrm{Uniq}(Y; X_2) + \mathrm{Syn}(Y; X_1, X_2),
\end{equation}
where $\mathrm{Red}(Y; X_1, X_2)$ represents redundant information accessible through either modality independently. The terms $\mathrm{Uniq}(Y; X_1)$ and $\mathrm{Uniq}(Y; X_2)$ denote information unique to each respective channel. $\mathrm{Syn}(Y; X_1, X_2)$ denotes \emph{synergistic information}. Synergistic information is absent from the marginal distributions of $X_1$ or $X_2$ in isolation, and occurs only through their joint statistical interaction. In clinical physiology, an elevated sympathetic HRV ratio (LF/HF) may indicate benign physical exertion when observed in isolation; however, when observed \emph{conditional on} a concomitant collapse in resting EEG spectral complexity and sleep fragmentation, it signals neuro autonomic decoupling driven by central neurodegeneration \citep{liu2014,boudaya2024}.

Standard fusion strategies struggle to capture this interaction. Late voting combines marginal predictions $\hat{y}_m$ after nonlinearities but discards feature level dependencies across modalities. Linear concatenation models additive effects only and cannot represent products across modalities without explicit polynomial term expansions \citep{qi2025,li2023synergy}.

The synthetic generators of Section~\ref{app:generators} represent these quantities. Scenarios S0--S5 are additive redundant or unique information regimes. S6 varies the observation noise of each sensor between subjects (the situation of the measurement model above). S7 makes the class the sign of the product of an HRV latent and an EEG latent, so that the information is purely synergistic (the $\mathrm{Syn}$ term of the decomposition).

\subsection{Why a Transformer specifically}
\label{sec:whytransformer}
Capturing both precision reweighting and nonadditive synergistic interactions requires a fusion operator that models pairwise modality interactions for each input without exponential parameter scaling. Tensor Fusion Networks \citep{zadeh2017tensor} compute the full outer product over modality feature spaces, but their parameter count scales as $\mathcal{O}(d^M)$, which becomes prohibitive for wearable devices with limited resources as modalities grow. Multimodal Transformers with pairwise directional attention \citep{tsai2019mult} require $\mathcal{O}(M^2)$ distinct cross-attention modules, which increases memory requirements.

Self-attention over modality tokens avoids this scaling cost \citep{vaswani2017}. By projecting each modality into a shared embedding space $\mathbb{R}^d$ and computing self-attention over the sequence $[e_{\text{cls}}, e_1, \ldots, e_M]$, pairwise token interactions are evaluated with quadratic complexity in token count $M$ but linear parameter scaling in the shared dimension $d$ \citep{jaegle2021perceiver,nagrani2021bottlenecks}. Softmax normalization bounds attention weights for each instance. Treating modalities as sequence tokens also decouples the encoder backbone from the input feature dimensions, enabling identical transformer weights to process different sensor counts (e.g., $M=4$ for cognitive decline and $M=2$ for metabolic risk) without altering internal projection matrices.

\subsection{Design criteria for composite biomarkers}
\label{sec:desiderata}
Grounded in the FDA-NIH BEST taxonomy \citep{best2016,califf2018}, the V3 validation framework \citep{goldsack2020v3,manta2020}, and edge computing principles \citep{warden2019tinyml,lin2020mcunet}, we formulate six explicit design criteria for pervasive composite digital biomarkers:

\begin{enumerate}
\item \textbf{Graded validity}: The composite score should vary continuously with the severity of underlying pathophysiology rather than collapsing to a coarse binary classification. This supports longitudinal trajectory tracking and monitoring of subtle disease progression \citep{califf2018,babrak2020}.
\item \textbf{Multimodal information capture}: The framework should capture complementary and synergistic information across distinct physiological subsystems, including information not available from isolated marginal channels \citep{williams2010pid,li2023synergy}.
\item \textbf{Attribution for each subject}: The contribution of each sensor to the score should be recoverable for each subject and session. For models based on attention this requires an attribution measured on the complete model (for example, modality removal), because attention weights alone describe attention allocation and not predictive contribution.
\item \textbf{Graceful degradation}: Prediction quality should degrade smoothly under transient noise bursts, sensor displacement, or intermittent channel loss so that artifacts in real world settings do not cause large performance drops \citep{goldsack2020v3,bent2020}.
\item \textbf{Architectural generality}: The core fusion architecture must remain independent of modality and transferable across clinical domains without requiring bespoke architectural redesigns \citep{baltrusaitis2019multimodal}.
\item \textbf{Feasibility on device}: Model memory and compute footprints must comply with sub-5~kB SRAM and very low power MCU execution budgets (e.g., ARM Cortex-M4/M33). This allows inference near the sensor without offloading unencrypted biosignals \citep{warden2019tinyml,ray2021tinyml,banbury2021micronets}.
\end{enumerate}

Table~\ref{tab:desiderata} compares specific implementations on measured criteria derived from these six (Section~\ref{sec:comparison}); we do not assign categorical ratings to whole method families.

\section{Proposed BioSync Framework}
\label{sec:method}

\subsection{Problem formulation}
A participant is represented by feature vectors specific to each modality $\{x_m\}_{m=1}^{M}$, extracted from a fixed length window by signal processing pipelines (Section~\ref{sec:features}). BioSync learns a function
\begin{equation}
f: \{x_m\}_{m=1}^{M} \mapsto (\hat{y}, \boldsymbol{\alpha})
\end{equation}
that outputs a continuous score $\hat{y} \in (0,1)$, the sigmoid output of a classifier trained on binary labels, defined as the \textbf{BioSync Index (BSI)}. The attention weight vector $\boldsymbol{\alpha}$, with $\sum_m \alpha_m = 1$, summarizes each modality's contribution to the fused token for that subject. Under the BEST taxonomy, the BSI is intended as a susceptibility/risk and monitoring biomarker: a composite, continuously valued indicator of physiological status rather than a diagnostic label \citep{best2016,califf2018}. The BSI is a continuous ranking score. Its continuity alone does not make it a calibrated risk probability or a measure of impairment severity, and correlation with a severity variable alone does not define the meaning of its scale. We therefore report its calibration and its stability across training repetitions separately (Section~\ref{app:calibration}) and do not interpret its numerical value as a probability or a severity unless that evidence supports it. We evaluate this formulation with $M=4$ for cognitive decline and $M=2$ for metabolic autonomic risk.

\subsection{Two application domains and their modality specific features}
\label{sec:features}

\textbf{Domain 1: cognitive decline ($M=4$).}

\textbf{HRV.} Interbeat interval series derived from wearable photoplethysmography (PPG) or ECG provide six features \citep{shaffer2017hrv}. Time domain metrics include the standard deviation of normal-to-normal intervals (SDNN), root mean square of successive differences (RMSSD), and percentage of successive intervals differing by $>50$~ms (pNN50). Frequency domain metrics comprise low frequency (LF: 0.04--0.15~Hz) and high frequency (HF: 0.15--0.40~Hz) spectral power, alongside the sympathovagal balance ratio (LF/HF), reflecting central autonomic dysfunction in neurodegeneration \citep{bateman2025,liu2014}.

\textbf{EEG.} Portable dry electrode wireless headbands yield six quantitative EEG features \citep{nuwer1999qeeg}. These comprise the theta-to-alpha power ratio (4--8~Hz / 8--12~Hz), gamma-band spectral entropy (30--45~Hz), frontal theta absolute power, parietal alpha absolute power, beta-band power (13--30~Hz), and Lempel-Ziv algorithmic complexity, tracking synaptic deterioration and cortical slowing in amnestic MCI \citep{katayama2023,klimesch1999eeg,boudaya2024}.

\textbf{Actigraphy.} Continuous wrist accelerometry supplies six behavioral and circadian features \citep{cosinorage2024,ijbnpa2025}. These comprise the index of rest and activity fragmentation, circadian relative amplitude, nighttime sleep efficiency, coefficient of variation of hourly step counts, interdaily stability (IS), and intradaily variability (IV), quantifying circadian disruption and motor volatility.

\textbf{Speech.} Brief spontaneous audio recordings provide four acoustic and linguistic features: pause-to-speech ratio, articulation rate (syllables/second), pitch (fundamental frequency $F_0$) standard deviation, and semantic density \citep{cay2024speech,xu2025language}. A smartphone microphone can record these during routine interactions, remaining independent of physiological hardware worn on the body.

\textbf{Domain 2: metabolic autonomic risk, structured around the AI-READI schema ($M=2$):} this domain uses variables documented for the AI-READI wearable activity monitor data domain \citep{aireadi2024}.

AI-READI is an NIH Bridge2AI project developing a FAIR multimodal dataset. Its public documentation reports more than 1,000 enrolled adults and a target of 4,000. T2D severity ranges from no T2D to disease managed with insulin. Participants are recruited at three U.S. sites. They wear a Garmin Vivosmart~5 during a 10-day period of monitoring at home with 5-second sampling \citep{aireadi2024}.
The documentation lists seven channels: heart rate (bpm), oxygen saturation (SpO$_2$\%, during sleep only), step count, calories from physical activity, respiration rate derived from HRV, sleep duration, and a proprietary 0--100 device stress score computed from heart rate and HRV. Records use an Open mHealth JSON schema; five of these variables form two tokens, with resting heart rate and respiration rate entering the \emph{Cardiac-Autonomic} token together with the device stress score. Step count and sleep duration form the \emph{Behavioral} token.

The diabetic cardiac autonomic neuropathy (CAN) literature motivates this grouping: resting tachycardia emerges early, HRV metrics collapse, and vagal modulation coupled to respiration diminishes with neuropathic denervation induced by chronic hyperglycemia \citep{vinik2003diabetic,spallone2011can,sletten2012compass}. Concurrently, daily physical step counts and restorative sleep duration diminish with diabetic complications and metabolic fatigue. The main text experiments use only AI-READI's public schema documentation. According to the project documentation for dataset version 3.0.0 \citep{aireadidocs}, the Garmin Vivosmart~5 variables are available to licensed users after they accept the AI-READI data license, and only variables such as ZIP code, sex, race/ethnicity, and medications require approval by the Data Access Committee. An evaluation on licensed AI-READI wearable recordings is reported in Section~\ref{app:realdata}.
The synthetic metabolic cohort in Section~\ref{sec:results} reproduces the documented schema through these two tokens and uses effect directions from published literature; the same configuration with two tokens is evaluated on real AI-READI wearable recordings in Section~\ref{app:realdata}.

\paragraph{Why summary features rather than raw signals} BioSync consumes summary features rather than raw windowed signals for four reasons. The literature reports effect directions for these established biomarkers. The inputs therefore remain clinically interpretable, and weights for each subject refer to named modalities rather than latent temporal codes. A wearable or phone can compute the features with modest effort. Encoding raw multichannel PPG and EEG alongside audio with convolutional or recurrent networks would raise inference cost by orders of magnitude and could conflict with on device constraints \citep{warden2019tinyml,lin2020mcunet}. Prospective wearable cohorts often contain only tens to hundreds of participants. Temporal encoders with high capacity can overfit at this scale. Fixed inputs isolate the question: how should modalities be fused? Temporal encoders are the natural extension (Section~\ref{sec:limitations}), especially when dynamics within each channel contain information discarded by summaries.

Multimodal learning does not require modalities to share a sensor type, sampling rate, or physiological system when each carries information about the same outcome \citep{baltrusaitis2019multimodal} (Section~\ref{sec:theory}). Within each cross-validation fold, every feature is standardized to zero mean and unit variance using training partition statistics only.

\subsection{Transformer fusion encoder}
\label{sec:transformer}
Figure~\ref{fig:architecture} traces the attention path from modality specific measurements and features to the fusion token. It then shows the BSI and modality weights. Section~\ref{sec:widedeep} introduces the parallel linear path separately; Figure~\ref{fig:overview} includes it in the high level view.

\begin{figure}[t]
\centering
\resizebox{0.95\textwidth}{!}{%
\begin{tikzpicture}[
    node distance=0.55cm and 0.55cm,
    sensor/.style={draw, rounded corners, fill=blue!6, text width=2.6cm, minimum height=0.9cm, align=center, font=\scriptsize, inner sep=2pt},
    feat/.style={draw, rounded corners, fill=orange!8, text width=2.6cm, minimum height=1.0cm, align=center, font=\scriptsize, inner sep=2pt},
    token/.style={draw, circle, fill=teal!15, minimum size=0.85cm, font=\scriptsize, align=center},
    block/.style={draw, rounded corners, fill=purple!8, text width=10.6cm, minimum height=0.9cm, align=center, font=\scriptsize, inner sep=3pt},
    outbox/.style={draw, rounded corners, fill=green!10, text width=7cm, minimum height=0.9cm, align=center, font=\small, inner sep=3pt},
    arr/.style={-{Latex[length=2mm]}, thick}
]

\node[sensor] (s1) at (0,0)   {PPG / ECG};
\node[sensor] (s2) [right=of s1] {Dry-electrode EEG};
\node[sensor] (s3) [right=of s2] {Wrist accelerometer};
\node[sensor] (s4) [right=of s3] {Microphone (speech)};

\node[feat, below=0.5cm of s1] (f1) {HRV features:\\ SDNN, RMSSD, LF/HF};
\node[feat, below=0.5cm of s2] (f2) {EEG features:\\ theta/alpha, entropy};
\node[feat, below=0.5cm of s3] (f3) {Actigraphy features:\\ fragmentation, sleep};
\node[feat, below=0.5cm of s4] (f4) {Speech features:\\ pause, rate, pitch var.};

\draw[arr] (s1) -- (f1); \draw[arr] (s2) -- (f2);
\draw[arr] (s3) -- (f3); \draw[arr] (s4) -- (f4);

\node[token, below=1.0cm of f1] (t1) {$e_1$};
\node[token, below=1.0cm of f2] (t2) {$e_2$};
\node[token, below=1.0cm of f3] (t3) {$e_3$};
\node[token, below=1.0cm of f4] (t4) {$e_4$};
\node[token, fill=gray!25] (tcls) at ($(t1)+(-2.0,0)$) {$e_{\text{cls}}$};

\draw[arr] (f1) -- node[right,font=\tiny]{$W_1 x_1{+}b_1$} (t1);
\draw[arr] (f2) -- (t2);
\draw[arr] (f3) -- (t3);
\draw[arr] (f4) -- (t4);

\node[block, below=0.9cm of $(t2)!0.5!(t3)$] (mha) {Multi-head self-attention over $[e_{\text{cls}}, e_1, e_2, e_3, e_4]$: \; $\mathrm{softmax}(QK^\top/\sqrt{d_h})V$ \; $\to$ residual $\to$ LayerNorm};
\node[block, below=0.3cm of mha] (ffn) {Position-wise feed-forward (GELU) $\to$ residual $\to$ LayerNorm};

\foreach \t in {tcls,t1,t2,t3,t4}{
  \draw[arr] (\t) -- (mha.north -| \t);
}

\node[outbox, below=0.7cm of ffn] (out) {\textbf{BioSync Index (BSI)}: \; $\hat y=\sigma(w^\top z+b_0)$ \\ + per-modality attention allocation $\boldsymbol\alpha$};
\draw[arr] (ffn) -- (out);

\end{tikzpicture}
}
\caption{BioSync attention path. Each modality is independently sensed, feature-extracted, and projected into a shared token embedding. A learned fusion token ($e_{\text{cls}}$) attends to the modality tokens through multi-head self-attention; the fused representation then yields an attention-path BSI contribution and per-subject attention-allocation weights (not predictive contributions or sensor reliabilities). The full model adds the parallel linear path defined in Section~\ref{sec:widedeep}.}
\label{fig:architecture}
\end{figure}

Each modality's standardized feature vector $x_m \in \mathbb{R}^{d_m}$ is linearly projected into a shared $d$-dimensional embedding, treated as one \emph{token}:
\begin{equation}
e_m = W_m x_m + b_m, \qquad e_m \in \mathbb{R}^{d}, \quad W_m \in \mathbb{R}^{d \times d_m}, \quad b_m \in \mathbb{R}^d.
\end{equation}
A learned \emph{fusion token} $e_{\text{cls}} \in \mathbb{R}^d$, analogous to the [CLS] token in vision and language Transformers, is prepended to the modality sequence, giving $E = [e_{\text{cls}}, e_1, \ldots, e_M] \in \mathbb{R}^{(M+1)\times d}$. One self-attention with multiple heads block updates each token by attending to all tokens in the sequence \citep{vaswani2017}:
\begin{equation}
\begin{aligned}
Q &= EW_Q, \qquad K = EW_K, \qquad V = EW_V,\\
\mathrm{Attn}(Q,K,V) &= \mathrm{softmax}\!\left(\frac{QK^{\top}}{\sqrt{d_h}}\right)V,
\end{aligned}
\end{equation}
computed independently for $h=1,\ldots,H$ heads of dimension $d_h = d/H$ and recombined with an output projection $W_O \in \mathbb{R}^{d \times d}$. A residual connection and layer normalization follow. The block then applies a position-wise feed-forward network with a GELU nonlinearity before the second residual connection and layer normalization; its fusion token output row, $z = \mathrm{seq}_{0} \in \mathbb{R}^d$, supplies the fused representation for the BSI and the attention interpretation:
\begin{equation}
\hat{y} = \sigma\!\left(w^{\top} z + b_0\right), \qquad \alpha_m = \frac{1}{H}\sum_{h=1}^{H} \mathrm{Attn}^{(h)}_{\text{cls}\rightarrow m},
\end{equation}
where $\mathrm{Attn}^{(h)}_{\text{cls}\rightarrow m}$ is the weight after softmax from the fusion token to modality token $m$ in head $h$. Because the fusion token query also attends to its own key, the raw modality weights sum to slightly less than one. We therefore renormalize $\alpha_m \propto \mathrm{Attn}_{\text{cls}\rightarrow m}$ so that $\sum_m \alpha_m = 1$ and report it as an attention allocation score for each subject. Three quantities must be kept apart:
\begin{equation}\label{eq:tripartite}
\begin{aligned}
&\underbrace{\alpha_m \in [0,1],\ \textstyle\sum_{m}\alpha_m = 1}_{\text{attention allocation}},\\
&\underbrace{\Delta\mathrm{AUC}_{-m} = \mathrm{AUC}\big(f(X)\big) - \mathrm{AUC}\big(f(X_{\setminus m})\big)}_{\text{predictive contribution}},\\
&\underbrace{q_m \in [0,1]}_{\text{sensor reliability}}.
\end{aligned}
\end{equation}
Attention allocation describes how the fusion token weights the modality tokens in one layer. Predictive contribution is how much a modality changes the AUC of the complete model (value paths, residual connections, and the parallel linear branch included) when it is removed at test time. Sensor reliability is the actual quality of the recorded signal. The attention vector is not a measure of the second or third quantity; Section~\ref{app:attribution} compares it with the contribution measured by removing each modality from the complete model. Two properties of the architecture help explain this. The keys are linear in the features standardized to zero mean, so a larger noise variance does not systematically move the key directions, while the value vectors are not normalized and grow with the noise. The parallel linear path also bypasses attention. In addition, strong variation in attention across subjects was not beneficial: in scenario S0, optimizer settings that raised the SD across subjects of the weights from about 0.02 to about 0.09 lowered the held-out AUC from 0.874 to 0.769 (Table~\ref{tab:diagS0}). The construction computes token similarity after learned projections and represents the pairwise interactions described in Section~\ref{sec:whytransformer}. The same attention block processes the cognitive decline domain with four tokens and the metabolic autonomic domain with two tokens, while each domain retains its own input projection for each modality.

\subsection{A wide-and-deep hybrid containing the concatenation baseline}
\label{sec:widedeep}
Sections~\ref{sec:theory} and \ref{sec:whytransformer} motivate attention for interactions across modalities, but they do not imply that pure attention is the most direct way to recover the unique and redundant information captured by a linear model. A small Transformer must learn a linear decision boundary that logistic regression represents explicitly. We therefore adapt the \emph{wide-and-deep} architecture introduced by \citet{cheng2016widedeep}. A linear (``wide'') path processes the concatenated standardized features in parallel with the Transformer (``deep'') path, and the two contributions are summed before the sigmoid:
\begin{equation}
\hat{y} = \sigma\!\Big(\underbrace{w_{\text{deep}}^{\top} z}_{\text{Transformer path}} + \underbrace{w_{\text{wide}}^{\top} \bar{x} + b_{\text{wide}}}_{\text{linear path}}\Big), \qquad \bar{x} = [x_1; x_2; \ldots; x_M] \in \mathbb{R}^{\sum_m d_m},
\end{equation}
where gradient descent jointly optimizes the Transformer parameters and both linear weight vectors, $w_{\text{deep}} \in \mathbb{R}^d$ and $w_{\text{wide}} \in \mathbb{R}^{\sum_m d_m}$, along with the scalar bias $b_{\text{wide}}$. Setting $w_{\text{deep}} = \mathbf{0}$ exactly collapses BioSync to an $\ell_2$-regularized logistic regression over the concatenated feature vector $\bar{x}$. The hypothesis class of BioSync contains the concatenation baseline as a strict nested subspace ($\mathcal{H}_{\text{wide}} \subset \mathcal{H}_{\text{BioSync}}$). Conversely, setting $w_{\text{wide}} = \mathbf{0}$ yields the pure attention model (Section~\ref{sec:transformer}). This containment establishes representational capacity only; it does not by itself imply better generalization or a useful contribution from the attention branch, and finite sample optimization can still yield lower held-out performance. We quantify what each branch contributes to the final output, and under which conditions the attention branch carries information that the linear branch cannot, in Section~\ref{app:branches}. The experiments use \emph{BioSync} to denote the full wide-and-deep architecture and compare it with both components (Sections~\ref{sec:mainresults} and \ref{sec:ablation}).

\section{Experimental Validation}
\label{sec:results}

\subsection{Rationale and scope}
Two synthetic cohorts informed by published literature support examination of the six criteria in Section~\ref{sec:desiderata}, with graded validity evaluated in Section~\ref{sec:correlation}. Sections~\ref{sec:mainresults}--\ref{sec:correlation} cover multimodal information capture and attention allocation and attribution at the modality level; Section~\ref{sec:robustness} tests graceful degradation, and the use of the same attention block across domains with four and two modalities tests architectural generality. We assess feasibility on the device from model structure rather than deployment measurements (Section~\ref{sec:deployment}). Section~\ref{sec:extra} adds experiments from a separately written second implementation. No patient data were used, and the reported metrics do not estimate clinical diagnostic performance.

\subsection{Cohort construction}
\label{sec:cohort}
\textbf{Cognitive cohort ($M=4$).} We generated $N=360$ profiles, divided equally between cases assigned to the healthy and MCI groups. Each profile received a continuous latent severity value $\text{sev} \in [0,1]$. Values were drawn from Beta(2,6) for the healthy group and Beta(4,2.2) for the MCI group. The overlap avoids trivial class separation. Twenty-two features were generated as linear functions of severity plus independent Gaussian noise. Six came from HRV, and another six came from EEG. Actigraphy supplied six features and speech four; Section~\ref{sec:features} gives their definitions and uses effect directions from published literature.

\textbf{Metabolic cohort, AI-READI schema ($M=2$).} An independent set of $N=360$ profiles contained 180 lower severity and 180 higher severity T2D-leaning cases; a latent glycemic/autonomic severity value was drawn from the same Beta(2,6) and Beta(4,2.2) distributions.

Five features were generated as linear functions of severity plus Gaussian noise and grouped into two AI-READI schema tokens; the Cardiac-Autonomic token contains resting heart rate and respiration rate, together with the device stress score. The Behavioral token contains step count and sleep duration. Directions from published literature indicate that resting heart rate and respiration rate increase with severity, as does the stress score. Step count and sleep duration decrease.

Because every feature is a linear function of one latent severity plus independent Gaussian noise (before clipping to physiological ranges), the log-odds of the class given the features is close to linear in the concatenated features. Parity between BioSync and logistic regression on these cohorts is therefore the expected outcome, and the cohorts cannot by themselves show an advantage of attention or of weighting that varies across subjects. Section~\ref{app:generators} reports scenarios that relax this assumption, and Section~\ref{app:realdata} reports real recordings.

In both cohorts, the continuous severity value is not used as a training target; it is reserved for the correlation analysis in Section~\ref{sec:correlation}. Features are standardized within each cross-validation fold using training partition statistics only.

\subsection{Experimental setup}
We used stratified cross-validation with five folds in each cohort. Separate logistic regression models provide a baseline for every modality or token, and logistic regression on all concatenated standardized features provides the early fusion baseline. The BioSync wide-and-deep encoder uses embedding width $d=8$, $H=2$ attention heads, and one block (Section~\ref{sec:widedeep}). We implemented BioSync in JAX with automatic differentiation. After tuning the learning rate, plain gradient descent was more stable than Adam at this dataset and model scale. All reported BioSync results therefore use plain gradient descent for 500 epochs with $\ell_2$ weight decay. Accuracy, F1 score, and AUC are reported as means across the five folds. The severity analysis uses pooled out-of-fold BSI values. We also evaluated BioSync without the wide linear path and swept embedding width and head count (Section~\ref{sec:ablation}).

\subsection{Main results}
\label{sec:mainresults}
\begin{figure}[t]
\centering
\includegraphics[width=0.95\textwidth]{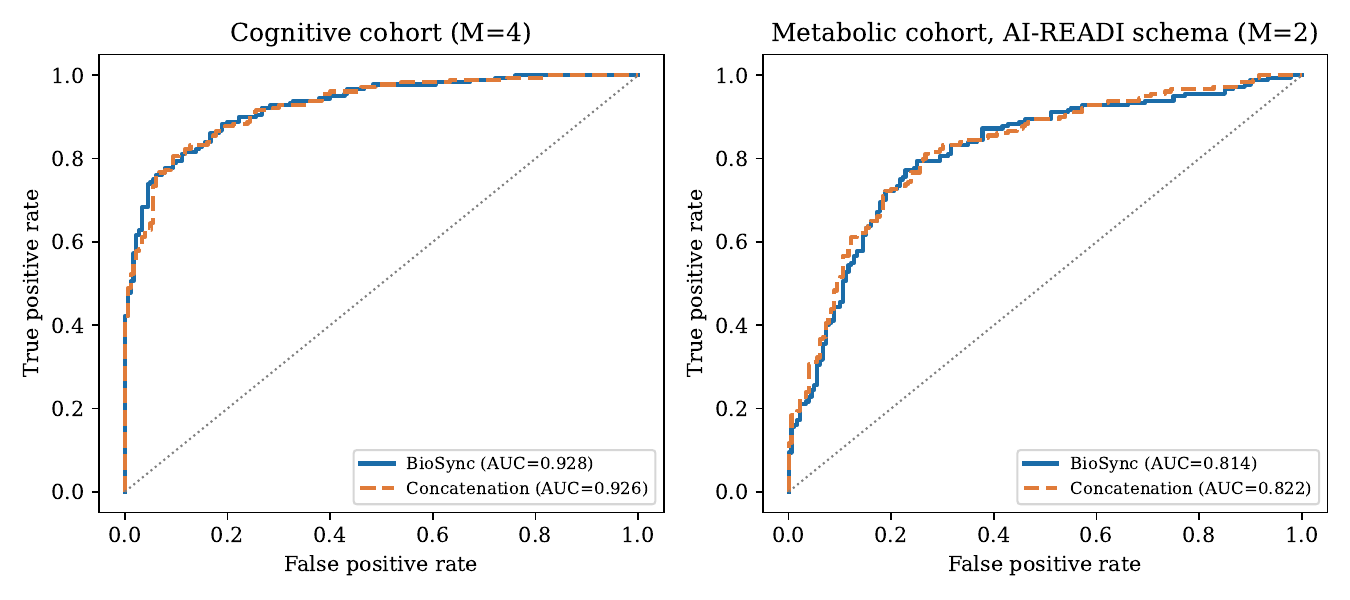}
\caption{ROC curves for BioSync versus the concatenation baseline, pooled out-of-fold predictions, both cohorts.}
\label{fig:roc}
\end{figure}

Table~\ref{tab:results} reports the cross-validated results. In the \textbf{cognitive cohort}, accuracy for a single modality ranged from 0.753 to 0.822 and AUC from 0.836 to 0.918. Concatenation exceeded every model for a single modality, with accuracy 0.850 and AUC 0.926. BioSync had the highest AUC (\textbf{0.928}), while its accuracy (0.842) and F1 score (0.839) were numerically below concatenation (0.850 and 0.847) and differed by less than one standard deviation across folds. In the \textbf{metabolic AI-READI schema cohort}, the Cardiac-Autonomic token (accuracy 0.728, AUC 0.799) outperformed the Behavioral token (accuracy 0.639, AUC 0.696). Concatenation reached accuracy 0.756 and AUC 0.823. BioSync had the highest accuracy (\textbf{0.764}) and F1 score (\textbf{0.766}), whereas its AUC (0.820, mean over folds) was slightly lower than concatenation's (0.823). The pooled out-of-fold AUCs shown in Figure~\ref{fig:roc} are 0.814 and 0.822. Figure~\ref{fig:roc} shows the pooled out-of-fold ROC curves.

BioSync ranks first in three of the six cohort and metric comparisons: cognitive cohort AUC and metabolic cohort accuracy and F1 score. The remaining differences are within or near one standard deviation across folds; these samples of $N=360$ do not support claims of large effects.

We use AUC as the primary discrimination metric because the BSI is continuous and AUC does not depend on a selected classification threshold. AUC is also commonly reported in the digital biomarker literature reviewed in Section~\ref{sec:related} \citep{qi2025}. On this metric, BioSync ranks first in the cognitive cohort; in the metabolic cohort, variation across folds does not support a clear difference from concatenation.

BioSync without the linear path did not exceed concatenation (Section~\ref{sec:ablation}). On cognitive cohort AUC, adding the wide path increased performance from 0.911 for pure attention to 0.928 for the full model, compared with 0.926 for concatenation.

\begin{table}[t]
\centering
\caption{Performance from cross-validation with five folds on both synthetic cohorts ($N=360$ each). Values are mean (standard deviation) across folds. Bold marks the best model on each metric within each cohort.}
\label{tab:results}
\small
\setlength{\tabcolsep}{4pt}
\resizebox{0.98\textwidth}{!}{%
\begin{tabular}{p{2.5cm}lccc}
\toprule
Cohort & Model & Accuracy & F1 score & AUC \\
\midrule
\multirow{6}{=}{Cognitive\\($M=4$)}
 & HRV only                     & 0.800 (0.045) & 0.798 (0.043) & 0.884 (0.044) \\
 & EEG only                     & 0.817 (0.034) & 0.813 (0.037) & 0.897 (0.034) \\
 & Actigraphy only              & 0.822 (0.034) & 0.821 (0.037) & 0.918 (0.019) \\
 & Speech only                  & 0.753 (0.028) & 0.750 (0.034) & 0.836 (0.035) \\
 & Early fusion (concatenation) & \textbf{0.850} (0.034) & \textbf{0.847} (0.036) & 0.926 (0.030) \\
 & \textbf{BioSync (wide-and-deep)} & 0.842 (0.044) & 0.839 (0.044) & \textbf{0.928} (0.028) \\
\midrule
\multirow{4}{=}{Metabolic,\\AI-READI schema\\($M=2$)}
 & Cardiac-Autonomic only       & 0.728 (0.047) & 0.721 (0.058) & 0.799 (0.038) \\
 & Behavioral only              & 0.639 (0.035) & 0.633 (0.041) & 0.696 (0.038) \\
 & Early fusion (concatenation) & 0.756 (0.032) & 0.758 (0.030) & \textbf{0.823} (0.026) \\
 & \textbf{BioSync (wide-and-deep)} & \textbf{0.764} (0.018) & \textbf{0.766} (0.018) & 0.820 (0.035) \\
\bottomrule
\end{tabular}}
\end{table}

\subsection{BSI and simulated severity}
\label{sec:correlation}
\begin{figure}[t]
\centering
\includegraphics[width=0.95\textwidth]{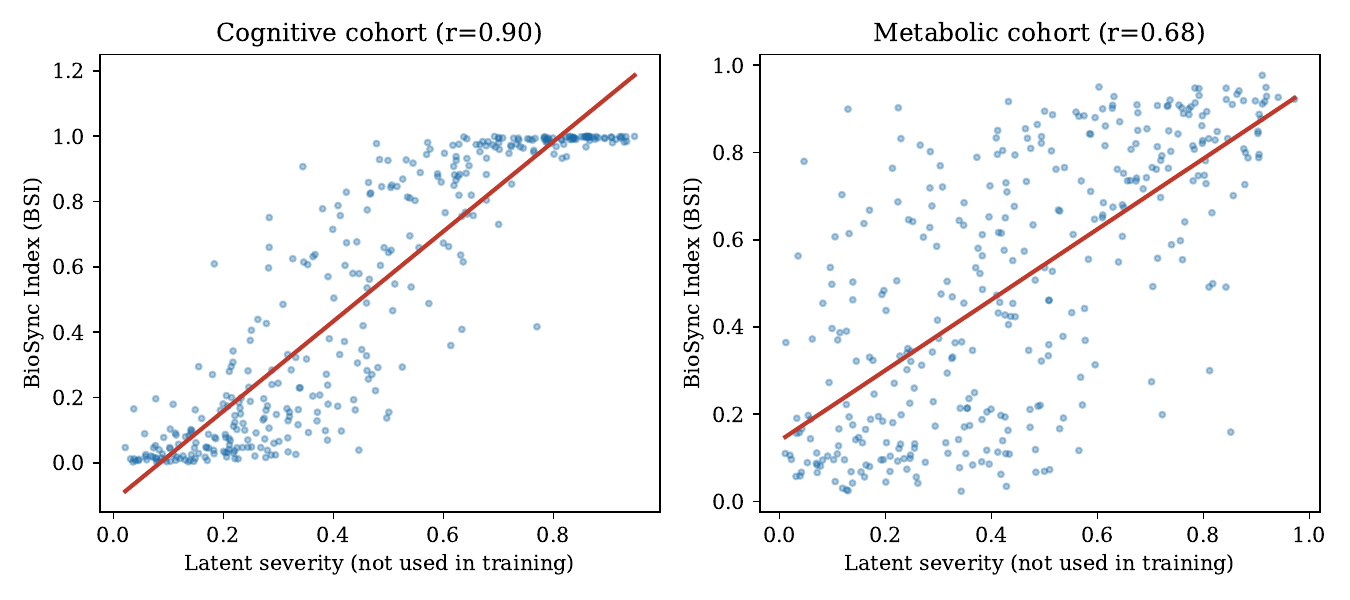}
\caption{BioSync Index (BSI) vs.\ the latent (not used for training) severity variable, both cohorts, with linear fit.}
\label{fig:severity}
\end{figure}

\begin{figure}[t]
\centering
\includegraphics[width=0.95\textwidth]{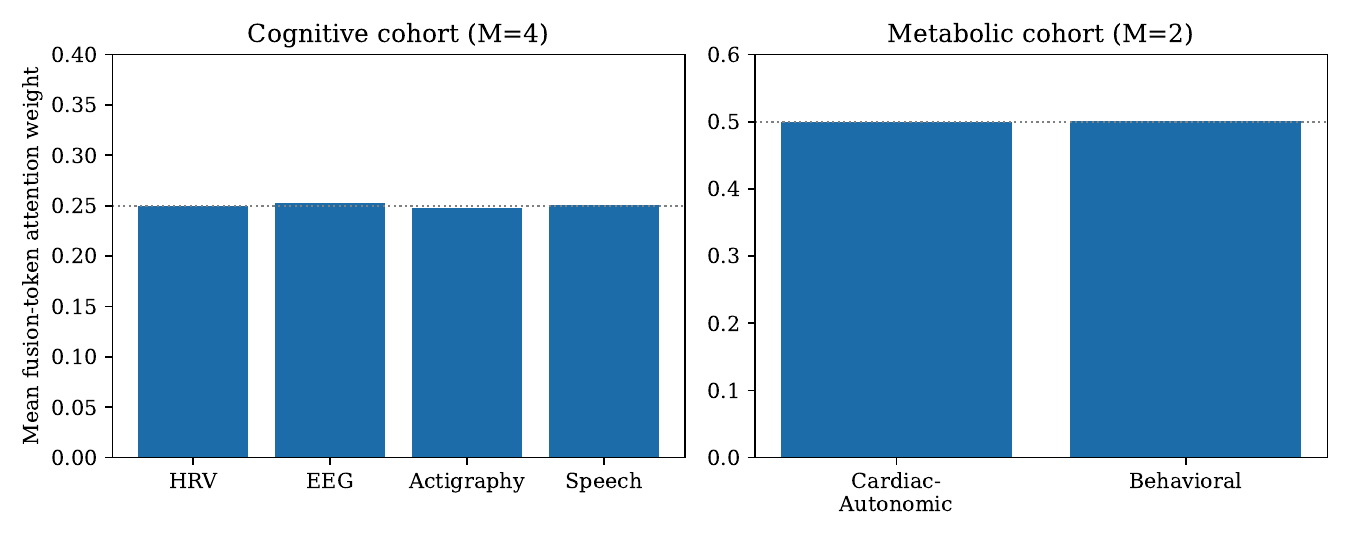}
\caption{Renormalized mean fusion token attention weight per modality, both cohorts. Dashed line marks the uniform reference ($1/M$).}
\label{fig:attention}
\end{figure}

Although latent severity was not a training target, pooled out-of-fold BSI values correlated with it: $r=0.905$ ($p<10^{-134}$) in the cognitive cohort and $r=0.681$ ($p<10^{-50}$) in the metabolic AI-READI schema cohort (Figure~\ref{fig:severity}). These results assess the \emph{graded validity} criterion in Section~\ref{sec:desiderata}. A linear model trained directly to regress concatenated features on severity achieved higher correlations: $r=0.952$ for the cognitive cohort and $r=0.696$ for the metabolic cohort. The regression model was trained for severity, so the comparison does not favor BioSync. Figure~\ref{fig:attention} shows renormalized fusion token attention weights close to uniform. They are approximately 0.25 per modality in the cognitive cohort and 0.50 per token in the metabolic cohort, a pattern consistent with the synthetic design where modality reliability is comparable and does not vary systematically by subject. The absence of weight collapse provides a clean data reference for the experiment with heterogeneous reliability in Section~\ref{sec:robustness}. Uniform mean weights alone, however, do not demonstrate adaptive interpretability. Section~\ref{sec:extra} tests whether the weights move when reliability varies and finds no consistent movement.

\subsection{Robustness to heterogeneous modality reliability}
\label{sec:robustness}
Section~\ref{sec:whytransformer} predicts that weighting that depends on the data should provide greatest utility when modality reliability varies across recordings; we evaluate this condition by replacing one randomly selected modality with high variance Gaussian noise for each test subject. The replacement probability $p$ defines the \emph{corruption rate}. This procedure models complete channel disruption rather than the continuous drift of motion artifacts, skin decoupling, or optical sensor displacement \citep{bent2020,zhang2015troika}. During training, each subject has probability 0.3 of having one randomly chosen modality replaced by the same kind of noise (modality dropout augmentation). BioSync receives a fresh random corruption in each of its 500 training epochs. Concatenation, which is fitted once, receives either three corrupted copies of the training set (as in the first submission) or 100 copies, which matches the exposure of BioSync more closely.

\begin{figure}[t]
\centering
\includegraphics[width=0.98\textwidth]{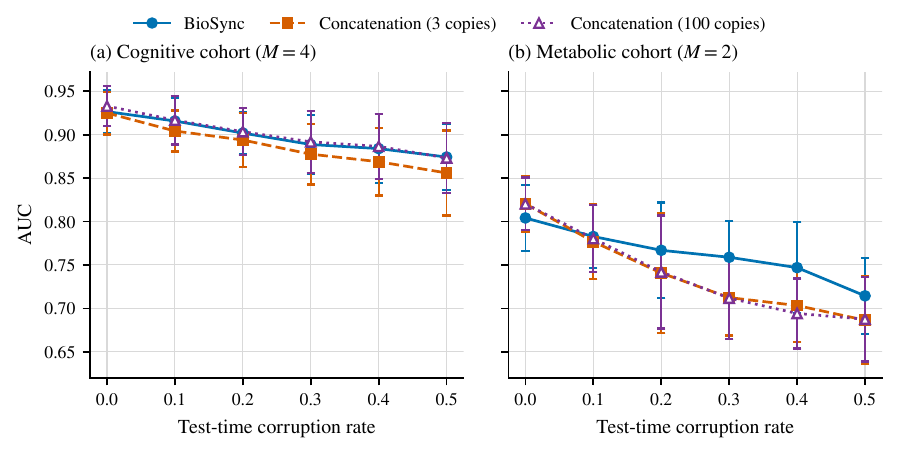}
\caption{AUC versus the rate of modality corruption at test time in the cognitive cohort (a) and the metabolic cohort (b). Points are means over 15 folds (three repeats of cross-validation with five folds), and error bars are the standard deviation across these 15 folds. BioSync draws a fresh corruption in every training epoch; concatenation is shown with 3 and with 100 corrupted copies of the training set.}
\label{fig:robustness}
\end{figure}

Figure~\ref{fig:robustness} reports AUC for $p \in \{0, 0.1, \ldots, 0.5\}$ as the mean over 15 folds (three repeats of cross-validation with five folds), with the standard deviation across the 15 folds as error bars. Two concatenation baselines are shown. Against the baseline with three corrupted copies, BioSync is ahead in the cognitive cohort at every nonzero rate, by +0.008 to +0.018 AUC (paired 95\% intervals exclude zero at $p=0.1$, 0.4, and 0.5; at $p=0.3$ the lower bound is +0.000). Against the matched baseline with 100 copies, the cognitive cohort advantage disappears: the paired differences lie between $-0.003$ and $+0.001$ for $p\ge 0.1$, every interval includes zero, and concatenation is slightly ahead at $p=0$ ($-0.006$ [$-0.009$, $-0.003$]). The advantage reported in the first submission therefore reflects unequal augmentation exposure and not attention. In the metabolic cohort, concatenation leads at $p=0$ (BioSync minus the matched baseline $-0.016$ [$-0.024$, $-0.008$]), the models are tied at $p=0.1$, and BioSync leads at $p=0.2$ to 0.5 by +0.025 to +0.053 against the matched baseline. All of these paired intervals exclude zero, and the largest difference is +0.053 [+0.035, +0.072] at $p=0.4$. This advantage persists when concatenation receives matched augmentation, so exposure does not explain it. A plausible reason is that a nonlinear model can limit the influence of a token replaced by Gaussian noise with large variance (standard deviation 4 on the standardized scale), which a linear model cannot, but we did not test this explanation. Corrupting two tokens also affects a larger part of the metabolic input than corrupting one of four cognitive tokens. These results measure one simplified form of the graceful-degradation criterion in Table~\ref{tab:desiderata}. Folds from repeated cross-validation are not independent, so the intervals are indicative. Section~\ref{sec:extra} shows that a separately written second implementation did not reproduce a BioSync advantage in the cognitive cohort.

\subsection{Architecture ablation}
\label{sec:ablation}
The first ablation isolates the linear path introduced in Section~\ref{sec:widedeep}. In the cognitive cohort, pure attention BioSync obtained accuracy 0.847 and AUC 0.911, whereas the wide-and-deep model obtained 0.842 and 0.928. On AUC, the primary discrimination metric, the full model exceeded pure attention (0.911) and concatenation (0.926). It did not exceed them on accuracy. Variation across folds does not support a large accuracy difference across the models.

\begin{figure}[t]
\centering
\includegraphics[width=0.95\textwidth]{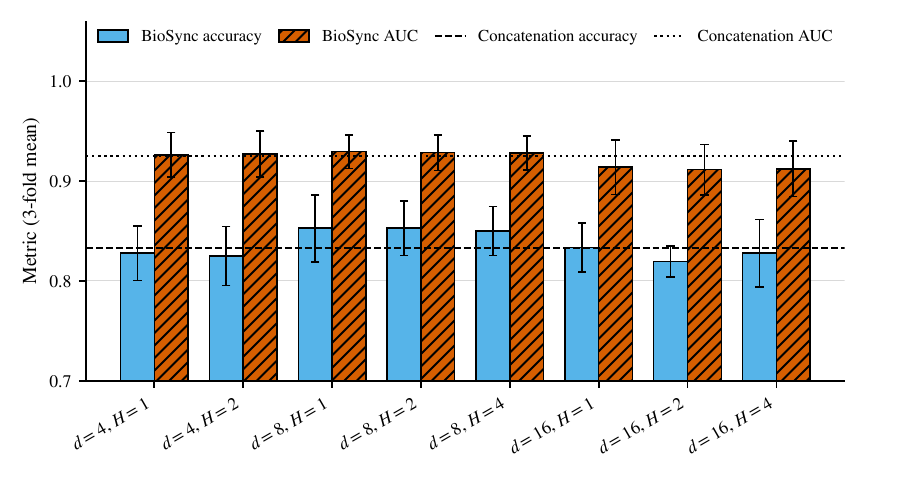}
\caption{Architecture sweep across embedding width $d$ and attention heads $H$ in the cognitive cohort. The full wide-and-deep model was evaluated by cross-validation using three folds. Bars show mean performance, and error bars show standard deviations across folds. Light bars indicate accuracy; hatched bars indicate AUC. Dashed and dotted horizontal lines mark the corresponding accuracy and AUC for concatenation under the same folds.}
\label{fig:ablation}
\end{figure}

We also varied cognitive cohort embedding width $d \in \{4,8,16\}$ and attention heads $H \in \{1,2,4\}$, subject to $H \mid d$ (Figure~\ref{fig:ablation}; cross-validation with three folds). Across the grid, accuracy ranged from 0.819 to 0.853 and AUC from 0.911 to 0.930. Under the same folds, concatenation reached accuracy 0.833 and AUC 0.925. The prespecified $d=8,H=2$ model obtained 0.853/0.929. The highest point estimate of AUC was $d=8,H=1$ (0.930). We retain $d=8,H=2$ because the robustness experiment preceded the sweep and changing the reported model afterward would constitute post hoc selection. Embedding width 16 gave lower AUC (0.911--0.914) than widths 4 and 8 (0.926--0.930), possibly reflecting overfitting at this sample size, and no configuration was clearly better than concatenation (AUC 0.925) under cross-validation with three folds.

\FloatBarrier
\subsection{Second implementation, artifact bursts, and a cohort in which outcomes depend only on interactions across modalities}
\label{sec:extra}
The earlier experiments replaced an entire channel with random noise and omitted interaction structure. Their generator parameters were also not fully specified. We addressed these limitations by writing a second implementation of the architecture in JAX \citep{bradbury2018jax}. The author developed it after the first pipeline. No independent external team validated it. It uses logistic regression and gradient-boosted trees from scikit-learn \citep{pedregosa2011sklearn}. The code will be released upon acceptance, as noted in the Data Availability statement. The cohort generator follows Section~\ref{sec:cohort}. We calibrated noise for each modality so that logistic regression AUCs for individual modalities match Table~\ref{tab:results} (target 0.88, 0.90, 0.92, 0.84). The model uses $d=8$ and $H=2$, with one block and a feed-forward width of $4d$ (1,119 parameters, Section~\ref{sec:deployment}). Learning rates and weight decay were selected on independent development cohorts, never on evaluation cohorts. Values are means (SD) over 15 folds (three repeats of stratified cross-validation with five folds). The artifact experiments use 10 folds (two repeats). Because the generator and hyperparameters differ from those of the original pipeline, the two sets of numbers are not directly comparable and should not be pooled. The original code for this section could not be recovered. We regenerated every number, table, and figure in this section with new code written from the same description. The original run seeds are unknown, so the values differ slightly from those previously reported. Section~\ref{sec:extra} uses the same hyperparameters in all scenarios (chosen on development cohorts), and the generator dials that were calibrated are listed in Section~\ref{app:params}.

\paragraph{Clean data and noise level} On the calibrated cohort, BioSync reached AUC 0.929 (0.022), compared with 0.925 (0.023) for concatenation and 0.898 (0.027) for pure attention (Table~\ref{tab:extra_clean}). The paired BioSync minus concatenation difference was +0.004 (SD 0.009), with BioSync higher in 9 of 15 folds. Pure attention was lower on average (-0.026 relative to concatenation); the BSI correlated with latent severity at $r=0.935$ (SD 0.003 across repeats). As the noise multiplier for features increases through $\kappa\in\{0.6,1,1.5,2\}$, the paired BioSync minus concatenation difference stays between +0.003 and +0.005, far below the SD across folds (Table~\ref{tab:extra_sweep}). The clean data differences in Table~\ref{tab:results} are compatible with parity.

\begin{table}[t]
\centering
\caption{Second implementation, calibrated cognitive cohort ($N=360$, $\kappa=1$). Values are mean (SD) over 15 folds. Accuracy and F1 use a 0.5 threshold.}
\label{tab:extra_clean}
\small
\begin{tabular}{lccc}
\toprule
Model & Accuracy & F1 score & AUC \\
\midrule
HRV only & 0.787 (0.042) & 0.787 (0.037) & 0.871 (0.045) \\
EEG only & 0.807 (0.042) & 0.802 (0.043) & 0.889 (0.034) \\
Actigraphy only & 0.822 (0.050) & 0.823 (0.046) & 0.917 (0.023) \\
Speech only & 0.805 (0.037) & 0.807 (0.033) & 0.868 (0.027) \\
Early fusion (concatenation) & 0.844 (0.041) & 0.844 (0.037) & 0.925 (0.023) \\
BioSync (wide-and-deep) & 0.860 (0.029) & 0.858 (0.026) & 0.929 (0.022) \\
Pure attention (no linear path) & 0.838 (0.038) & 0.834 (0.036) & 0.898 (0.027) \\
\bottomrule
\end{tabular}
\end{table}

\begin{table}[t]
\centering
\caption{AUC as the noise multiplier for features $\kappa$ increases (second implementation). The penultimate column reports the mean paired BioSync minus concatenation AUC difference with its SD across folds; the final column counts folds in which BioSync was higher.}
\label{tab:extra_sweep}
\small
\begin{tabular}{lccccc}
\toprule
$\kappa$ & Concat. & BioSync & Pure att. & Difference & Wins \\
\midrule
0.6 & 0.934 & 0.940 & 0.919 & +0.005 (0.006) & 12/15 \\
1.0 & 0.925 & 0.929 & 0.898 & +0.004 (0.009) & 9/15 \\
1.5 & 0.904 & 0.910 & 0.869 & +0.005 (0.009) & 10/15 \\
2.0 & 0.881 & 0.884 & 0.833 & +0.003 (0.010) & 11/15 \\
\bottomrule
\end{tabular}
\end{table}

\paragraph{Temporally correlated artifact bursts} Each recording is divided into 48 analysis windows. Artifact episodes follow a Markov chain with two states \citep{gilbert1960,elliott1963}; their mean length is six windows, and their stationary burden is $\rho\in\{0,0.1,\dots,0.5\}$. Windows inside an episode add an offset to every feature of the affected modality. Offsets are half-normal and independent across features and episodes. We scale them by the feature's SD across subjects, then average them into the summary at the recording level. The same burden can thus produce different errors when episodes are few and long rather than many and short. In scenario A, one burst process hits HRV and EEG together, as wristband slippage and dry electrode motion affect PPG during activity \citep{zhang2015troika}. Actigraphy and speech remain unaffected. Each modality has an independent burst process in scenario B. Scenario C reproduces the protocol in Section~\ref{sec:robustness} that replaces an entire channel with noise. Logistic regression on a development cohort set the offset scale (12.7 times the feature SD) so that scenario B AUC fell by about 0.1 at $\rho=0.3$. BioSync was not used for tuning. All models trained on the same augmented training sets (six copies, $\rho=0,\dots,0.5$, $p=0.3$ for scenario C). Testing used four independent corruption draws per recording.

\begin{figure}[t]
\centering
\includegraphics[width=\textwidth]{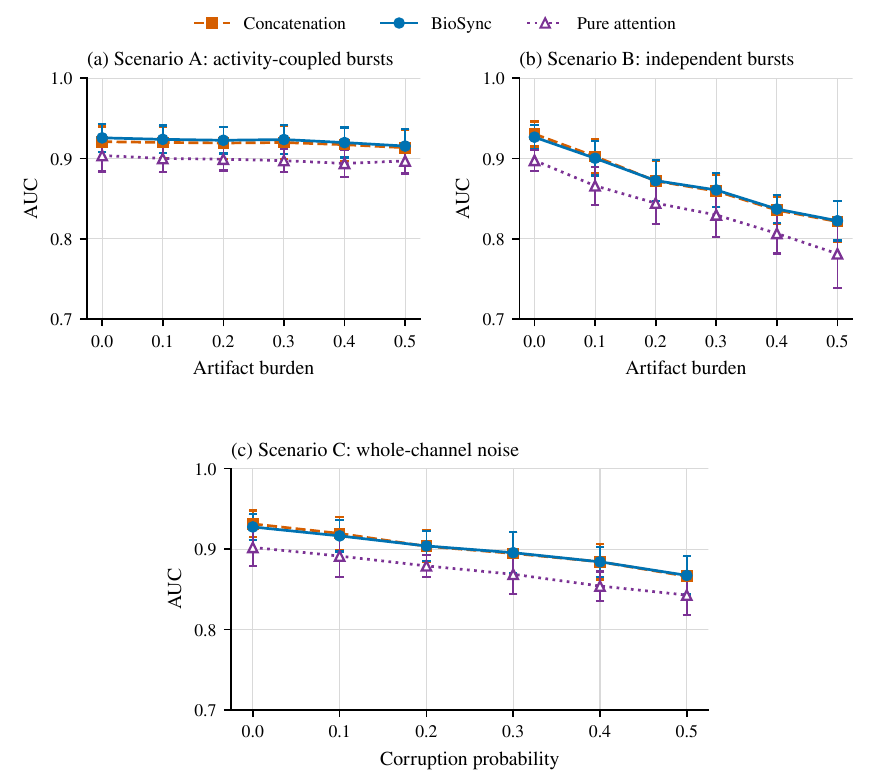}
\caption{Burst and noise replacement experiments with the second implementation (synthetic data). Panels (a)--(c): AUC (mean $\pm$ SD over 10 folds) versus artifact burden for concatenation, BioSync, and pure attention under scenario A (bursts coupled to activity hitting HRV and EEG together), scenario B (independent bursts in each modality), and scenario C (whole channel noise replacement).}
\label{fig:extra}
\end{figure}

Concatenation and BioSync were indistinguishable in every scenario (Figure~\ref{fig:extra}). The BioSync minus concatenation AUC difference ranged from +0.002 to +0.005 in scenario A, from -0.004 to +0.001 in scenario B, and from -0.004 to +0.001 in scenario C. Each range was far below the SD across folds of 0.02--0.03. Scenario A changed every model little (concatenation AUC 0.921 to 0.913) because actigraphy and speech remained clean, allowing augmentation to learn a fixed reduction in the weights of HRV and EEG without adaptation for each recording. Scenarios B and C produced larger declines in concatenation AUC (0.930 to 0.821 and 0.931 to 0.866, respectively), because any channel could fail in these settings. BioSync did not degrade more gracefully, and the advantage in Section~\ref{sec:robustness} was not reproduced. In the first submission, scenario C was reported to decline from 0.931 to 0.734; that value could not be reproduced, and the present value is close to the value reported for the corruption experiment of Section~\ref{sec:robustness} in the first submission at the same corruption rate (0.865 at $p=0.5$).

\begin{figure}[t]
\centering
\includegraphics[width=\textwidth]{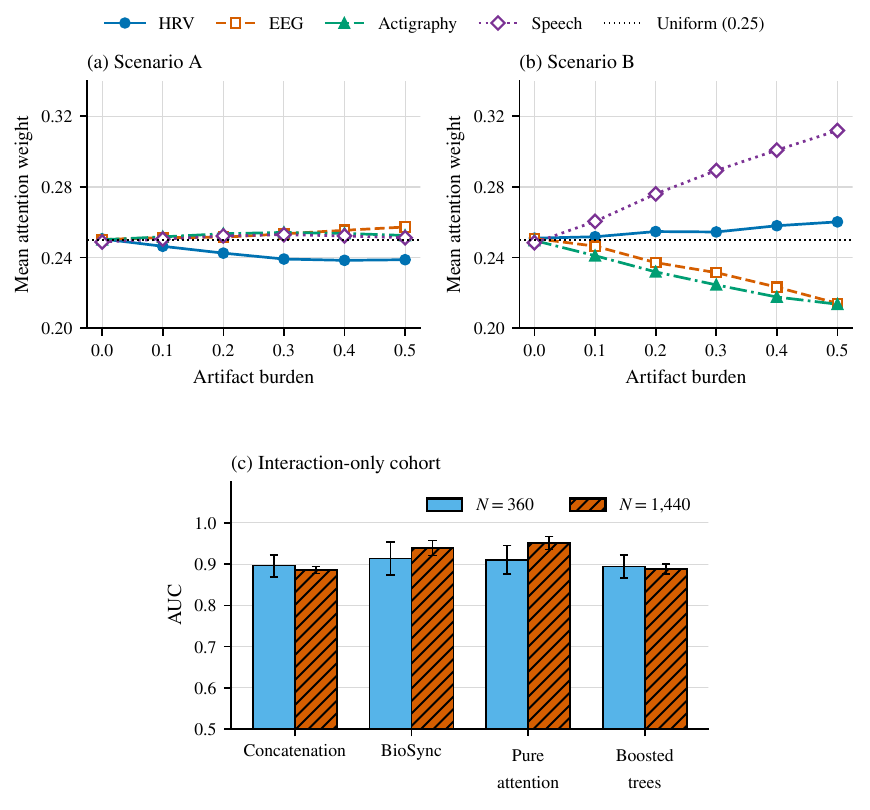}
\caption{Attention weights and the cohort in which outcomes depend only on interactions across modalities (second implementation, synthetic data). Panels (a) and (b): mean fusion token attention weight per modality under scenarios A and B (dotted line: uniform weight 0.25). Panel (c): AUC on the cohort in which outcomes depend only on interactions across modalities at $N=360$ and $N=1{,}440$ (mean $\pm$ SD over folds).}
\label{fig:extrab}
\end{figure}

\paragraph{Attention weights under heterogeneous reliability} The mean fusion token weights stayed between 0.214 and 0.312 at every burden, against a uniform value of 0.25 (Figure~\ref{fig:extrab}, panels (a) and (b)). The direction was inconsistent. Within a burden level in scenario A, the correlation between a modality's realized artifact fraction and the weight it received ranged from -0.08 to +0.02 for HRV and from +0.03 to +0.06 for EEG, although scenario A corrupts the two identically. In this setting, the weights do not reliably identify the corrupted channel, and we do not claim that they track modality reliability.

\paragraph{Cohort in which the outcome depends only on interactions} To test the synergy argument of Section~\ref{sec:theory} formalized by Partial Information Decomposition \citep{williams2010pid}, we generated a synthetic cohort in which the diagnostic class is the sign of the product of an HRV latent and an EEG latent ($Y = \mathrm{sign}(z_{\text{HRV}} \cdot z_{\text{EEG}})$). Neither modality alone carries marginal information: at $N=360$, logistic regression AUC for a single modality was 0.534 for HRV and 0.514 for EEG. Actigraphy and speech carry weak marginal signals (effect 0.28 per feature, set so that concatenation reaches AUC 0.89 at $N=1{,}440$). The noise of the HRV and EEG features was fixed at 0.6 in advance. With 360 profiles, BioSync (0.913) and pure attention (0.910) exceeded concatenation (0.896) and boosted trees (0.894). With 1,440 profiles, BioSync reached 0.939 and pure attention 0.951, compared with 0.886 for concatenation and 0.888 for boosted trees (Figure~\ref{fig:extrab}, panel (c)). The attention models therefore recovered interaction information that the linear baseline could not. In the first submission, this advantage appeared only at 1,440 profiles. In the present cohort it is already present at 360, so the sample size at which attention starts to help depends on the strength of the interaction and is not a fixed threshold (Section~\ref{app:branches} maps it over sample size and interaction noise). The linear path did not help when the interaction dominated: pure attention was at least as good as BioSync at $N=1{,}440$. Hyperparameters for this cohort were chosen on separate development cohorts at $N=360$ only.

\subsection{Published reference values for real cohort validation}
\label{sec:fieldpositioning}
The main results and ablations above provide the only controlled comparisons in this study. Published digital biomarker studies differ in cohorts, tasks, feature sets, and metrics, and the synthetic effect sizes used here were set by us. We therefore do not rank BioSync against published methods. Table~\ref{tab:reference} instead lists reported values from the literature reviewed in Section~\ref{sec:related} as targets that a real cohort evaluation of BioSync should be judged against. These values are not directly comparable with one another or with the synthetic results in Table~\ref{tab:results}.

\begin{table}[htbp]
\centering
\caption{Published performance values used as reference targets for real cohort validation.}
\label{tab:reference}
\small
\begin{tabularx}{\textwidth}{l>{\raggedright\arraybackslash}Xlc}
\toprule
Source & Population / task & Metric & Value \\
\midrule
\citet{teh2022mci} & Meta-analysis, MCI and pre-frailty digital biomarkers (mostly single modality) & Pooled sens./spec. & $\approx$0.80 \\
\citet{qi2025} & 45 MCI-focused AI models (scoping review of 431 studies) & Mean AUC & 0.821 \\
\citet{qi2025} & 21 AD-focused AI models (same review) & Mean AUC & 0.887 \\
\citet{li2023synergy} & Late fusion of digital cognitive tests and wearable measures & Accuracy & 0.889 \\
\citet{xu2025language} & Late fusion of language digital biomarkers & Accuracy & 0.918 \\
\bottomrule
\end{tabularx}
\end{table}

The reported means for real cohorts (AUC 0.82--0.89 across AI models in the review) are lower than the synthetic cognitive cohort AUC of 0.928. This gap is expected because the synthetic features are linear functions of a single latent severity plus independent noise. It should not be read as expected performance in real world settings. Real recordings add variability between subjects and correlated noise; device effects and label uncertainty add further difficulty.

\FloatBarrier
\subsection{Comparison against alternative fusion designs}
\label{sec:comparison}
Table~\ref{tab:desiderata} compares specific implementations on measured criteria. An earlier version of this comparison rated whole method families with categorical symbols, which is not justified: soft voting can produce continuous scores, nonlinear models on concatenated features can model interactions, and tree and neural models can support attribution for each subject. The criteria, defined in the caption, are discrimination averaged over the synthetic scenarios, the ability to model interactions (the cohort in which outcomes depend only on interactions across modalities), calibration, robustness to lost modalities, model size, and the attribution provided by each model. Where we did not measure a criterion for an implementation, the table says so. Architectural extensibility (the same encoder with two or four tokens, Section~\ref{sec:features}) and feasibility on the device (analytic parameter count, Section~\ref{sec:deployment}) are architectural properties, not demonstrated transfer or measured deployment performance, and we do not rate them. Averaged over the seven single-axis synthetic scenarios S0--S6, the mean AUC was 0.829 for concatenated logistic regression, 0.847 for soft voting, 0.809 for boosted trees, 0.806 for the MLP, and 0.833 for BioSync. On the interaction-only cohort at $N=1{,}440$ the AUC was 0.892 for logistic regression, 0.890 for soft voting, 0.919 for boosted trees, 0.975 for the MLP, and 0.935 for BioSync. At $N=360$ the corresponding AUCs were 0.953 for the MLP, 0.955 for the SVM, and 0.904 for BioSync. No design dominates. BioSync is competitive on the additive scenarios, and its nonlinear branch helps when interactions dominate. However, tuned nonlinear models on concatenated features (the MLP and the SVM) exceeded BioSync on the interaction-only cohort, so the attention branch is not the only, nor the most efficient, way to model interactions, and soft voting is the strongest design on several additive scenarios. BioSync alone offers an attention vector, but that vector is attention allocation and not a contribution measure.
\par

\begin{table}[t]
\centering
\caption{Comparison of specific implementations on measured criteria (synthetic data; see the notes below the table). n/m: not measured.}
\label{tab:desiderata}
\footnotesize
\setlength{\tabcolsep}{3pt}\renewcommand{\arraystretch}{1.15}
\begin{tabular}{p{0.18\linewidth}p{0.09\linewidth}p{0.11\linewidth}p{0.07\linewidth}p{0.10\linewidth}p{0.17\linewidth}p{0.14\linewidth}}
\toprule
Implementation & Mean AUC S0--S6 & AUC, interaction $N{=}1440$ & Mean ECE & AUC, 30\% modality loss & Model size, kB / MACs & Per-subject attribution \\
\midrule
Concatenation + logistic regression & 0.829 & 0.892 & 0.065 & 0.845 & 0.09 / 22 & coefficients \\
Soft voting over per-modality LR & 0.847 & 0.890 & n/m & 0.858 & 0.10 / 22 & per-modality scores \\
Boosted trees (tuned) & 0.809 & 0.919 & 0.103 & n/m & 26.2--43.7 / 0$^\dagger$ & tree attributions \\
MLP (tuned) & 0.806 & 0.975 & n/m & n/m & 1.54 / 368 & gradient / permutation \\
SVM, RBF kernel (tuned) & 0.838 & 0.969 & n/m & n/m & 19.6--45.1 / 4.7--10.8k$^\ddagger$ & permutation \\
BioSync (wide-and-deep) & 0.833 & 0.935 & 0.058 & 0.846 & 4.6 (1.3 INT8) / 4{,}450 & attention allocation$^\ast$ \\
\bottomrule
\end{tabular}
\end{table}
\noindent\begin{minipage}{\linewidth}\footnotesize\emph{Notes to Table~\ref{tab:desiderata}.} Criteria: mean AUC over scenarios S0--S6 (Table~\ref{tab:grid}); AUC on the interaction-only cohort at $N=1440$, which measures the ability to model interactions; mean expected calibration error over S0, S2, and S6 (Table~\ref{tab:calib}); AUC in S0 when each modality is lost with probability 0.3 (Table~\ref{tab:missingS0}); stored model size in kB (float32) and multiply-accumulates (MACs) per inference, measured on the fitted models (range over S0, S6, and S7 at $N=1440$; tuned models refit on the first outer fold of four development seeds; Appendix~\ref{app:confirm}); and the attribution that the model offers. For BioSync the size is that of the quality-gated network of Appendix~\ref{app:confirm} (1,159 parameters; weight-only INT8 quantization changed the AUC by at most 0.0008). $^\dagger$Trees need only comparisons along one path per tree (325--450 per boosted ensemble, 1,030--2,919 per forest in the cohorts measured); no multiply-accumulates. $^\ddagger$Thousand multiply-accumulates (SVM: support vectors times features), plus one kernel exponential per support vector (213--490). $^\ast$Attention allocation is not a measure of contribution or reliability (Section~\ref{app:attribution}). Sizes and operation counts are measured; none of the models was run on microcontroller hardware, so latency, energy, and transfer to other domains are not rated.\end{minipage}\par\medskip

\FloatBarrier
\subsection{Additional evaluations}
\label{sec:review}
This subsection describes the synthetic generator parameters and their sources (Sec.~\ref{app:params}), the robustness of the conclusions to the choice of generator (Sec.~\ref{app:generators}), the contribution of each branch (Sec.~\ref{app:branches}), attention allocation versus predictive contribution (Sec.~\ref{app:attribution}), calibration, stability, and the comparison with other fusion designs (Sec.~\ref{app:calibration}), and the evaluation on real recordings (Sec.~\ref{app:realdata}). Additional analyses (variants of the method, a quality gate and a mixture model with a confirmation run, further real datasets, and a real data test of the quality gate) are reported in Appendices~\ref{app:variants}, \ref{app:confirm}, and~\ref{app:realpass}.

\subsubsection{Generator parameters}
\label{app:params}
This subsection lists every parameter of the synthetic generators. The generator is an independently designed simulation. Effect directions follow the literature cited in Section~\ref{sec:features}, but the numerical intercepts, slopes, and noise scales are author choices and are not empirical estimates from a named cohort. Each feature is $x = \text{intercept} + \text{slope}\times\text{severity} + \varepsilon$, with $\varepsilon\sim\mathcal{N}(0,\text{SD}^2)$, clipped to the floor and ceiling shown (a dash means no bound). Severity is drawn from Beta(2,\,6) for the lower severity group and Beta(4,\,2.2) for the higher severity group. The class label is the group. Features are standardized inside each cross-validation fold.

\begin{table}[htbp]
\centering
\caption{Cognitive cohort ($M=4$, 22 features), first pipeline (Table~\ref{tab:results}) and the base of the second implementation. In the second implementation the noise SD of each modality is multiplied by a calibrated factor so that its logistic regression AUC for a single modality matches the target (0.955 for HRV, 0.932 for EEG, 0.769 for actigraphy, 1.149 for speech; targets 0.88, 0.90, 0.92, 0.84), and by the multiplier $\kappa$ in the noise sweep.}
\label{tab:genparams}
\small
\resizebox{\linewidth}{!}{%
\begin{tabular}{llrrrrr}
\toprule
Modality & Feature & Intercept & Slope & Noise SD & Floor & Ceiling \\
\midrule
HRV & SDNN & 65 & -22 & 9 & 8 & -- \\
HRV & RMSSD & 42 & -16 & 7 & 5 & -- \\
HRV & pNN50 & 18 & -9 & 4 & 0.5 & 60 \\
HRV & LF power & 650 & -150 & 90 & 50 & -- \\
HRV & HF power & 550 & 80 & 90 & 50 & -- \\
HRV & LF/HF ratio & 1.3 & 1.1 & 0.35 & 0.2 & -- \\
EEG & theta-to-alpha ratio & 0.9 & 0.7 & 0.18 & 0.2 & -- \\
EEG & Gamma spectral entropy & 0.78 & -0.28 & 0.09 & 0.1 & 1 \\
EEG & Frontal theta power & 1 & 0.55 & 0.2 & 0.1 & -- \\
EEG & Parietal alpha power & 1.1 & -0.45 & 0.2 & 0.1 & -- \\
EEG & Beta power & 0.9 & -0.2 & 0.15 & 0.1 & -- \\
EEG & Lempel--Ziv complexity & 0.62 & -0.22 & 0.08 & 0.05 & 1 \\
ACT & Rest--activity fragmentation & 0.28 & 0.32 & 0.09 & 0.02 & 1 \\
ACT & Circadian amplitude & 0.72 & -0.3 & 0.1 & 0.05 & 1 \\
ACT & Sleep efficiency & 0.86 & -0.18 & 0.07 & 0.3 & 0.99 \\
ACT & Step count CV & 0.35 & 0.28 & 0.08 & 0.05 & 1.2 \\
ACT & Interdaily stability & 0.68 & -0.24 & 0.09 & 0.05 & 1 \\
ACT & Intradaily variability & 0.55 & 0.3 & 0.1 & 0.05 & 1.5 \\
Speech & Pause-to-speech ratio & 0.18 & 0.22 & 0.07 & 0.02 & 0.9 \\
Speech & Articulation rate & 3.4 & -0.9 & 0.4 & 0.8 & -- \\
Speech & Pitch variability & 0.55 & -0.2 & 0.1 & 0.05 & 1 \\
Speech & Semantic density & 0.62 & -0.24 & 0.09 & 0.05 & 1 \\
\bottomrule
\end{tabular}}
\end{table}

\begin{table}[htbp]
\centering
\caption{Metabolic cohort, AI-READI schema ($M=2$, 5 features), first pipeline. Same functional form as Table~\ref{tab:genparams}.}
\label{tab:genparams2}
\small
\resizebox{\linewidth}{!}{%
\begin{tabular}{llrrrrr}
\toprule
Token & Feature & Intercept & Slope & Noise SD & Floor & Ceiling \\
\midrule
Cardiac-Autonomic & Resting heart rate & 66 & 9 & 6 & 45 & 110 \\
Cardiac-Autonomic & Respiration rate & 14 & 2.2 & 1.8 & 8 & 26 \\
Cardiac-Autonomic & Device stress score & 32 & 22 & 10 & 0 & 100 \\
Behavioral & Step count & 7800 & -2600 & 1400 & 500 & -- \\
Behavioral & Sleep duration (h) & 7.1 & -0.9 & 0.6 & 3 & 10 \\
\bottomrule
\end{tabular}}
\end{table}

\paragraph{Scenario generators (Section~\ref{app:generators})} All scenarios use 22 features in four modalities (6, 6, 6, 4), severity as above with $N=360$ and balanced classes, a random effect direction ($\pm1$) and gain (uniform on $[1,2]$) per feature, and unit variance Gaussian noise unless stated. S0: each feature equals direction $\times$ gain $\times$ severity plus noise. S1: each modality is driven by its own latent, $0.6\,\text{severity}+\sqrt{1-0.6^2}\,u_m$, with $u_m$ independent and matched in mean and SD to severity. S2: each subject expresses impairment in a random subset of modalities (each with probability 0.5, at least one). Modalities not expressed keep 15\% of their gain. S3: the speech modality has zero gain. S4: modalities 1 and 3 follow a convex function of severity ($s^2/0.35$), modalities 2 and 4 a steep logistic function ($1/(1+e^{-12(s-0.45)})$), and a quarter of subjects have one randomly chosen modality with reversed sign. S5: modality noise SDs of 0.8, 1.2, 1.8, and 0.6, heavy-tailed Student-$t$ ($3$ degrees of freedom, unit variance) noise with within modality AR(1) correlation 0.7, and shared across modalities noise of SD 0.6, 0.6, 0.6, and 0.4. S6: S0 plus, for each subject and modality independently with probability 0.35, extra Gaussian noise of SD $\sqrt{8}$ (total SD 3). A noisy quality indicator $q=1-\text{degraded}+\mathcal{N}(0,0.2^2)$ is available to the quality aware models. S7: the class is the sign of the product of an HRV latent and an EEG latent (both standard normal). HRV and EEG features are the latent plus noise of SD 0.6 (fixed in advance). Actigraphy and speech features equal $\pm0.283$ by class plus unit noise, with 0.2825 calibrated so that concatenated logistic regression reaches AUC 0.89 at $N=1{,}440$.

\paragraph{Dials calibrated or fixed} Noise factors for each modality are listed above. The burst offset scale was 12.7 times the feature SD, calibrated so that logistic regression AUC falls by about 0.1 at burden 0.3 in scenario B. The linear effect was 0.2825 and latent noise was 0.6 for S7. Hyperparameters of the networks were chosen on development cohorts (clean generator: learning rate 0.02, weight decay 0.0001, 500 full batch epochs; interaction cohort: learning rate 0.05, 500 epochs) and are not tuned per scenario.
\FloatBarrier

\subsubsection{Generator robustness and realism}
\label{app:generators}
The cohorts of Section~\ref{sec:cohort} use one latent severity variable and independent Gaussian noise, so the log-odds of the class is close to linear in the features and parity with concatenation is expected. To test whether the conclusions depend on this choice, we ran the models on eight generators (nine settings, with S7 at two sample sizes; Section~\ref{app:params} defines them): S0 the design with a single latent variable, S1 latent factors specific to each modality, S2 heterogeneous impairment profiles, S3 an uninformative modality, S4 nonlinear relations and relations with inconsistent signs, S5 structured and heavy-tailed noise, S6 modality reliability that varies across subjects, and S7 a cohort in which outcomes depend only on interactions across modalities at two sample sizes. Each scenario used eight fresh data seeds (5000--5007), cross-validation with five folds inside each seed, boosted trees, random forest, MLP, and SVM tuned with inner cross-validation with three folds, and BioSync hyperparameters fixed from the development cohorts (not tuned per scenario). The analysis plan was written before the runs (evidence/analysis\_plan\_2026-11-05.md). Deviations are logged there. Tables~\ref{tab:grid}, \ref{tab:gridpaired}, and \ref{tab:gridpaired2} give the results. Intervals are bootstrap 95\% intervals over the eight seeds.

\paragraph{Results} Against concatenated logistic regression, BioSync was ahead with an interval that excludes zero in S0, S1, S2, S4, S7 ($N=360$) and S7 ($N=1440$), tied in S3, S5 and S6, and behind in none. The paired differences ranged from +0.001 to +0.011 AUC in S0--S6, so the gains there are small, and were +0.034 (S7 ($N=360$)) and +0.043 (S7 ($N=1440$)) in the cohort in which outcomes depend only on interactions across modalities. The best model by mean AUC was soft voting in S0, S1, S3 and S5, the SVM in S2, S4 and S7 ($N=360$), random forest in S6 and the MLP in S7 ($N=1440$). Soft voting was ahead of BioSync in S0, S1, S2, S3, S4, S5 and S6 and tied in none. BioSync was ahead of boosted trees in S0, S1, S2, S3, S5, S7 ($N=360$) and S7 ($N=1440$), of the MLP in S0, S1, S2, S3, S5 and S6, of pure attention in S0, S1, S2, S3, S4, S5 and S6, and of the residual variant in S0, S1, S2, S3, S4, S6, S7 ($N=360$) and S7 ($N=1440$). Results across generators are consistent with the main text. The wide-and-deep model matches or slightly exceeds concatenated logistic regression in S0--S6 and clearly exceeds it in the cohort in which outcomes depend only on interactions across modalities. It does not beat soft voting or the best tuned nonlinear baseline in most scenarios. The only setting in which the attention branch clearly adds information is the cohort in which outcomes depend only on interactions across modalities (Section~\ref{app:branches}). The severity correlation of Section~\ref{sec:correlation} depends on the generator and is not independent evidence for the BSI. Section~\ref{app:realdata} compares the generator with real recordings.

\IfFileExists{tables_B.tex}{\begin{widepage}\begin{table}[t]
\centering
\caption{Scenario grid (synthetic): AUC, mean (SD) over data seeds; five-fold cross-validation within each seed; baselines tuned with inner three-fold CV; BioSync hyper-parameters fixed from development cohorts. Seeds 5000--5007.}
\label{tab:grid}
\small
\resizebox{\linewidth}{!}{%
\begin{tabular}{lcccccccccc}
\toprule
Scenario & Seeds & Concat.\ LR & Soft vote & Boosted trees & Random forest & MLP & SVM (RBF) & BioSync & Pure attention & BioSync-residual \\
\midrule
S0 single latent & 8 & 0.866 (0.020) & 0.883 (0.019) & 0.835 (0.022) & 0.862 (0.018) & 0.842 (0.025) & 0.871 (0.017) & 0.870 (0.020) & 0.830 (0.022) & 0.861 (0.023) \\
S1 modality-specific latents & 8 & 0.759 (0.033) & 0.776 (0.032) & 0.719 (0.034) & 0.753 (0.033) & 0.718 (0.040) & 0.768 (0.030) & 0.763 (0.030) & 0.720 (0.031) & 0.751 (0.026) \\
S2 heterogeneous profiles & 8 & 0.758 (0.025) & 0.768 (0.021) & 0.736 (0.041) & 0.764 (0.014) & 0.723 (0.032) & 0.775 (0.025) & 0.762 (0.023) & 0.713 (0.030) & 0.747 (0.025) \\
S3 uninformative modality & 8 & 0.861 (0.021) & 0.876 (0.019) & 0.828 (0.026) & 0.855 (0.021) & 0.835 (0.026) & 0.864 (0.020) & 0.865 (0.022) & 0.817 (0.021) & 0.855 (0.025) \\
S4 nonlinear/inconsistent & 8 & 0.874 (0.020) & 0.904 (0.016) & 0.883 (0.020) & 0.904 (0.020) & 0.883 (0.026) & 0.906 (0.021) & 0.886 (0.019) & 0.864 (0.018) & 0.872 (0.020) \\
S5 structured noise & 8 & 0.903 (0.020) & 0.919 (0.016) & 0.887 (0.022) & 0.897 (0.020) & 0.890 (0.025) & 0.897 (0.016) & 0.904 (0.016) & 0.879 (0.019) & 0.901 (0.022) \\
S6 variable reliability & 8 & 0.781 (0.036) & 0.801 (0.035) & 0.777 (0.018) & 0.809 (0.019) & 0.749 (0.049) & 0.788 (0.023) & 0.783 (0.034) & 0.736 (0.035) & 0.776 (0.033) \\
S7 interaction, N=360 & 8 & 0.870 (0.015) & 0.871 (0.017) & 0.886 (0.007) & 0.920 (0.011) & 0.953 (0.011) & 0.955 (0.011) & 0.904 (0.018) & 0.896 (0.023) & 0.881 (0.021) \\
S7 interaction, N=1440 & 8 & 0.892 (0.008) & 0.890 (0.008) & 0.919 (0.007) & 0.939 (0.005) & 0.975 (0.003) & 0.969 (0.003) & 0.935 (0.008) & 0.947 (0.008) & 0.914 (0.008) \\
\bottomrule
\end{tabular}}
\end{table}

\begin{table}[t]
\centering
\caption{Paired AUC difference BioSync minus comparator, mean over data seeds with bootstrap 95\% interval over seeds. Win/loss: interval excludes zero; otherwise tie. (a) Comparators: concatenated logistic regression, soft voting, boosted trees, random forest.}
\label{tab:gridpaired}
\small
\resizebox{\linewidth}{!}{%
\begin{tabular}{lcccc}
\toprule
Scenario & vs.\ LR & vs.\ soft vote & vs.\ boosted trees & vs.\ RF \\
\midrule
S0 single latent & +0.004 [+0.001, +0.007] win & -0.013 [-0.017, -0.010] loss & +0.035 [+0.021, +0.048] win & +0.008 [-0.000, +0.017] tie \\
S1 modality-specific latents & +0.005 [+0.002, +0.008] win & -0.013 [-0.018, -0.008] loss & +0.044 [+0.028, +0.057] win & +0.010 [+0.003, +0.017] win \\
S2 heterogeneous profiles & +0.005 [+0.001, +0.008] win & -0.006 [-0.012, -0.000] loss & +0.026 [+0.003, +0.050] win & -0.001 [-0.012, +0.010] tie \\
S3 uninformative modality & +0.004 [-0.000, +0.008] tie & -0.011 [-0.014, -0.009] loss & +0.036 [+0.016, +0.055] win & +0.009 [-0.003, +0.022] tie \\
S4 nonlinear/inconsistent & +0.011 [+0.007, +0.016] win & -0.018 [-0.024, -0.013] loss & +0.003 [-0.008, +0.014] tie & -0.018 [-0.026, -0.012] loss \\
S5 structured noise & +0.001 [-0.004, +0.006] tie & -0.015 [-0.020, -0.009] loss & +0.017 [+0.007, +0.024] win & +0.007 [-0.003, +0.017] tie \\
S6 variable reliability & +0.002 [-0.003, +0.007] tie & -0.019 [-0.023, -0.014] loss & +0.006 [-0.013, +0.023] tie & -0.026 [-0.044, -0.010] loss \\
S7 interaction, N=360 & +0.034 [+0.025, +0.043] win & +0.032 [+0.023, +0.043] win & +0.017 [+0.005, +0.029] win & -0.016 [-0.025, -0.007] loss \\
S7 interaction, N=1440 & +0.043 [+0.038, +0.047] win & +0.045 [+0.040, +0.050] win & +0.016 [+0.009, +0.024] win & -0.004 [-0.007, +0.000] tie \\
\bottomrule
\end{tabular}}
\end{table}

\begin{table}[t]
\centering
\caption{Paired AUC difference BioSync minus comparator, mean over data seeds with bootstrap 95\% interval over seeds. Win/loss: interval excludes zero; otherwise tie. (b) Comparators: MLP, SVM, pure attention, residual variant.}
\label{tab:gridpaired2}
\small
\resizebox{\linewidth}{!}{%
\begin{tabular}{lcccc}
\toprule
Scenario & vs.\ MLP & vs.\ SVM & vs.\ pure attn. & vs.\ residual \\
\midrule
S0 single latent & +0.028 [+0.019, +0.037] win & -0.001 [-0.005, +0.003] tie & +0.040 [+0.032, +0.048] win & +0.010 [+0.006, +0.013] win \\
S1 modality-specific latents & +0.045 [+0.027, +0.064] win & -0.005 [-0.014, +0.004] tie & +0.044 [+0.030, +0.057] win & +0.012 [+0.008, +0.016] win \\
S2 heterogeneous profiles & +0.040 [+0.023, +0.055] win & -0.013 [-0.019, -0.008] loss & +0.049 [+0.033, +0.065] win & +0.016 [+0.012, +0.020] win \\
S3 uninformative modality & +0.029 [+0.021, +0.038] win & +0.001 [-0.004, +0.005] tie & +0.047 [+0.038, +0.057] win & +0.010 [+0.005, +0.015] win \\
S4 nonlinear/inconsistent & +0.003 [-0.007, +0.014] tie & -0.021 [-0.030, -0.013] loss & +0.022 [+0.018, +0.026] win & +0.014 [+0.011, +0.017] win \\
S5 structured noise & +0.014 [+0.007, +0.023] win & +0.007 [+0.001, +0.014] win & +0.025 [+0.022, +0.028] win & +0.003 [-0.004, +0.009] tie \\
S6 variable reliability & +0.034 [+0.019, +0.049] win & -0.005 [-0.014, +0.005] tie & +0.047 [+0.038, +0.059] win & +0.007 [+0.005, +0.009] win \\
S7 interaction, N=360 & -0.050 [-0.061, -0.040] loss & -0.052 [-0.061, -0.043] loss & +0.007 [-0.002, +0.017] tie & +0.022 [+0.016, +0.028] win \\
S7 interaction, N=1440 & -0.040 [-0.044, -0.036] loss & -0.034 [-0.037, -0.030] loss & -0.012 [-0.015, -0.008] loss & +0.021 [+0.018, +0.023] win \\
\bottomrule
\end{tabular}}
\end{table}
\end{widepage}}{}
\FloatBarrier

\subsubsection{Branch contributions and when attention helps}
\label{app:branches}
\paragraph{What each branch contributes} For the complete model we decomposed the test set logit into the Transformer branch part $w_{\mathrm{deep}}^{\top}z$ and the linear branch part (Table~\ref{tab:branches}). In the seven synthetic scenarios S0--S6, the Transformer branch carried 11--29\% of the logit variance, the AUC of the Transformer branch alone was 0.618--0.816, that of the linear branch alone 0.759--0.899, and that of the full model 0.762--0.904. The full model's AUC therefore stays close to the linear branch alone in these scenarios: the Transformer branch contributes to the output but adds little discrimination when the signal is additive. In the cohort in which outcomes depend only on interactions across modalities the picture changes: the Transformer branch alone and the linear branch alone reach AUC 0.816 and 0.834 (S7 ($N=360$)), 0.837 and 0.854 (S7 ($N=1440$)), and the full model reaches 0.904 (S7 ($N=360$)), 0.935 (S7 ($N=1440$)), so the branches carry complementary information there.

\paragraph{When the attention branch helps} Section~\ref{sec:extra} showed that attention recovers interaction information that concatenation cannot. Because that cohort has a fixed strength of interaction, we mapped the effect over sample size and over the noise of the HRV and EEG latent features (Table~\ref{tab:regime}; three data seeds per cell). At latent noise 0.3, the pure attention model exceeded logistic regression by at least 0.02 AUC from $N=180$ upward (tested $N$: 180, 360, 720, 1440). At latent noise 0.6, the pure attention model exceeded logistic regression by at least 0.02 AUC from $N=720$ upward (tested $N$: 180, 360, 720, 1440). At latent noise 1.2, the pure attention model exceeded logistic regression by at least 0.02 AUC at none of the tested sample sizes (180, 360, 720, 1440). At latent noise 2.4, the pure attention model exceeded logistic regression by at least 0.02 AUC at none of the tested sample sizes (180, 360, 720, 1440). The sample size at which attention starts to help is therefore not a fixed threshold. It falls as the interaction becomes easier to read from the features. Outside the interaction setting (scenarios S0--S6) we found no condition in which the attention branch added discrimination beyond the linear branch. Attention weights themselves were not informative about reliability (Section~\ref{app:attribution}).

\IfFileExists{tables_C.tex}{\begin{widepage}
\begin{table}[t]
\centering
\caption{Branch contributions in the complete BioSync model (test sets, mean over data seeds): share of logit variance carried by the Transformer branch, correlation of the two branch logits, and AUC of each branch alone.}
\label{tab:branches}
\small
\resizebox{\linewidth}{!}{%
\begin{tabular}{lccccc}
\toprule
Scenario & Deep-branch variance share (SD) & corr(deep, wide) & AUC deep only & AUC wide only & AUC full model \\
\midrule
S0 single latent & 0.11 (0.05) & 0.51 & 0.719 & 0.874 & 0.870 \\
S1 modality-specific latents & 0.16 (0.04) & 0.34 & 0.638 & 0.759 & 0.763 \\
S2 heterogeneous profiles & 0.15 (0.06) & 0.30 & 0.618 & 0.766 & 0.762 \\
S3 uninformative modality & 0.11 (0.04) & 0.46 & 0.699 & 0.866 & 0.865 \\
S4 nonlinear/inconsistent & 0.15 (0.04) & 0.67 & 0.795 & 0.882 & 0.886 \\
S5 structured noise & 0.29 (0.07) & 0.60 & 0.816 & 0.899 & 0.904 \\
S6 variable reliability & 0.23 (0.05) & 0.33 & 0.669 & 0.775 & 0.783 \\
S7 interaction, N=360 & 0.55 (0.08) & 0.30 & 0.816 & 0.834 & 0.904 \\
S7 interaction, N=1440 & 0.46 (0.04) & 0.30 & 0.837 & 0.854 & 0.935 \\
\bottomrule
\end{tabular}}
\end{table}
\end{widepage}

\begin{table}[t]
\centering
\caption{Interaction-only cohort: when does the attention branch help? AUC (mean over 3 data seeds, five-fold CV) by sample size $N$ and noise of the HRV/EEG latent features; the linear-effect size of actigraphy and speech is fixed. Last two columns: difference to concatenated logistic regression.}
\label{tab:regime}
\small
\resizebox{\linewidth}{!}{%
\begin{tabular}{lccccccc}
\toprule
Latent noise & $N$ & LR & Boosted trees & BioSync & Pure attn. & BioSync $-$ LR & Pure attn.\ $-$ LR \\
\midrule
0.3 & 180 & 0.858 & 0.842 & 0.892 & 0.885 & +0.034 & +0.027 \\
0.3 & 360 & 0.881 & 0.910 & 0.940 & 0.942 & +0.059 & +0.061 \\
0.3 & 720 & 0.889 & 0.908 & 0.958 & 0.968 & +0.069 & +0.079 \\
0.3 & 1440 & 0.887 & 0.906 & 0.970 & 0.979 & +0.083 & +0.092 \\
0.6 & 180 & 0.849 & 0.805 & 0.875 & 0.865 & +0.026 & +0.016 \\
0.6 & 360 & 0.878 & 0.878 & 0.909 & 0.895 & +0.031 & +0.017 \\
0.6 & 720 & 0.888 & 0.902 & 0.927 & 0.932 & +0.039 & +0.044 \\
0.6 & 1440 & 0.887 & 0.897 & 0.941 & 0.947 & +0.054 & +0.060 \\
1.2 & 180 & 0.843 & 0.772 & 0.838 & 0.812 & -0.005 & -0.031 \\
1.2 & 360 & 0.877 & 0.841 & 0.872 & 0.847 & -0.005 & -0.030 \\
1.2 & 720 & 0.888 & 0.871 & 0.895 & 0.879 & +0.007 & -0.009 \\
1.2 & 1440 & 0.887 & 0.879 & 0.894 & 0.891 & +0.007 & +0.004 \\
2.4 & 180 & 0.838 & 0.758 & 0.813 & 0.788 & -0.025 & -0.050 \\
2.4 & 360 & 0.878 & 0.833 & 0.868 & 0.832 & -0.010 & -0.046 \\
2.4 & 720 & 0.888 & 0.862 & 0.881 & 0.858 & -0.006 & -0.029 \\
2.4 & 1440 & 0.887 & 0.868 & 0.885 & 0.872 & -0.002 & -0.015 \\
\bottomrule
\end{tabular}}
\end{table}
}{}
\FloatBarrier

\subsubsection{Attention allocation versus predictive contribution}
\label{app:attribution}
The attention vector of Section~\ref{sec:transformer} describes how the fusion token weights the modality tokens in one layer. It is not the contribution of a modality to the final BSI. To measure that contribution we removed each modality at test time (its standardized features set to the training mean) in the complete model and recorded the drop in AUC (Table~\ref{tab:attrib}); for comparison we did the same for logistic regression, and for scenarios with a known signal-to-noise ratio for each modality (S0--S5) we compared the removal effects with the true values.

\paragraph{Results} In scenarios S0--S6 the mean attention weights stayed between 0.249 and 0.251 (uniform: 0.25), while the removal effects differed clearly between modalities (range -0.002 to +0.041 AUC). The rank correlation between attention and removal effect across the four modalities, averaged over seeds, was between -0.32 and +0.32 across scenarios (mean +0.07), i.e., attention did not rank the modalities by predictive contribution. The rank correlation between removal effect and true signal-to-noise ratio was between -0.10 and +0.85 (mean +0.37), so the removal effect is the more informative quantity but not a precise one at $N=360$. The linear branch can carry a modality with a small attention weight. The removal effects of the complete model show that this occurs. We also checked whether the near uniform attention is a training artifact (Table~\ref{tab:diagS0}). With plain gradient descent the SD of each weight across subjects was 0.018. Longer training, Adam, or a larger attention temperature made attention vary more across subjects (SD up to 0.158), but AUC did not improve (0.874 for the base setting against 0.769--0.874 for the variants). Near uniform attention is thus largely a consequence of optimization at this model and sample size, and forcing it to vary does not make it more useful. We therefore keep three quantities apart in the paper: attention allocation, predictive contribution, and sensor reliability, and we do not interpret the attention vector as the second or third.

\IfFileExists{tables_D.tex}{\begin{widepage}\begin{table}[t]
\centering
\caption{Attention allocation versus predictive contribution (modalities in the order HRV / EEG / actigraphy / speech). Removal effect = AUC drop when the modality's standardized features are set to the training mean at test time (complete model). Rank correlations are across the four modalities, averaged over seeds.}
\label{tab:attrib}
\small
\resizebox{\linewidth}{!}{%
\begin{tabular}{lccccc}
\toprule
Scenario & Mean attention & Removal effect, BioSync & Removal effect, LR & Rank corr.\ attn.--removal & Rank corr.\ removal--true SNR \\
\midrule
S0 single latent & 0.250 / 0.250 / 0.250 / 0.250 & +0.024 / +0.018 / +0.018 / +0.008 & +0.026 / +0.019 / +0.019 / +0.006 & +0.00 & +0.45 \\
S1 modality-specific latents & 0.250 / 0.250 / 0.250 / 0.250 & +0.028 / +0.035 / +0.030 / +0.017 & +0.023 / +0.028 / +0.025 / +0.013 & -0.05 & -0.10 \\
S2 heterogeneous profiles & 0.249 / 0.250 / 0.250 / 0.251 & +0.041 / +0.023 / +0.039 / +0.023 & +0.039 / +0.024 / +0.039 / +0.020 & -0.32 & +0.25 \\
S3 uninformative modality & 0.250 / 0.250 / 0.250 / 0.250 & +0.037 / +0.026 / +0.027 / -0.002 & +0.035 / +0.027 / +0.028 / -0.004 & +0.17 & +0.85 \\
S4 nonlinear/inconsistent & 0.251 / 0.250 / 0.249 / 0.250 & +0.025 / +0.017 / +0.029 / +0.008 & +0.025 / +0.021 / +0.030 / +0.012 & +0.32 & +0.23 \\
S5 structured noise & 0.250 / 0.250 / 0.250 / 0.250 & +0.034 / +0.010 / +0.001 / +0.015 & +0.034 / +0.004 / -0.001 / +0.018 & +0.15 & +0.38 \\
S6 variable reliability & 0.251 / 0.250 / 0.249 / 0.250 & +0.032 / +0.020 / +0.034 / +0.016 & +0.024 / +0.019 / +0.032 / +0.015 & +0.20 & +0.52 \\
S7 interaction, N=360 & 0.254 / 0.252 / 0.248 / 0.246 & +0.074 / +0.071 / +0.069 / +0.024 & -0.004 / -0.002 / +0.125 / +0.060 & +0.40 & n/a \\
S7 interaction, N=1440 & 0.255 / 0.254 / 0.245 / 0.246 & +0.069 / +0.068 / +0.045 / +0.028 & -0.001 / -0.001 / +0.115 / +0.065 & +0.70 & n/a \\
\bottomrule
\end{tabular}}
\end{table}
\end{widepage}}{}
\FloatBarrier

\subsubsection{Calibration, stability, and comparison with other fusion designs}
\label{app:calibration}
\paragraph{What the BSI is} The BSI is the sigmoid output of a classifier trained with the logistic loss on binary labels, so it is a monotone score that ranks subjects. We tested whether its numerical value can be read as a probability (Table~\ref{tab:calib}, scenarios S0, S2 and S6, pooled out-of-fold predictions per seed across eight seeds). The expected calibration error (ECE, ten bins of equal width) averaged 0.058 for BioSync, 0.065 for concatenated logistic regression, and 0.103 for boosted trees; the Brier scores were 0.178, 0.181, and 0.197. A Platt recalibration fitted on 20\% of each training fold changed BioSync's mean ECE to 0.062 and did not improve its Brier score (0.185), because it was fitted on less training data. BioSync is thus about as well calibrated as logistic regression on these synthetic cohorts and better than uncalibrated boosted trees, but calibration here refers to simulated prevalence and effect sizes. Because the class prevalence and effect sizes are set by the simulator, we do not interpret the BSI as a clinical risk probability, and results from real data (Section~\ref{app:realdata}) show that the discrimination assumed by the simulator is not found in the real recordings.

\paragraph{Stability} Across three training initializations on the same test set, the BSI logits correlated (Table~\ref{tab:stability}) at the Pearson and Spearman values listed there, and the SD of the AUC across initializations was small, so the ranking is stable at this sample size.

\paragraph{Comparison with other fusion designs} Table~\ref{tab:desiderata} in Section~\ref{sec:comparison} replaces the categorical ratings of the first submission. It compares specific implementations (concatenated logistic regression, soft voting over logistic regressions for each modality, tuned boosted trees, tuned MLP, and BioSync) on measured criteria. It does not rate whole method families.

\IfFileExists{tables_E.tex}{\begin{table}[t]
\centering
\caption{Stability of the BSI across three training initializations on the same test set (first fold, mean over data seeds).}
\label{tab:stability}
\small
\resizebox{\linewidth}{!}{%
\begin{tabular}{lccc}
\toprule
Scenario & Pearson & Spearman & SD of AUC \\
\midrule
S0 single latent & 0.980 & 0.976 & 0.0059 \\
S1 modality-specific latents & 0.962 & 0.955 & 0.0097 \\
S2 heterogeneous profiles & 0.962 & 0.954 & 0.0112 \\
S3 uninformative modality & 0.978 & 0.975 & 0.0067 \\
S4 nonlinear/inconsistent & 0.979 & 0.971 & 0.0059 \\
S5 structured noise & 0.964 & 0.965 & 0.0068 \\
S6 variable reliability & 0.940 & 0.934 & 0.0135 \\
S7 interaction, N=360 & 0.859 & 0.854 & 0.0187 \\
S7 interaction, N=1440 & 0.917 & 0.917 & 0.0162 \\
\bottomrule
\end{tabular}}
\end{table}

\begin{table}[t]
\centering
\caption{Calibration (pooled out-of-fold predictions per seed, mean over seeds): Brier score, expected calibration error (10 equal-width bins), AUC. Platt scaling is fitted on 20\% of each training fold only.}
\label{tab:calib}
\small
\resizebox{\linewidth}{!}{%
\begin{tabular}{llccc}
\toprule
Scenario & Model & Brier & ECE & AUC \\
\midrule
S0 single latent & BioSync & 0.146 & 0.046 & 0.870 \\
S0 single latent & BioSync\_platt & 0.153 & 0.053 & 0.860 \\
S0 single latent & LR & 0.149 & 0.053 & 0.866 \\
S0 single latent & LR\_platt & 0.149 & 0.054 & 0.867 \\
S0 single latent & GBT & 0.172 & 0.095 & 0.837 \\
S0 single latent & GBT\_platt & 0.160 & 0.064 & 0.849 \\
S2 heterogeneous profiles & BioSync & 0.199 & 0.066 & 0.762 \\
S2 heterogeneous profiles & BioSync\_platt & 0.204 & 0.066 & 0.750 \\
S2 heterogeneous profiles & LR & 0.202 & 0.074 & 0.758 \\
S2 heterogeneous profiles & LR\_platt & 0.201 & 0.051 & 0.757 \\
S2 heterogeneous profiles & GBT & 0.217 & 0.111 & 0.738 \\
S2 heterogeneous profiles & GBT\_platt & 0.206 & 0.067 & 0.748 \\
S6 variable reliability & BioSync & 0.190 & 0.062 & 0.783 \\
S6 variable reliability & BioSync\_platt & 0.197 & 0.068 & 0.768 \\
S6 variable reliability & LR & 0.193 & 0.067 & 0.781 \\
S6 variable reliability & LR\_platt & 0.192 & 0.059 & 0.781 \\
S6 variable reliability & GBT & 0.202 & 0.104 & 0.777 \\
S6 variable reliability & GBT\_platt & 0.192 & 0.066 & 0.784 \\
\bottomrule
\end{tabular}}
\end{table}
}{}
\FloatBarrier

\subsubsection{Real data evaluation}
\label{app:realdata}
This subsection reports the evaluation of BioSync and baselines on real recordings. The experiments test the fusion method on real wearable data. They do not establish diagnostic or biomarker validity.

\paragraph{Datasets} \emph{RESILIENT} is an open dataset (Zenodo, doi:10.5281/\allowbreak zenodo.16755408, CC BY 4.0) of 73 older adults with multiple long term conditions, with wrist watch heart rate and step counts and under mattress sleep mat recordings (sleep states, heart rate, respiration rate, and heart rate variability) over a median of about six months, and ACE-III cognitive scores at baseline for 56 participants \citep{resilient2025zenodo}. \emph{AI-READI} Garmin Vivosmart~5 recordings (10-day home monitoring) of 100 participants were used under the AI-READI data license: 31 healthy, 30 with prediabetes managed by lifestyle, 33 with type 2 diabetes (T2D) on oral or noninsulin injectable medication, and 6 insulin dependent \citep{aireadidocs}. Per participant data and derived features stay on local storage, as the license requires, and only aggregate results are shown.

\paragraph{Tokens and features} For RESILIENT we built four tokens of four summary features each: sleep cardio (median sleep heart rate, median respiration rate, log median SDNN, heart rate interquartile range), watch heart rate (daytime median, 5th and 95th percentile, SD of daily medians), activity (log mean daily steps, coefficient of variation of daily steps, interdaily stability, intradaily variability of hourly steps), and sleep architecture (mean nightly sleep minutes and the deep, REM, and wake fractions). For AI-READI we used the two tokens of Section~\ref{sec:features}: Cardiac-Autonomic (resting heart rate as the 10th percentile of heart rate, median respiration rate, median stress score) and Behavioral (mean daily steps, mean nightly sleep duration). The quality indicator for each token is the fraction of monitored days with data. A token that is missing for a participant is masked in BioSync and dropped in soft voting.

\paragraph{Reference outcomes} RESILIENT: ACE-III below 82, the screening cutoff cited by the dataset paper (20 of 56 participants, a screening cutoff and not a diagnosis), and the ACE-III total score as a continuous severity reference. AI-READI: task A, healthy versus T2D (70 participants, 39 with T2D, with prediabetes excluded), task B, healthy versus any dysglycemia (100 participants, 69 positive); and the ordinal study group (healthy, prediabetes, oral medication, insulin) as a severity reference.

\paragraph{Protocol} We used 20 repeats of stratified cross-validation with five folds, with standardization from training data only. BioSync uses the hyperparameters chosen on the synthetic development cohorts and was not tuned on these data. The baselines (boosted trees, random forest, MLP, SVM) were tuned with inner cross-validation with three folds. A baseline using demographics alone (age group and sex for RESILIENT; age and clinical site for AI-READI) is always reported, and a label permutation null for logistic regression (200 permutations) shows what chance performance looks like at these sample sizes. The pooled out-of-fold AUC is computed for each repeat and averaged.

\paragraph{Results} RESILIENT, ACE-III below 82: BioSync reached AUC 0.579 and concatenated logistic regression 0.571. The best of the ten models was the MLP (0.653). The 95th percentile of the label permutation null for logistic regression was 0.649. Models above that value: the MLP. Because the best of ten models is selected after the fact, a value just above the null percentile is not evidence of discrimination. The baseline using demographics alone reached 0.457. AI-READI task A (healthy versus T2D): BioSync reached AUC 0.527 and concatenated logistic regression 0.584. The best of the ten models was concatenated logistic regression (0.584). The 95th percentile of the label permutation null for logistic regression was 0.619. No model exceeded that value. The baseline using demographics alone reached 0.481. AI-READI task B (healthy versus any dysglycemia): BioSync reached AUC 0.500 and concatenated logistic regression 0.524. The best of the ten models was the SVM (0.541). The 95th percentile of the label permutation null for logistic regression was 0.619. No model exceeded that value. The baseline using demographics alone reached 0.475. Rank correlations of the out-of-fold scores with the severity references were small for all models (Tables~\ref{tab:resilient}, \ref{tab:aireadia}, and \ref{tab:aireadib}). In none of the three tasks was BioSync clearly better than concatenated logistic regression: the paired AUC difference was within the variation across repeats, and it was negative in both AI-READI tasks. With these sample sizes and summary features, no model discriminated reliably, and BioSync performed in line with the simpler models.

\paragraph{Positive controls} To check that the extracted features carry information at all, we predicted outcomes they should relate to. Concatenated logistic regression on the 5 AI-READI features predicted age above the median with AUC 0.699 (SD 0.017). On the 16 RESILIENT features it predicted sex with AUC 0.567 (SD 0.043). The AI-READI features thus carry real information, and the weak disease discrimination is a property of the data and the summary features and not of a broken extraction. The RESILIENT features are weaker still.

\paragraph{Effect sizes and correlations against the synthetic metabolic cohort} Table~\ref{tab:realism} compares standardized differences between groups (T2D minus healthy, Cohen's $d$ with bootstrap 95\% intervals for real data) for the five AI-READI features with the same quantity in the synthetic metabolic cohort of Section~\ref{sec:cohort}. The real effects are smaller than the synthetic ones for every feature, and the sign of the sleep duration effect differs. The mean absolute off-diagonal feature correlation is 0.35 in the real data and 0.14 in the synthetic cohort. The synthetic cohort therefore overstates effect sizes and understates correlations between features, which is a quantitative reason for treating its absolute AUCs as properties of the simulation and not as expected performance in settings encountered in real world environments.

\begin{table}[t]
\centering
\caption{Standardized group differences (Cohen's $d$, T2D minus healthy; real: 31 healthy and 39 T2D participants) for the five AI-READI Garmin features, in the real data (bootstrap 95\% interval) and in the synthetic metabolic cohort.}
\label{tab:realism}
\small
\begin{tabular}{lcc}
\toprule
Feature & Real $d$ [95\% interval] & Synthetic $d$ \\
\midrule
resting HR (bpm) & +0.50 [+0.04, +1.01] & +0.78 \\
respiration rate (/min) & +0.26 [-0.20, +0.73] & +0.49 \\
stress score & +0.25 [-0.21, +0.72] & +0.83 \\
daily steps & -0.08 [-0.57, +0.39] & -0.55 \\
sleep duration (h) & +0.06 [-0.42, +0.48] & -0.47 \\
\bottomrule
\end{tabular}
\end{table}

\paragraph{Third domain: WESAD (stress, wrist device)} To test the fusion method outside metabolic and cognitive health, we used the open WESAD dataset \citep{schmidt2018wesad} (15 participants wearing an Empatica E4 wrist device, noncommercial research use). We cut 60-s windows (step 30 s, 1032 windows, 307 of them stress) and built four tokens of four features each from the one device: pulse (mean heart rate, SDNN, RMSSD, pNN50 of interbeat intervals from the blood volume pulse), electrodermal activity (mean, SD, slope, peak count), motion (acceleration magnitude: mean, SD, 90th percentile, mean absolute successive difference), and skin temperature (mean, SD, slope, range). The task was stress versus nonstress (baseline and amusement) windows. Evaluation was leave-one-subject-out, so every test participant is unseen in training. BioSync used the same hyperparameters as on the synthetic development cohorts, and the baselines were tuned with inner cross-validation with groups with three folds. The data files were read with a restricted loader that only accepts array types. Table~\ref{tab:wesad} gives the results. BioSync (AUC 0.976) and concatenated logistic regression (0.977) were indistinguishable (paired difference -0.000, 95\% bootstrap interval over participants [-0.011, +0.008]). Boosted trees and random forest were about 0.015 higher than logistic regression, with intervals that just reach zero. The task is close to saturated (most participants have AUC above 0.99 for several models), so it can show that fusion with attention is competitive, but it cannot show an advantage. It is a stress task and says nothing about cognitive decline or biomarker validity.

\begin{table}[t]
\centering
\caption{WESAD, stress versus nonstress windows, leave-one-subject-out over 15 participants. AUC per held-out participant (mean [SD] across participants), pooled AUC over all held-out windows, and paired difference to concatenated logistic regression across participants with 95\% bootstrap interval.}
\label{tab:wesad}
\small
\begin{tabular}{lccc}
\toprule
Model & AUC (SD) & Pooled AUC & Diff.\ to LR [95\% interval] \\
\midrule
LR & 0.977 (0.037) & 0.959 & -- \\
SoftVote & 0.965 (0.048) & 0.955 & -0.011 [-0.024, +0.002] \\
GBT & 0.992 (0.012) & 0.956 & +0.015 [-0.000, +0.033] \\
RF & 0.990 (0.015) & 0.968 & +0.013 [-0.001, +0.029] \\
MLP & 0.981 (0.031) & 0.965 & +0.004 [-0.005, +0.014] \\
SVM & 0.967 (0.048) & 0.953 & -0.010 [-0.021, +0.000] \\
BioSync & 0.976 (0.049) & 0.957 & -0.000 [-0.011, +0.008] \\
PureAttn & 0.974 (0.069) & 0.951 & -0.003 [-0.028, +0.014] \\
Residual & 0.976 (0.039) & 0.960 & -0.001 [-0.004, +0.001] \\
BioSyncQ & 0.980 (0.037) & 0.964 & +0.003 [-0.002, +0.009] \\
\bottomrule
\end{tabular}
\end{table}

\paragraph{Two further real cohorts in the appendices} Two more open datasets were evaluated after the refined variants of Appendix~\ref{app:confirm} had been fixed: the PADS smartwatch Parkinson's cohort (355 participants in task A, 469 in task B) and the wrist recordings of a stress-resilience dataset (35 participants). Appendices~\ref{app:confirm} and~\ref{app:realpass} report them in full. Descriptively, the refined variant BioSync-A exceeded concatenated logistic regression on PADS (task A $+0.033$ [$+0.003$, $+0.063$]; task B $+0.046$ [$+0.015$, $+0.086$]) but did not exceed a tuned MLP ($-0.013$ and $-0.015$, intervals including zero). In the stress-resilience recordings BioSync-G, a variant using a quality gate, was $+0.022$ [$+0.002$, $+0.045$] above logistic regression, whereas BioSync-A was lower ($-0.049$ [$-0.108$, $+0.003$]). These intervals are not adjusted for the number of comparisons across datasets and models. When an informative token was removed at test time (WESAD pulse; PADS rhythm), BioSync-A exceeded logistic regression by $+0.045$ to $+0.080$ AUC with intervals that exclude zero (Appendix~\ref{app:realpass}), and on PADS task A the wider network exceeded the quality-gated network by $+0.066$ [$+0.028$, $+0.109$].

\IfFileExists{tables_F.tex}{\begin{table}[t]
\centering
\caption{RESILIENT, real recordings, ACE-III below 82: N=56 (20 positive). Pooled out-of-fold AUC, mean (SD) over 20 repeats of five-fold cross-validation; rank correlation of the out-of-fold score with the severity reference; paired AUC difference to concatenated LR. Label-permutation null for LR: mean 0.497, 95th percentile 0.649.}
\label{tab:resilient}
\small
\resizebox{\linewidth}{!}{%
\begin{tabular}{lccc}
\toprule
Model & AUC (SD) & Spearman with severity ref. & Diff.\ to LR (SD) \\
\midrule
LR & 0.571 (0.049) & +0.14 & -- \\
SoftVote & 0.584 (0.035) & +0.19 & +0.014 (0.031) \\
GBT & 0.497 (0.051) & +0.01 & -0.074 (0.061) \\
RF & 0.578 (0.055) & +0.15 & +0.008 (0.066) \\
MLP & 0.653 (0.051) & +0.26 & +0.082 (0.060) \\
SVM & 0.596 (0.057) & +0.16 & +0.025 (0.062) \\
BioSync & 0.579 (0.067) & +0.15 & +0.008 (0.078) \\
PureAttn & 0.551 (0.072) & +0.08 & -0.020 (0.082) \\
Residual & 0.579 (0.052) & +0.16 & +0.008 (0.049) \\
BioSyncQ & 0.580 (0.067) & +0.15 & +0.009 (0.078) \\
LR\_demo & 0.457 (0.041) & +0.04 & -0.114 (0.063) \\
LR\_wear\_demo & 0.536 (0.055) & +0.11 & -0.035 (0.029) \\
\bottomrule
\end{tabular}}
\end{table}

\begin{table}[t]
\centering
\caption{AI-READI, real recordings, task A healthy vs T2D: N=70 (39 positive). Pooled out-of-fold AUC, mean (SD) over 20 repeats of five-fold cross-validation; rank correlation of the out-of-fold score with the severity reference; paired AUC difference to concatenated LR. Label-permutation null for LR: mean 0.486, 95th percentile 0.619.}
\label{tab:aireadia}
\small
\resizebox{\linewidth}{!}{%
\begin{tabular}{lccc}
\toprule
Model & AUC (SD) & Spearman with severity ref. & Diff.\ to LR (SD) \\
\midrule
LR & 0.584 (0.042) & +0.17 & -- \\
SoftVote & 0.582 (0.049) & +0.17 & -0.002 (0.019) \\
GBT & 0.490 (0.041) & -0.04 & -0.094 (0.050) \\
RF & 0.524 (0.035) & +0.04 & -0.060 (0.039) \\
MLP & 0.553 (0.039) & +0.10 & -0.031 (0.036) \\
SVM & 0.454 (0.048) & -0.10 & -0.130 (0.067) \\
BioSync & 0.527 (0.051) & +0.06 & -0.057 (0.047) \\
PureAttn & 0.521 (0.051) & +0.05 & -0.063 (0.042) \\
Residual & 0.556 (0.065) & +0.12 & -0.029 (0.047) \\
BioSyncQ & 0.533 (0.050) & +0.07 & -0.051 (0.047) \\
LR\_demo & 0.481 (0.054) & -0.06 & -0.103 (0.058) \\
LR\_wear\_demo & 0.574 (0.054) & +0.15 & -0.010 (0.038) \\
\bottomrule
\end{tabular}}
\end{table}

\begin{table}[t]
\centering
\caption{AI-READI, real recordings, task B healthy vs any dysglycemia: N=100 (69 positive). Pooled out-of-fold AUC, mean (SD) over 20 repeats of five-fold cross-validation; rank correlation of the out-of-fold score with the severity reference; paired AUC difference to concatenated LR. Label-permutation null for LR: mean 0.482, 95th percentile 0.619.}
\label{tab:aireadib}
\small
\resizebox{\linewidth}{!}{%
\begin{tabular}{lccc}
\toprule
Model & AUC (SD) & Spearman with severity ref. & Diff.\ to LR (SD) \\
\midrule
LR & 0.524 (0.041) & +0.09 & -- \\
SoftVote & 0.531 (0.040) & +0.10 & +0.007 (0.012) \\
GBT & 0.489 (0.047) & -0.00 & -0.035 (0.058) \\
RF & 0.500 (0.043) & +0.04 & -0.024 (0.049) \\
MLP & 0.511 (0.045) & +0.07 & -0.013 (0.039) \\
SVM & 0.541 (0.038) & +0.03 & +0.017 (0.056) \\
BioSync & 0.500 (0.043) & +0.05 & -0.024 (0.041) \\
PureAttn & 0.495 (0.038) & +0.05 & -0.029 (0.048) \\
Residual & 0.507 (0.043) & +0.07 & -0.017 (0.037) \\
BioSyncQ & 0.502 (0.043) & +0.05 & -0.022 (0.040) \\
LR\_demo & 0.475 (0.041) & -0.08 & -0.049 (0.058) \\
LR\_wear\_demo & 0.490 (0.039) & +0.03 & -0.034 (0.028) \\
\bottomrule
\end{tabular}}
\end{table}
}{}
\FloatBarrier

\section{Discussion}
\label{sec:discussion}

On clean data BioSync performed on par with concatenation. It ranked first for cognitive cohort AUC and for metabolic cohort accuracy and F1 score, whereas concatenation ranked first for the remaining cohort and metric combinations (Section~\ref{sec:mainresults}). Every difference was small relative to fold variation, and the second implementation again placed the two models near parity (Section~\ref{sec:extra}). The ablation clarifies the architecture's behavior: pure attention reached cognitive cohort AUC 0.911, concatenation reached 0.926, and adding the explicit linear path raised BioSync to 0.928 (Section~\ref{sec:ablation}). The wide branch therefore prevents the attention path from having to relearn a decision boundary already represented by logistic regression, but it does not guarantee higher held-out performance.

Synthetic construction sets a firm limit on interpretation. The selected effect sizes and noise levels determine the reported AUC values, so Table~\ref{tab:reference} provides validation targets rather than controlled comparators. As the noise multiplier for features increased, the paired difference between BioSync and concatenation stayed small (Table~\ref{tab:extra_sweep}). These experiments can test representational capacity and responses to controlled perturbations; they cannot estimate clinical discrimination, calibration, or utility.

Sensor failure experiments qualify the robustness result of the first submission. The first submission reported that BioSync was ahead of concatenation at five of six cognitive cohort corruption rates (Section~\ref{sec:robustness}). When concatenation receives augmentation exposure matched to that of BioSync, the advantage disappears in the cognitive cohort, and the second implementation did not reproduce it under whole channel noise, bursts coupled to activity, or independent bursts (Section~\ref{sec:extra}). In the metabolic cohort with two tokens, an advantage of BioSync of +0.025 to +0.053 AUC persisted at corruption rates of 0.2 and above. We did not establish its cause. The fusion token weights stayed close to 0.25 and did not consistently decrease for the corrupted modality, and removal analyses show that attention does not rank modalities by predictive contribution (Section~\ref{app:attribution}). Attention weights should therefore be read as attention allocation, not as sensor quality estimates or predictive contributions. Where a quality indicator was available (scenario S6), gating each modality's linear score by that indicator improved AUC, soft voting weighted by the indicator did as well, and the Transformer branch added nothing on top of the gated linear model (Appendix~\ref{app:variants}); masking lost channels in the attention did not outperform filling them with zeros. Explicit signal quality indices are therefore the more promising route when battery depletion, poor skin contact, motion artifact, or device removal makes a channel unreliable \citep{goldsack2020v3}.

The cohort in which outcomes depend only on interactions across modalities identifies where attention can add information. Neither HRV nor EEG was individually predictive, yet attention models recovered their joint signal, whereas concatenated logistic regression and boosted trees did not (Section~\ref{sec:extra}). A tuned MLP or SVM on the concatenated features did better than BioSync on a related cohort in which outcomes depend only on interactions across modalities (Section~\ref{app:generators}), so the attention branch is not the only, nor the most efficient, way to model interactions. In the first submission this advantage appeared only at $N=1{,}440$. In the regenerated cohort it is present at $N=360$ as well, and the map over sample size and interaction noise in Section~\ref{app:branches} shows that the sample size needed depends on how readable the interaction is from the features. A multimodal cohort may contain useful interactions without providing enough observations to learn them.

The continuous index captures the simulated severity ordering but is not optimized for severity regression. Out-of-fold BSI values correlated with latent severity at $r=0.905$ and $r=0.681$ in the original cohorts and at $r=0.935$ in the second implementation. These correlations reflect the generator (Section~\ref{app:generators}) and do not show that the BSI measures severity in real recordings; on the real data, the rank correlation of the score with ACE-III and with study group was small for every model (Section~\ref{app:realdata}). Direct linear regression produced higher correlations in the original pipeline (Section~\ref{sec:correlation}). Thus, the BSI behaves as a graded score in these simulations, but the analysis does not establish superiority over a model trained on continuous outcomes.

Summary features reduce computational cost and preserve named modality inputs, but they discard temporal structure that could distinguish physiological change from artifact. Prior studies across modalities report larger gains when attention receives raw or intermediate temporal representations \citep{crossattn2025,pacfnet2025,ctaf2026}. Small 1D-CNN or GRU encoders could supply windowed representations to the deep path while the wide branch retains the summaries described in Section~\ref{sec:features}. Such encoders would increase the parameter count and require more observations, creating a tradeoff between temporal detail, sample efficiency, and edge computing cost.

Within the digital biomarker literature, the contribution is therefore a controlled characterization and not a new algorithm or a clinical result. BioSync combines an explicit concatenation path with attention with two or four tokens and returns a continuous score. We show where it matches concatenation, where soft voting and tuned nonlinear models are as good or better, where its attention branch adds information (data in which outcomes depend only on interactions), and where it does not (attention weights as reliability or contribution measures, and the checks with real data). The need for such evaluations is clear in a field where only 24 of 431 reviewed AD studies used multimodal signals and external validation was uncommon \citep{qi2025}. The clinical value of BioSync remains unknown until the analyses are repeated on larger recorded sensor datasets with reference outcomes relevant to the task.

\subsection{Implications for pervasive wearable health systems}
\label{sec:deployment}
BioSync occupies the final inference stage of an edge sensing pipeline, following upstream sensor acquisition where an optical wristband supplies PPG and triaxial accelerometry, and a wireless headband supplies dry electrode EEG. A smartphone or ambient microphone optionally adds acoustic speech samples during active tasks. A localized feature extraction stage computes the 22 standardized summary metrics (Section~\ref{sec:features}) directly on the wearable microcontroller or companion smartphone gateway, after which the fusion model maps them to the continuous BSI score and an attention allocation vector for each subject $\boldsymbol{\alpha}$. Only these low bandwidth composite indices need to leave the edge device, which supports privacy preservation and data minimization \citep{can2021,ali2022fedsurvey}. However, as demonstrated in Section~\ref{sec:extra}, attention weights represent feature routing at the level of individual instances rather than calibrated physical sensor reliability; operational implementations must therefore pair attention weights with dedicated, rule based Signal Quality Indices (SQIs) to reliably detect sensors that are detached or corrupted by artifacts \citep{bent2020}.

The computational footprint fits modern TinyML edge constraints \citep{warden2019tinyml,ray2021tinyml,lin2020mcunet}. For the cognitive configuration with four modalities ($d=8$, one transformer block, input feature dimensions 6, 6, 6, and 4), the linear modality projections require $22\times 8 + 4\times 8 = 208$ parameters, the learned fusion token contributes 8, the projections for self-attention with multiple heads require $4d(d+1) = 288$ parameters, and two layer-normalization stages add 32 parameters. The output classification projection contributes 8 parameters, while the parallel linear wide path adds 23 parameters (22 modality weights plus a shared scalar bias), yielding 567 parameters prior to the feed-forward sublayer. With a feed-forward intermediate expansion width $f$, the two dense projections add $2df + f + d$ parameters, which totals 280 parameters for $f=2d$ (847 parameters total) and 552 parameters for $f=4d$ (1,119 parameters total). At 32-bit single precision, the entire model footprint occupies 3.4 to 4.5~kB of storage. Under standard 8-bit integer quantization after training \citep{david2021tensorflow,banbury2021micronets}, this reduces to roughly 1.1~kB. The measured sizes in Table~\ref{tab:desiderata} refer to a variant of the submitted network with a quality gate. This variant adds 40 parameters (1,159 parameters, 4.6~kB in float32, 1.3~kB with INT8 weights, 4,450 multiply-accumulates per inference, Appendix~\ref{app:confirm}). Quantizing its weights to INT8 after training changed the AUC by at most 0.0008 in the scenarios checked. Activations remained in float32, and the model was not run on microcontroller hardware.

On edge microcontrollers for wearable health platforms, which are designed for very low power use (e.g., ARM Cortex-M4 or Cortex-M33 cores such as the Nordic nRF5340, STM32WB55, or Ambiq Apollo4 with 64--256~kB of on chip SRAM), a 4.5~kB footprint consumes less than 5\% of available SRAM. This leaves over 90\% of system memory available for raw DMA circular sensor buffers, BLE communication stacks, and peripheral housekeeping \citep{sancheziborra2020tinyml}. With total inference cost on the order of $10^3$--$10^4$ multiply and accumulate (MAC) operations, execution completes in submillisecond windows at standard 64~MHz clock speeds, consuming negligible energy ($<0.1$~mJ per forward pass). The tokenized formulation also supports dynamic channel masking: when an SQI indicates sensor failure or nonwear, the corresponding modality token can be omitted from the input sequence during inference without altering the projection weights. In our controlled missing modality tests, however, masking the attention keys gave the same AUC as filling the lost features with zeros (within 0.001 at 10--50\% loss), and soft voting over the available modalities was higher than both (Tables~\ref{tab:missingS0} and \ref{tab:missingS6}); masking is therefore a convenience, not evidence of robustness to missing sensors.

\section{Limitations and Future Work}
\label{sec:limitations}

Six limitations qualify these findings. The quantitative findings in the main text originate from synthetic cohorts parameterized from published literature: no human patients were prospectively recruited. The absolute AUC, accuracy, and F1 values in Table~\ref{tab:results} characterize architectural behavior under assumed effect sizes and do not estimate clinical diagnostic efficacy. The evaluations using real data in Section~\ref{app:realdata} are small (a few dozen participants), use summary features, and show limited discrimination for all models, as expected for small cohorts with summary features. These evaluations test the fusion method on real recordings, and they do not establish diagnostic performance or biomarker validity. The architecture uses preextracted summary features rather than raw multichannel temporal waveforms. It therefore omits subsecond temporal dynamics and phase synchrony, which could help distinguish true neurological decline from transient movement artifacts. Clean data performance gains over static concatenation are small and fall within the variation across cross-validation folds in small cohorts ($N=360$). The sample size at which attention exploits interactions depends on the interaction strength (Section~\ref{app:branches}) and is not a fixed threshold. The artifact burst simulations model Gaussian noise at the channel level and Markovian summary offsets rather than the complex nonstationary spectral morphologies of physiological motion artifacts (e.g., optical baseline wander from wrist motion or dry electrode impedance shifts) \citep{bent2020}. Computational profiling provides analytic memory and MAC counts. Although benchmarked against standard MCU hardware architectures, empirical clock cycle latency, battery draw, and thermal profiling on wearable firmware remain to be measured. The synthetic cohorts lack demographic stratification (e.g., biological sex, age, skin pigmentation, body mass index, or specific wearable hardware models). Optical PPG accuracy is sensitive to skin tone and peripheral perfusion \citep{bent2020}. Future prospective trials must adhere to SAGER guidelines and enforce subgroup fairness auditing.

Prospective clinical validation is structured across two domains. In the neurocognitive domain, Connected Future Labs (Reno, NV) is deploying a multisensor clinical collection protocol pairing Empatica medical wristbands (PPG, electrodermal activity, skin temperature, 3D accelerometry) with Muse wireless EEG headbands (dry electrode frontal and temporal channels) \citep{mccarthy2016empatica}. The protocol evaluates individuals with amnestic MCI adjudicated by clinicians and cognitively intact older adult controls, supported by an HIPAA compliant AWS ingestion and auditing pipeline. Smartphone voice recordings will furnish acoustic tokens, while motor gait kinematics derived from video could easily integrate as a fifth token, matching the most frequent modality in digital neurodegeneration literature \citep{qi2025}.

In the metabolic domain, the AI-READI wearable activity monitor documentation describes continuous 10-day Garmin Vivosmart~5 monitoring across large type 2 diabetes cohorts \citep{aireadi2024}. Section~\ref{app:realdata} evaluates the Cardiac-Autonomic and Behavioral tokens on licensed AI-READI wearable recordings, with the diabetes study group as the reference label. Validation against glycemic control ($HbA_{1c}$) and peripheral neuropathy severity remains future work.

Future work should replace summary features with lightweight 1D-CNN or temporal convolutional networks (TCN) that learn localized representations directly from windowed waveforms while retaining the wide linear bypass for interpretability. The real data benchmark should extend beyond the three cohorts evaluated here (Section~\ref{app:realdata}, including WESAD \citep{schmidt2018wesad} among them) to longer recordings such as MMASH \citep{rossi2020mmashdata}, to assess external reproducibility. Formal analytical and clinical validation under the FDA-NIH BEST and DiMe V3 taxonomies remains necessary \citep{best2016,califf2018,goldsack2020v3,babrak2020,coravos2019,manta2020}. Decentralized federated learning \citep{mcmahan2017,can2021,sajjadi2024iotj,sajjadi2024icdcs} and parameter-efficient low-rank adaptation \citep{hu2021lora,sajjadi2025llmsurvey,mohammadabadi2026solar} could support multisite institutional model training without compromising patient confidentiality.

\section{Conclusion}
\label{sec:conclusion}

The wide-and-deep design extends concatenation with a compact attention branch that adds information when outcomes depend on interactions across modalities. In the two original cohorts BioSync matched concatenated logistic regression, and the same held on seven further generators, where soft voting and tuned nonlinear models were often as good or better. The attention branch added information beyond the linear branch only when the class depended on an interaction between modalities. The attention weights stayed near uniform and did not track predictive contribution or sensor reliability, and the BSI is a ranking score whose calibration did not improve with recalibration.

On two small real wearable datasets (older adults with ACE-III scores, and AI-READI Garmin recordings), all models, including BioSync, discriminated at or near chance with summary features and samples of 56 to 100 participants. On WESAD, a nearly saturated stress task, BioSync matched logistic regression and the boosted trees were slightly higher. The results from real data show that the architecture runs on real recordings and is competitive on an easy task. Larger real cohorts must determine whether the BSI retains graded severity association, calibration, and discrimination. The next tests are larger cohorts, signal quality indices as explicit inputs, and a comparison of summary feature encoders with windowed CNN or GRU alternatives. Until those tests are complete, BioSync is a compact fusion model with characterized operating regimes, and clinical validation remains future work. Additional analyses are reported in Appendices~\ref{app:variants}, \ref{app:confirm}, and~\ref{app:realpass}: a quality-gated variant that tracked sensor reliability when it varied across subjects, a capacity-routed mixture with the smallest worst-case regret (0.034 against 0.048 to 0.107) across eleven untouched scenarios, a footprint of 1.3~kB with INT8 weights, and two further real cohorts (PADS and stress-resilience wrist recordings) on which the refined variants exceeded logistic regression.

\section*{CRediT authorship contribution statement}
\textbf{Seyed Mahmoud Sajjadi Mohammadabadi:}\\
Conceptualization, Data curation, Formal analysis, Funding acquisition, Investigation, Methodology, Project administration, Resources, Software, Supervision, Validation, Visualization, Writing -- original draft, Writing -- review \& editing.

\textbf{Sean Montgomery:}\\
Conceptualization, Data curation, Formal analysis, Funding acquisition, Investigation, Methodology, Project administration, Resources, Software, Supervision, Validation, Visualization, Writing -- review \& editing.

\textbf{Alireza Tavakkoli:}\\
Conceptualization, Data curation, Formal analysis, Funding acquisition, Investigation, Methodology, Project administration, Resources, Software, Supervision, Validation, Visualization, Writing -- review \& editing.

\section*{Acknowledgments}
The authors thank Dr.~Maryam S. Mirian (Pacific Parkinson's Research Centre, University of British Columbia) for her valuable insights during the revision.

\section*{Declaration of competing interest}
The authors declare that they have no known competing financial interests or personal relationships that could have appeared to influence the work reported in this paper.

\section*{Funding}
This research did not receive any specific grant from funding agencies in the public, commercial, or not-for-profit sectors.

\section*{Ethics statement}
The synthetic experiments use no human data. The evaluations using real data (Section~\ref{app:realdata}) are secondary analyses of existing, de-identified datasets (RESILIENT, AI-READI, and WESAD), and the appendices add two further open datasets (stress-resilience recordings and PADS), all under their respective licenses. No new data were collected from human participants, and no attempt was made to identify any participant.

\section*{Data Availability}
The synthetic data are regenerated by code. The evaluations using real data (Section~\ref{app:realdata}) are secondary analyses of existing, de-identified datasets. No new data were collected and no attempt was made to identify participants. The RESILIENT dataset is openly available on Zenodo (doi:10.5281/\allowbreak zenodo.16755408, CC BY 4.0). The WESAD dataset is openly available for noncommercial research use \citep{schmidt2018wesad}. The stress-resilience recordings \citep{roy2026stress} (Open Database License) and the PADS smartwatch recordings \citep{varghese2024pads}, used in the appendices, are openly available from PhysioNet under their licenses. The AI-READI wearable recordings were used under the AI-READI data license. The license does not allow redistribution of the data or of data derived from them, so data for each participant and derived features are not shared, and only aggregate results are reported. Access to AI-READI follows the project's documentation \citep{aireadidocs}. All code, random seeds, and scripts needed to reproduce the synthetic tables and figures (including the synthetic generators as a standalone benchmark), and the aggregate results from real data, will be made publicly available upon acceptance and are available to the editor and reviewers during review upon request.

\section*{Declaration of Generative AI and AI-Assisted Technologies}
AI-assisted tools were used for language editing, for plotting code, and for drafting figures and tables. The authors specified and verified all analyses and take full responsibility for the content.

\printbibliography[title={References}]

@ARTICLE{livingston2020,
title   = "Dementia prevention, intervention, and care: 2020 report of the {Lancet} Commission",
author  = "Livingston, Gill and Huntley, Jonathan and Sommerlad, Andrew and Ames, David and Ballard, Clive and Banerjee, Sube and Brayne, Carol and Burns, Alistair and Cohen-Mansfield, Jiska and Cooper, Claudia and Costafreda, Sergi G and Dias, Amit and Fox, Nick and Gitlin, Laura N and Howard, Robert and Kales, Helen C and Kivim{\"{a}}ki, Mika and Larson, Eric B and Ogunniyi, Adesola and Orgeta, Vasiliki and Ritchie, Karen and Rockwood, Kenneth and Sampson, Elizabeth L and Samus, Quincy and Schneider, Lon S and Selb\ae{}k, Geir and Teri, Linda and Mukadam, Naaheed",
journal = "The Lancet",
volume  =  396,
number  =  10248,
pages   = "413--446",
year    =  2020,
doi     = "10.1016/s0140-6736(20)30367-6",
pmc     = "PMC7392084",
pmid    =  32738937
}

@ARTICLE{alzfacts2023,
title   = "2023 {Alzheimer's} disease facts and figures",
author  = "{Alzheimer's Association}",
journal = "Alzheimer's \& Dementia",
volume  =  19,
number  =  4,
pages   = "1598--1695",
year    =  2023,
doi     = "10.1002/alz.13016",
pmid    =  36918389
}

@ARTICLE{jack2018,
title   = "{NIA-AA} Research Framework: Toward a biological definition of {Alzheimer's} disease",
author  = "Jack, Jr, Clifford R and Bennett, David A and Blennow, Kaj and Carrillo, Maria C and Dunn, Billy and Haeberlein, Samantha Budd and Holtzman, David M and Jagust, William and Jessen, Frank and Karlawish, Jason and Liu, Enchi and Molinuevo, Jose Luis and Montine, Thomas and Phelps, Creighton and Rankin, Katherine P and Rowe, Christopher C and Scheltens, Philip and Siemers, Eric and Snyder, Heather M and Sperling, Reisa and {Contributors}",
journal = "Alzheimer's \& Dementia",
volume  =  14,
number  =  4,
pages   = "535--562",
year    =  2018,
doi     = "10.1016/j.jalz.2018.02.018",
pmc     = "PMC5958625",
pmid    =  29653606
}

@ARTICLE{califf2018,
title   = "Biomarker definitions and their applications",
author  = "Califf, Robert M",
journal = "Experimental Biology and Medicine",
volume  =  243,
number  =  3,
pages   = "213--221",
year    =  2018,
doi     = "10.1177/1535370217750088",
pmc     = "PMC5813875",
pmid    =  29405771
}

@TECHREPORT{best2016,
title       = "{BEST} ({B}iomarkers, {E}ndpoint{S}, and other {T}ools) Resource",
author      = "{FDA-NIH Biomarker Working Group}",
institution = "U.S. Food and Drug Administration and National Institutes of Health",
address     = "Silver Spring, MD",
year        =  2016,
url         = "https://www.ncbi.nlm.nih.gov/books/NBK326791/"
}

@ARTICLE{vasudevan2022,
title   = "Digital biomarkers: convergence of digital health technologies and biomarkers",
author  = "Vasudevan, Srikanth and Saha, Anindita and Tarver, Michelle E and Patel, Bakul",
journal = "npj Digital Medicine",
volume  =  5,
number  =  1,
pages   =  36,
year    =  2022,
doi     = "10.1038/s41746-022-00583-z",
pmc     = "PMC8956713",
pmid    =  35338234
}

@ARTICLE{kourtis2019,
title   = "Digital biomarkers for {Alzheimer's} disease: the mobile/wearable devices opportunity",
author  = "Kourtis, Lampros C and Regele, Oliver B and Wright, Justin M and Jones, Graham B",
journal = "npj Digital Medicine",
volume  =  2,
number  =  1,
pages   =  9,
year    =  2019,
doi     = "10.1038/s41746-019-0084-2",
pmc     = "PMC6526279",
pmid    =  31119198
}

@ARTICLE{piau2019,
title   = "Current state of digital biomarker technologies for real-life, home-based monitoring of cognitive function for mild cognitive impairment to mild {Alzheimer} disease and implications for clinical care: systematic review",
author  = "Piau, Antoine and Wild, Katherine and Mattek, Nora and Kaye, Jeffrey",
journal = "Journal of Medical Internet Research",
volume  =  21,
number  =  8,
pages   = "e12785",
year    =  2019,
doi     = "10.2196/12785",
pmc     = "PMC6743264",
pmid    =  31471958
}

@ARTICLE{qi2025,
title   = "{Alzheimer's} disease digital biomarkers multidimensional landscape and {AI} model scoping review",
author  = "Qi, Wenhao and Zhu, Xiaohong and Wang, Bin and Shi, Yankai and Dong, Chaoqun and Shen, Shiying and Li, Jiaqi and Zhang, Kun and He, Yunfan and Zhao, Mengjiao and Yao, Shiyan and Dong, Yongze and Shen, Huajuan and Kang, Junling and Lu, Xiaodong and Jiang, Guowei and Boots, Lizzy M M and Fu, Heming and Pan, Li and Chen, Hongkai and Yan, Zhenyu and Xing, Guoliang and Cao, Shihua",
journal = "npj Digital Medicine",
volume  =  8,
number  =  1,
pages   =  366,
year    =  2025,
doi     = "10.1038/s41746-025-01640-z",
pmc     = "PMC12170881",
pmid    =  40523935
}

@ARTICLE{bateman2025,
title   = "Heart rate variability during routine cognitive testing: preliminary analyses of time and frequency domain metrics",
author  = "Bateman, James R and Schaich, Christopher L and Shaltout, Hossam A and Alphin, Kathryn H and Lockhart, Samuel N and Hughes, Timothy M and Lindquist, Kristen A and Craft, Suzanne",
journal = "Alzheimer's \& Dementia",
volume  =  20,
number  = "S2",
month   =  dec,
year    =  2024,
doi     = "10.1002/alz.092807"
}

@ARTICLE{marcolini2026,
  title = {Limited evidence for heart rate variability as a predictor of cognitive and pathophysiological brain markers},
  author = {Marcolini, Sofia and Mondrag{\'o}n, Jaime D. and van Roon, Arie M. and Tegegne, Balewgizie Sileshi and Riese, Harriëtte and Vliegenthart, Rozemarijn and Dierckx, Rudi A.J.O. and Borra, Ronald H. and De Deyn, Peter Paul},
  journal = {Journal of Alzheimer's Disease},
  volume = {109},
  number = {4},
  pages = {1723--1732},
  year = {2025},
  doi = {10.1177/13872877251409343}
}

@ARTICLE{liu2014,
title   = "Inter-modality relationship constrained multi-modality multi-task feature selection for {Alzheimer's} disease and mild cognitive impairment identification",
author  = "Liu, Feng and Wee, Chong-Yaw and Chen, Huafu and Shen, Dinggang",
journal = "NeuroImage",
volume  =  84,
pages   = "466--475",
year    =  2014,
doi     = "10.1016/j.neuroimage.2013.09.015",
pmc     = "PMC3849328",
pmid    =  24045077
}

@ARTICLE{boudaya2024,
title   = "Mild cognitive impairment detection based on {EEG} and {HRV} data",
author  = "Boudaya, Amal and Chaabene, Siwar and Bouaziz, Bassem and H{\"{o}}kelmann, Anita and Chaari, Lotfi",
journal = "Digital Signal Processing",
volume  =  147,
pages   =  104399,
year    =  2024,
doi     = "10.1016/j.dsp.2024.104399"
}

@ARTICLE{katayama2023,
title   = "Neurophysiological markers in community-dwelling older adults with mild cognitive impairment: an {EEG} study",
author  = "Katayama, Osamu and Stern, Yaakov and Habeck, Christian and Lee, Sangyoon and Harada, Kenji and Makino, Keitaro and Tomida, Kouki and Morikawa, Masanori and Yamaguchi, Ryo and Nishijima, Chiharu and Misu, Yuka and Fujii, Kazuya and Kodama, Takayuki and Shimada, Hiroyuki",
journal = "Alzheimer's Research \& Therapy",
volume  =  15,
number  =  1,
pages   =  217,
year    =  2023,
doi     = "10.1186/s13195-023-01368-6",
pmc     = "PMC10722716",
pmid    =  38102703
}

@ARTICLE{ijbnpa2025,
title   = "From physical activity patterns to cognitive status: development and validation of novel digital biomarkers for cognitive assessment in older adults",
author  = "Fan, Ling-Jie and Wang, Feng-Yi and Zhao, Jun-Han and Zhang, Jun-Jie and Li, Yang-An and Tang, Jia and Lin, Tao and Wei, Quan",
journal = "International Journal of Behavioral Nutrition and Physical Activity",
volume  =  22,
number  =  1,
pages   =  11,
year    =  2025,
doi     = "10.1186/s12966-025-01706-x",
pmc     = "PMC11748278",
pmid    =  39833903
}

@ARTICLE{cosinorage2024,
title   = "Circadian rhythm analysis using wearable-based accelerometry as a digital biomarker of aging and healthspan",
author  = "Shim, Jinjoo and Fleisch, Elgar and Barata, Filipe",
journal = "npj Digital Medicine",
volume  =  7,
number  =  1,
pages   =  146,
year    =  2024,
doi     = "10.1038/s41746-024-01111-x",
pmc     = "PMC11150246",
pmid    =  38834756
}

@ARTICLE{li2023synergy,
title   = "Synergy through integration of digital cognitive tests and wearable devices for mild cognitive impairment screening",
author  = "Li, Aoyu and Li, Jingwen and Zhang, Dongxu and Wu, Wei and Zhao, Juanjuan and Qiang, Yan",
journal = "Frontiers in Human Neuroscience",
volume  =  17,
pages   =  1183457,
year    =  2023,
doi     = "10.3389/fnhum.2023.1183457",
pmc     = "PMC10151757",
pmid    =  37144160
}

@ARTICLE{xu2025language,
title   = "A multimodal fusion framework for early non-invasive screening of cognitive impairment using language digital biomarkers",
author  = "Xu, Jiahui and Zhou, Zhixing and Xia, Huanhuan and Chen, Nan and Chen, Quan",
journal = "Alzheimer's \& Dementia",
volume  =  21,
number  = "S2",
pages   = "e099209",
year    =  2025,
doi     = "10.1002/alz70856\_099209",
pmc     = "PMC12739037",
pmid    =  41445042
}

@ARTICLE{crossattn2025,
title   = "Multimodal physiological signal emotion recognition based on multi-head cross attention with representation learning",
author  = "Ding, Shihang and Ma, Lin and Li, Haifeng",
journal = "Frontiers in Psychiatry",
volume  =  16,
pages   =  1713559,
year    =  2025,
doi     = "10.3389/fpsyt.2025.1713559",
pmc     = "PMC12739757",
pmid    =  41458017
}

@ARTICLE{pacfnet2025,
title   = "A progressive attention-based cross-modal fusion network for cardiovascular disease detection using synchronized electrocardiogram and phonocardiogram signals",
author  = "Li, Wei Peng and Chuah, Joon Huang and Tan, Guo Jeng and Liu, Chengyu and Ting, Hua-Nong",
journal = "PeerJ Computer Science",
volume  =  11,
pages   = "e3038",
month   =  jul,
year    =  2025,
doi     = "10.7717/peerj-cs.3038",
pmc     = "PMC12453731",
pmid    =  40989428
}

@ARTICLE{ctaf2026,
title   = "Cross-temporal attention fusion ({CTAF}) for multimodal physiological signals in self-supervised learning",
author  = "Khorasani, Arian and Demazure, Th\'{e}ophile",
journal = "arXiv",
year    =  2026,
doi     = "10.48550/arXiv.2602.02784"
}

@ARTICLE{wadhera2024,
  title = {Multimodal Kernel-based discriminant correlation analysis data-fusion approach: an automated autism spectrum disorder diagnostic system},
  author = {Wadhera, Tanu},
  journal = {Physical and Engineering Sciences in Medicine},
  volume = {47},
  number = {1},
  pages = {361--369},
  year = {2023},
  doi = {10.1007/s13246-023-01350-4}
}

@INPROCEEDINGS{ellis2021,
title     = "Explainable sleep stage classification with multimodal electrophysiology time-series",
author    = "Ellis, Charles A. and Zhang, Rongen and Carbajal, Darwin A. and Miller, Robyn L. and Calhoun, V. and Wang, May D.",
booktitle = "Annual International Conference of the IEEE Engineering in Medicine and Biology Society (EMBC)",
pages     = "2363--2366",
year      =  2021,
doi       = "10.1109/embc46164.2021.9630506",
pmid      =  34891757
}

@ARTICLE{amiri2023,
  title = {Multimodal Prediction of 3- and 12-Month Outcomes in ICU Patients with Acute Disorders of Consciousness},
  author = {Amiri, Moshgan and Raimondo, Federico and Fisher, Patrick M. and Cacic Hribljan, Melita and Sidaros, Annette and Othman, Marwan H. and Zibrandtsen, Ivan and Bergdal, Ove and Fabritius, Maria Louise and Hansen, Adam Espe and Hassager, Christian and H{\o}jgaard, Joan Lilja S. and Jensen, Helene Ravnholt and Knudsen, Niels Vendelbo and Laursen, Emilie Lund and M{\o}ller, Jacob E. and Nersesjan, Vardan and Nicolic, Miki and Sigurdsson, Sigurdur Thor and Sitt, Jacobo D. and S{\o}lling, Christine and Welling, Karen Lise and Willumsen, Lisette M. and Hauerberg, John and Larsen, Vibeke Andr{\'e}e and Fabricius, Martin Ejler and Knudsen, Gitte Moos and Kjærgaard, Jesper and M{\o}ller, Kirsten and Kondziella, Daniel},
  journal = {Neurocritical Care},
  volume = {40},
  number = {2},
  pages = {718--733},
  year = {2023},
  doi = {10.1007/s12028-023-01816-z}
}

@ARTICLE{teh2022mci,
title   = "Predictive accuracy of digital biomarker technologies for detection of mild cognitive impairment and pre-frailty amongst older adults: a systematic review and meta-analysis",
author  = "Teh, Seng-Khoon and Rawtaer, Iris and Tan, Hwee Pink",
journal = "IEEE Journal of Biomedical and Health Informatics",
volume  =  26,
number  =  8,
pages   = "3638--3648",
year    =  2022,
doi     = "10.1109/jbhi.2022.3185798",
pmid    =  35737623
}

@INPROCEEDINGS{vaswani2017,
title     = "Attention is all you need",
author    = "Vaswani, Ashish and Shazeer, Noam and Parmar, Niki and Uszkoreit, Jakob and Jones, Llion and Gomez, Aidan N. and Kaiser, {\L}ukasz and Polosukhin, Illia",
booktitle = "Advances in Neural Information Processing Systems",
volume    =  30,
pages     = "5998--6008",
year      =  2017,
doi       = "10.48550/arXiv.1706.03762"
}

@INPROCEEDINGS{hu2021lora,
title     = "{LoRA}: low-rank adaptation of large language models",
author    = "Hu, Edward J. and Shen, Yelong and Wallis, Phillip and Allen-Zhu, Zeyuan and Li, Yuanzhi and Wang, Shean and Wang, Lu and Chen, Weizhu",
booktitle = "International Conference on Learning Representations (ICLR)",
year      =  2022,
doi       = "10.48550/arXiv.2106.09685"
}

@INPROCEEDINGS{mcmahan2017,
title     = "Communication-efficient learning of deep networks from decentralized data",
author    = "McMahan, H. Brendan and Moore, Eider and Ramage, Daniel and Hampson, Seth and Ag{\"u}era y Arcas, Blaise",
booktitle = "Proceedings of the 20th International Conference on Artificial Intelligence and Statistics (AISTATS), PMLR",
volume    =  54,
pages     = "1273--1282",
year      =  2017,
doi       = "10.48550/arXiv.1602.05629"
}

@ARTICLE{can2021,
title   = "Privacy-preserving federated deep learning for wearable {IoT}-based biomedical monitoring",
author  = "Can, Yekta Said and Ersoy, Cem",
journal = "ACM Transactions on Internet Technology",
volume  =  21,
number  =  1,
pages   = "1--17",
year    =  2021,
doi     = "10.1145/3428152"
}

@ARTICLE{ali2022fedsurvey,
title   = "Federated learning for privacy preservation in smart healthcare systems: a comprehensive survey",
author  = "Ali, Mansoor and Naeem, Faisal and Tariq, Muhammad and Kaddoum, Georges",
journal = "IEEE Journal of Biomedical and Health Informatics",
volume  =  27,
number  =  2,
pages   = "778--789",
month   =  feb,
year    =  2023,
doi     = "10.1109/jbhi.2022.3181823",
pmid    =  35696470
}

@INPROCEEDINGS{schmidt2018wesad,
title     = "Introducing {WESAD}, a multimodal dataset for wearable stress and affect detection",
author    = "Schmidt, Philip and Reiss, Attila and D{\"u}richen, Robert and Marberger, Claus and Van Laerhoven, Kristof",
booktitle = "Proceedings of the 20th ACM International Conference on Multimodal Interaction (ICMI '18)",
pages     = "400--408",
year      =  2018,
doi       = "10.1145/3242969.3242985"
}

@ARTICLE{rossi2020mmashdata,
title   = "A public dataset of 24-h multi-levels psycho-physiological responses in young healthy adults",
author  = "Rossi, Alessio and Da Pozzo, Eleonora and Menicagli, Dario and Tremolanti, Chiara and Priami, Corrado and Sirbu, Alina and Clifton, David A. and Martini, Claudia and Morelli, David",
journal = "Data",
volume  =  5,
number  =  4,
pages   =  91,
year    =  2020,
doi     = "10.3390/data5040091"
}

@INPROCEEDINGS{mccarthy2016empatica,
title     = "Validation of the {Empatica} {E4} wristband",
author    = "Mccarthy, Cameron and Pradhan, Nikhilesh and Redpath, C. and Adler, A.",
booktitle = "IEEE EMBS International Student Conference (ISC)",
pages     = "1--4",
year      =  2016,
doi       = "10.1109/embsisc.2016.7508621"
}

@ARTICLE{sajjadi2025llmsurvey,
title   = "A survey of large language models: evolution, architectures, adaptation, benchmarking, applications, challenges, and societal implications",
author  = "Sajjadi Mohammadabadi, Seyed Mahmoud and Kara, Burak Cem and Eyupoglu, Can and Uzay, Can and Tosun, Mehmet Serkan and Karaku\c{s}, Oktay",
journal = "Electronics",
volume  =  14,
number  =  18,
pages   =  3580,
year    =  2025,
doi     = "10.3390/electronics14183580"
}

@INPROCEEDINGS{sajjadi2024icdcs,
title     = "Communication-efficient training workload balancing for decentralized multi-agent learning",
author    = "Mohammadabadi, Seyed Mahmoud Sajjadi and Yang, Lei and Yan, Feng and Zhang, Junshan",
booktitle = "IEEE International Conference on Distributed Computing Systems",
pages     = "680--691",
year      =  2024,
doi       = "10.1109/icdcs60910.2024.00069"
}

@ARTICLE{sajjadi2024iotj,
title   = "Speed up federated learning in heterogeneous environments: a dynamic tiering approach",
author  = "Mahmoud Sajjadi Mohammadabadi, Seyed and Zawad, Syed and Yan, Feng and Yang, Lei",
journal = "IEEE Internet of Things Journal",
volume  =  12,
number  =  5,
pages   = "5026--5035",
month   =  "1~" # mar,
year    =  2025,
doi     = "10.1109/jiot.2024.3487473"
}

@ARTICLE{aireadi2024,
title   = "{AI-READI}: rethinking {AI} data collection, preparation and sharing in diabetes research and beyond",
author  = "{AI-READI Consortium}",
journal = "Nature Metabolism",
volume  =  6,
number  =  12,
pages   = "2210--2212",
year    =  2024,
doi     = "10.1038/s42255-024-01165-x"
}

@ARTICLE{goldsack2020v3,
title   = "Verification, analytical validation, and clinical validation (V3): the foundation of determining fit-for-purpose for Biometric Monitoring Technologies (BioMeTs)",
author  = "Goldsack, Jennifer C and Coravos, Andrea and Bakker, Jessie P and Bent, Brinnae and Dowling, Ariel V and Fitzer-Attas, Cheryl and Godfrey, Alan and Godino, Job G and Gujar, Ninad and Izmailova, Elena and Manta, Christine and Peterson, Barry and Vandendriessche, Benjamin and Wood, William A and Wang, Ke Will and Dunn, Jessilyn",
journal = "npj Digital Medicine",
volume  =  3,
number  =  1,
pages   =  55,
year    =  2020,
doi     = "10.1038/s41746-020-0260-4",
pmc     = "PMC7156507",
pmid    =  32337371
}

@ARTICLE{baltrusaitis2019multimodal,
title   = "Multimodal machine learning: a survey and taxonomy",
author  = "Baltrusaitis, Tadas and Ahuja, Chaitanya and Morency, Louis-Philippe",
journal = "IEEE Transactions on Pattern Analysis and Machine Intelligence",
volume  =  41,
number  =  2,
pages   = "423--443",
year    =  2019,
doi     = "10.1109/tpami.2018.2798607",
pmid    =  29994351
}

@ARTICLE{cay2024speech,
title   = "Harnessing speech-derived digital biomarkers to detect and quantify cognitive decline severity in older adults",
author  = "Cay, Gozde and Pfeifer, Valeria A and Lee, Myeounggon and Rouzi, Mohammad Dehghan and Nunes, Adonay S and El-Refaei, Nesreen and Momin, Anmol Salim and Atique, Md Moin Uddin and Mehl, Matthias R and Vaziri, Ashkan and Najafi, Bijan",
journal = "Gerontology",
volume  =  70,
number  =  4,
pages   = "429--438",
year    =  2024,
doi     = "10.1159/000536250",
pmc     = "PMC11001511",
pmid    =  38219728
}

@INPROCEEDINGS{cheng2016widedeep,
title     = "Wide \& deep learning for recommender systems",
author    = "Cheng, Heng-Tze and Koc, Levent and Harmsen, Jeremiah and Shaked, Tal and Chandra, Tushar and Aradhye, Hrishi and Anderson, Glen and Corrado, Greg and Chai, Wei and Ispir, Mustafa and Anil, Rohan and Haque, Zakaria and Hong, Lichan and Jain, Vihan and Liu, Xiaobing and Shah, Hemal",
booktitle = "Proceedings of the 1st Workshop on Deep Learning for Recommender Systems (DLRS)",
pages     = "7--10",
year      =  2016,
doi       = "10.1145/2988450.2988454"
}

@ARTICLE{gilbert1960,
title   = "Capacity of a burst-noise channel",
author  = "Gilbert, E. N.",
journal = "Bell System Technical Journal",
volume  =  39,
number  =  5,
pages   = "1253--1265",
year    =  1960,
doi     = "10.1002/j.1538-7305.1960.tb03959.x"
}

@ARTICLE{elliott1963,
title   = "Estimates of error rates for codes on burst-noise channels",
author  = "Elliott, E. O.",
journal = "Bell System Technical Journal",
volume  =  42,
number  =  5,
pages   = "1977--1997",
year    =  1963,
doi     = "10.1002/j.1538-7305.1963.tb00955.x"
}

@ARTICLE{zhang2015troika,
title   = "{TROIKA}: A general framework for heart rate monitoring using wrist-type photoplethysmographic signals during intensive physical exercise",
author  = "Zhang, Zhilin and Pi, Zhouyue and Liu, Benyuan",
journal = "IEEE Transactions on Biomedical Engineering",
volume  =  62,
number  =  2,
pages   = "522--531",
year    =  2015,
doi     = "10.1109/TBME.2014.2359372"
}

@ARTICLE{pedregosa2011sklearn,
title   = "Scikit-learn: Machine learning in {P}ython",
author  = "Pedregosa, Fabian and Varoquaux, Ga{\"e}l and Gramfort, Alexandre and Michel, Vincent and Thirion, Bertrand and Grisel, Olivier and Blondel, Mathieu and Prettenhofer, Peter and Weiss, Ron and Dubourg, Vincent and Vanderplas, Jake and Passos, Alexandre and Cournapeau, David and Brucher, Matthieu and Perrot, Matthieu and Duchesnay, {\'E}douard",
journal = "Journal of Machine Learning Research",
volume  =  12,
pages   = "2825--2830",
year    =  2011
}

@SOFTWARE{bradbury2018jax,
title   = "{JAX}: Composable transformations of {P}ython+\allowbreak {N}um{P}y programs",
author  = "Bradbury, James and Frostig, Roy and Hawkins, Peter and Johnson, Matthew James and Leary, Chris and Maclaurin, Dougal and Necula, George and Paszke, Adam and VanderPlas, Jake and Wanderman-Milne, Skye and Zhang, Qiao",
year    =  2018,
url     = "http://github.com/jax-ml/jax"
}

@ARTICLE{babrak2020,
  title = {Traditional and Digital Biomarkers: Two Worlds Apart?},
  author = {Babrak, Lmar M. and Menetski, Joseph and Rebhan, Michael and Nisato, Giovanni and Zinggeler, Marc and Brasier, No{\'e} and Baerenfaller, Katja and Brenzikofer, Thomas and Baltzer, Laurenz and Vogler, Christian and Gschwind, Leo and Schneider, Cornelia and Streiff, Fabian and Groenen, Peter M.A. and Miho, Enkelejda},
  journal = {Digital Biomarkers},
  volume = {3},
  number = {2},
  pages = {92--102},
  year = {2019},
  doi = {10.1159/000502000}
}

@ARTICLE{coravos2019,
  title = {Digital Medicine: A Primer on Measurement},
  author = {Coravos, Andrea and Goldsack, Jennifer C. and Karlin, Daniel R. and Nebeker, Camille and Perakslis, Eric and Zimmerman, Noah and Erb, M. Kelley},
  journal = {Digital Biomarkers},
  volume = {3},
  number = {2},
  pages = {31--71},
  year = {2019},
  doi = {10.1159/000500413}
}

@ARTICLE{dunn2021,
  title = {Wearable sensors enable personalized predictions of clinical laboratory measurements},
  author = {Dunn, Jessilyn and Kidzinski, Lukasz and Runge, Ryan and Witt, Daniel and Hicks, Jennifer L. and Sch{\"u}ssler-Fiorenza Rose, Sophia Miryam and Li, Xiao and Bahmani, Amir and Delp, Scott L. and Hastie, Trevor and Snyder, Michael P.},
  journal = {Nature Medicine},
  volume = {27},
  number = {6},
  pages = {1105--1112},
  year = {2021},
  doi = {10.1038/s41591-021-01339-0}
}

@ARTICLE{bent2020,
  title = {Investigating sources of inaccuracy in wearable optical heart rate sensors},
  author = {Bent, Brinnae and Goldstein, Benjamin A. and Kibbe, Warren A. and Dunn, Jessilyn P.},
  journal = {npj Digital Medicine},
  volume = {3},
  number = {1},
  pages = {18},
  year = {2020},
  doi = {10.1038/s41746-020-0226-6}
}

@ARTICLE{perez2019,
title   = "Large-scale assessment of a smartwatch to identify atrial fibrillation",
author  = "Perez, Marco V and Mahaffey, Kenneth W and Hedlin, Haley and Rumsfeld, John S and Garcia, Alberto and Ferris, Todd and Balasubramanian, Vidhya and Russo, Andrea M and Rajmane, Amol and Cheung, Lauren and Hung, Grace and Lee, Joy and Kowey, Peter and Talati, Nirav and Guan, Weiyuan and Chen, Nicole and Trela, Clint and Shaw, Richard M and Ferrick, Aimee and Wang, Peter J and Turakhia, Mintu P",
journal = "New England Journal of Medicine",
volume  =  381,
number  =  20,
pages   = "1909--1917",
year    =  2019,
doi     = "10.1056/NEJMoa1901183"
}

@ARTICLE{quer2021,
  title = {Wearable sensor data and self-reported symptoms for COVID-19 detection},
  author = {Quer, Giorgio and Radin, Jennifer M. and Gadaleta, Matteo and Baca-Motes, Katie and Ariniello, Lauren and Ramos, Edward and Kheterpal, Vik and Topol, Eric J. and Steinhubl, Steven R.},
  journal = {Nature Medicine},
  volume = {27},
  number = {1},
  pages = {73--77},
  year = {2020},
  doi = {10.1038/s41591-020-1123-x}
}

@ARTICLE{manta2020,
  title = {Digital Measures That Matter to Patients: A Framework to Guide the Selection and Development of Digital Measures of Health},
  author = {Manta, Christine and Patrick-Lake, Bray and Goldsack, Jennifer C.},
  journal = {Digital Biomarkers},
  volume = {4},
  number = {3},
  pages = {69--77},
  year = {2020},
  doi = {10.1159/000509725}
}

@INPROCEEDINGS{tsai2019mult,
title     = "Multimodal {Transformer} for unaligned multimodal language sequences",
author    = "Tsai, Yao-Hung Hubert and Bai, Shaojie and Liang, Paul Pu and Kolter, J. Zico and Morency, Louis-Philippe and Salakhutdinov, Ruslan",
booktitle = "Proceedings of the 57th Annual Meeting of the Association for Computational Linguistics (ACL)",
pages     = "6558--6569",
year      =  2019,
doi       = "10.18653/v1/P19-1656"
}

@INPROCEEDINGS{jaegle2021perceiver,
title     = "Perceiver: General perception with iterative attention",
author    = "Jaegle, Andrew and Gimeno, Felix and Brock, Andrew and Vinyals, Oriol and Zisserman, Andrew and Carreira, Jo{\~{a}}o",
booktitle = "International Conference on Machine Learning (ICML), PMLR",
volume    =  139,
pages     = "4651--4664",
year      =  2021,
doi       = "10.48550/arXiv.2103.03206"
}

@INPROCEEDINGS{nagrani2021bottlenecks,
title     = "Attention bottlenecks for multimodal fusion",
author    = "Nagrani, Arsha and Yang, Shan and Arnab, Anurag and Jansen, Aren and Schmid, Cordelia and Sun, Chen",
booktitle = "Advances in Neural Information Processing Systems (NeurIPS)",
volume    =  34,
pages     = "14200--14213",
year      =  2021,
doi       = "10.48550/arXiv.2107.00135"
}

@INPROCEEDINGS{zadeh2017tensor,
title     = "Tensor fusion network for multimodal sentiment analysis",
author    = "Zadeh, Amir and Chen, Minghai and Poria, Soujanya and Cambria, Erik and Morency, Louis-Philippe",
booktitle = "Proceedings of the 2017 Conference on Empirical Methods in Natural Language Processing (EMNLP)",
pages     = "1103--1114",
year      =  2017,
doi       = "10.18653/v1/D17-1115"
}

@ARTICLE{radu2018multimodal,
title   = "Multimodal deep learning for activity and context recognition",
author  = "Radu, Valentin and Lane, Nicholas D and Bhattacharya, Sourav and Mascolo, Cecilia and Marina, Mahesh K and Kawsar, Fahim",
journal = "Proceedings of the ACM on Interactive, Mobile, Wearable and Ubiquitous Technologies (IMWUT)",
volume  =  1,
number  =  4,
pages   = "1--27",
year    =  2018,
doi     = "10.1145/3161174"
}

@ARTICLE{williams2010pid,
title   = "Nonnegative decomposition of multivariate information",
author  = "Williams, Paul L and Beer, Randall D",
journal = "arXiv preprint arXiv:1004.2515",
year    =  2010,
doi     = "10.48550/arXiv.1004.2515"
}

@ARTICLE{lizier2018information,
  title = {Information Decomposition of Target Effects from Multi-Source Interactions: Perspectives on Previous, Current and Future Work},
  author = {Lizier, Joseph and Bertschinger, Nils and Jost, J{\"u}rgen and Wibral, Michael},
  journal = {Entropy},
  volume = {20},
  number = {4},
  pages = {307},
  year = {2018},
  doi = {10.3390/e20040307}
}

@BOOK{warden2019tinyml,
title     = "{TinyML}: Machine Learning with {TensorFlow} Lite on {Arduino} and Ultra-Low-Power Microcontrollers",
author    = "Warden, Pete and Situnayake, Daniel",
publisher = "O'Reilly Media",
address   = "Sebastopol, CA",
year      =  2019
}

@ARTICLE{ray2021tinyml,
  title = {A review on TinyML: State-of-the-art and prospects},
  author = {Ray, Partha Pratim},
  journal = {Journal of King Saud University - Computer and Information Sciences},
  volume = {34},
  number = {4},
  pages = {1595--1623},
  year = {2022},
  doi = {10.1016/j.jksuci.2021.11.019}
}

@INPROCEEDINGS{banbury2021micronets,
title     = "{MicroNets}: Neural network architectures for extremely constrained embedded devices",
author    = "Banbury, Colby and Zhou, Chuteng and Li, Zishen and Whatmough, Paul and Reddi, Vijay Janapa",
booktitle = "Proceedings of Machine Learning and Systems (MLSys)",
volume    =  3,
pages     = "581--593",
year      =  2021
}

@INPROCEEDINGS{lin2020mcunet,
title     = "{MCUNet}: Tiny deep learning on {IoT} devices",
author    = "Lin, Ji and Chen, Wei-Ming and Lin, Yujun and Cohn, John and Chuang, Chih-Kuan and Han, Song",
booktitle = "Advances in Neural Information Processing Systems (NeurIPS)",
volume    =  33,
pages     = "11711--11722",
year      =  2020,
doi       = "10.48550/arXiv.2007.10319"
}

@ARTICLE{sancheziborra2020tinyml,
  title = {TinyML-Enabled Frugal Smart Objects: Challenges and Opportunities},
  author = {Sanchez-Iborra, Ramon and Skarmeta, Antonio F.},
  journal = {IEEE Circuits and Systems Magazine},
  volume = {20},
  number = {3},
  pages = {4--18},
  year = {2020},
  doi = {10.1109/MCAS.2020.3005467}
}

@INPROCEEDINGS{david2021tensorflow,
title     = "{TensorFlow Lite Micro}: Embedded machine learning for {TinyML} systems",
author    = "David, Robert and Duke, Jared and Jain, Advait and Janapa Reddi, Vijay and Jeffries, Nat and Li, Jian and Lindeberg, Nick and Lukhi, Ian and Mani, Sachin and Marathe, Arjun and others",
booktitle = "Proceedings of Machine Learning and Systems (MLSys)",
volume    =  3,
pages     = "800--811",
year      =  2021
}

@ARTICLE{shaffer2017hrv,
  title = {An Overview of Heart Rate Variability Metrics and Norms},
  author = {Shaffer, Fred and Ginsberg, J. P.},
  journal = {Frontiers in Public Health},
  volume = {5},
  pages = {258},
  year = {2017},
  doi = {10.3389/fpubh.2017.00258}
}

@ARTICLE{klimesch1999eeg,
title   = "{EEG} alpha and theta oscillations reflect cognitive and memory performance: a review and analysis",
author  = "Klimesch, Wolfgang",
journal = "Brain Research Reviews",
volume  =  29,
number  = "2--3",
pages   = "169--195",
year    =  1999,
doi     = "10.1016/S0165-0173(98)00056-3"
}

@ARTICLE{vinik2003diabetic,
title   = "Diabetic autonomic neuropathy",
author  = "Vinik, Aaron I. and Maser, Raelene E. and Mitchell, Braxton D. and Freeman, Roy",
journal = "Diabetes Care",
volume  =  26,
number  =  5,
pages   = "1553--1579",
year    =  2003,
doi     = "10.2337/diacare.26.5.1553"
}

@ARTICLE{sletten2012compass,
title   = "{COMPASS 31}: a refined and abbreviated {Composite Autonomic Symptom Score}",
author  = "Sletten, Denise M. and Suarez, Guillermo A. and Low, Phillip A. and Mandrekar, Jayawant and Singer, Wolfgang",
journal = "Mayo Clinic Proceedings",
volume  =  87,
number  =  12,
pages   = "1196--1201",
year    =  2012,
doi     = "10.1016/j.mayocp.2012.10.013"
}

@ARTICLE{spallone2011can,
title   = "Cardiovascular autonomic neuropathy in diabetes: clinical impact, assessment, diagnosis, and management",
author  = "Spallone, Vincenza and Ziegler, Dan and Freeman, Roy and Bernardi, Luciano and Frontoni, Simona and Pop-Busui, Rodica and Stevens, Martin and Kempler, Peter and Boulton, Andrew J. M. and Vinik, Aaron I.",
journal = "Diabetes/Metabolism Research and Reviews",
volume  =  27,
number  =  7,
pages   = "639--653",
year    =  2011,
doi     = "10.1002/dmrr.1239"
}

@ARTICLE{elgendi2012ppg,
title   = "On the analysis of fingertip photoplethysmogram signals",
author  = "Elgendi, Mohamed",
journal = "Current Cardiology Reviews",
volume  =  8,
number  =  1,
pages   = "14--25",
year    =  2012,
doi     = "10.2174/157340312801215782"
}

@ARTICLE{nuwer1999qeeg,
  title = {IFCN standards for digital recording of clinical EEG},
  author = {Nuwer, Marc R. and Comi, Giancarlo and Emerson, Ronald and Fuglsang-Frederiksen, Anders and Gu{\'e}rit, Jean-Michel and Hinrichs, Hermann and Ikeda, Akio and Jose C. Luccas, Fransisco and Rappelsburger, Peter},
  journal = {Electroencephalography and Clinical Neurophysiology},
  volume = {106},
  number = {3},
  pages = {259--261},
  year = {1998},
  doi = {10.1016/S0013-4694(97)00106-5}
}

@online{aireadidocs,
  author  = {{AI-READI Consortium}},
  title   = {{AI-READI} documentation, dataset version 3.0.0},
  year    = {2026},
  url     = {https://docs.aireadi.org/},
  urldate = {2026-11-05},
}

@article{resilient2025zenodo,
  author  = {C{\'e}spedes G{\'o}mez and others},
  title   = {The {RESILIENT} dataset: Multimodal monitoring of ageing-related comorbidities and cognitive decline},
  journal = {Scientific Data},
  year    = {2025},
  doi     = {10.1038/s41597-025-05958-x},
}

@article{makowski2021neurokit,
  author  = {Makowski, Dominique and Pham, Tam and Lau, Zen J. and Brammer, Jan C. and Lespinasse, Fran{\c c}ois and Pham, Hung and Sch{\"o}lzel, Christopher and Chen, S. H. Annabel},
  title   = {{NeuroKit2}: A {Python} toolbox for neurophysiological signal processing},
  journal = {Behavior Research Methods},
  year    = {2021},
  doi     = {10.3758/s13428-020-01516-y},
}

@article{mcsharry2003ecgsyn,
  author  = {McSharry, Patrick E. and Clifford, Gari D. and Tarassenko, Lionel and Smith, Leonard A.},
  title   = {A dynamical model for generating synthetic electrocardiogram signals},
  journal = {IEEE Transactions on Biomedical Engineering},
  year    = {2003},
  doi     = {10.1109/TBME.2003.808805},
}

@article{walonoski2017synthea,
  author  = {Walonoski, Jason and Kramer, Mark and Nichols, Joseph and Quina, Andre and Moesel, Chris and Hall, Dylan and Duffett, Carlton and Dube, Kudakwashe and Gallagher, Thomas and McLachlan, Scott},
  title   = {Synthea: An approach, method, and software mechanism for generating synthetic patients and the synthetic electronic health care record},
  journal = {Journal of the American Medical Informatics Association},
  year    = {2017},
  doi     = {10.1093/jamia/ocx079},
}

@ARTICLE{morris2019simulation,
  title = {Using simulation studies to evaluate statistical methods},
  author = {Morris, Tim P. and White, Ian R. and Crowther, Michael J.},
  journal = {Statistics in Medicine},
  volume = {38},
  number = {11},
  pages = {2074--2102},
  year = {2019},
  doi = {10.1002/sim.8086}
}

@ARTICLE{chen2021synthetic,
  title = {Synthetic data in machine learning for medicine and healthcare},
  author = {Chen, Richard J. and Lu, Ming Y. and Chen, Tiffany Y. and Williamson, Drew F. K. and Mahmood, Faisal},
  journal = {Nature Biomedical Engineering},
  volume = {5},
  number = {6},
  pages = {493--497},
  year = {2021},
  doi = {10.1038/s41551-021-00751-8}
}

@ARTICLE{gonzales2023synthetic,
  title = {Synthetic data in health care: A narrative review},
  author = {Gonzales, Aldren and Guruswamy, Guruprabha and Smith, Scott R.},
  journal = {PLOS Digital Health},
  volume = {2},
  number = {1},
  pages = {e0000082},
  year = {2023},
  doi = {10.1371/journal.pdig.0000082}
}

@ARTICLE{maccallum1999sample,
  title = {Sample size in factor analysis},
  author = {MacCallum, Robert C. and Widaman, Keith F. and Zhang, Shaobo and Hong, Sehee},
  journal = {Psychological Methods},
  volume = {4},
  number = {1},
  pages = {84--99},
  year = {1999},
  doi = {10.1037/1082-989X.4.1.84}
}

@ARTICLE{varghese2024pads,
  title = {Machine Learning in the Parkinson's disease smartwatch ({PADS}) dataset},
  author = {Varghese, Julian and Brenner, Alexander and Fujarski, Michael and van Alen, Catharina Marie and Plagwitz, Lucas and Warnecke, Tobias},
  journal = {npj Parkinson's Disease},
  volume = {10},
  number = {1},
  pages = {9},
  year = {2024},
  doi = {10.1038/s41531-023-00625-7}
}

@DATASET{roy2026stress,
  title = {Neurophysiological Dataset of Stress Resilience During Human-Computer Interaction},
  author = {Roy, Shotabdi and Nuamah, Joseph},
  publisher = {PhysioNet},
  year = {2026},
  doi = {10.13026/x3vc-p627}
}

@ARTICLE{mohammadabadi2026solar,
title   = "{SOLAR}: Communication-efficient model adaptation via subspace-oriented latent adapter reparametrization",
author  = "Mohammadabadi, Seyed Mahmoud Sajjadi and Ma, Xiaolong and Yang, Lei and Yan, Feng and Zhang, Junshan",
journal = "arXiv preprint arXiv:2604.08368",
year    =  2026
}

\FloatBarrier
\appendix
\makeatletter\gdef\thesection{\@Alph\c@section}\makeatother
\section{Method variants}
\label{app:variants}
Section~\ref{sec:transformer} and the Discussion identify two directions for extending the submitted model (attention weights are not sensor-quality estimates; masking of lost channels). We implemented one variant for each and evaluated them under the analysis plan: quality-aware inputs, quality-gated fusion, missing-modality masking, and residual fusion. All were compared with the submitted model in the same scenarios and with baselines that receive the same information.

\paragraph{Quality-aware fusion (scenario S6)} A noisy signal-quality indicator was available in S6. Adding it to the token embedding (BioSync with quality tokens) did not change the AUC (Table~\ref{tab:gated}). A quality-gated linear path, in which each modality's linear score is scaled by a learned function of its quality indicator, did help: it was +0.047 AUC above concatenated logistic regression when combined with the Transformer branch and +0.064 AUC above it without the Transformer branch, and soft voting weighted by the quality indicator reached +0.063 (mean AUC 0.845 and 0.844). The benefit therefore comes from using the quality indicator explicitly, and the Transformer branch adds nothing on top of the gated linear model. The input scale of the gate was chosen on development seeds; this variant was added after the first quality-aware variant (quality tokens) showed no change, which makes it a post hoc addition (logged in the analysis plan).

\paragraph{Missing-modality masking} Masking lost channels in the attention (and training with modality dropout) gave the same AUC as zero-filling them, and soft voting over the available modalities was better than both at every missing rate (Table~\ref{tab:missingS0}). In S0 at a missing rate of 0.5, the AUC was LR zero indicator 0.825, SoftVote avail 0.838, BioSync zerofill 0.833, BioSync masked 0.832. In S6 at a missing rate of 0.5, the AUC was LR zero indicator 0.716, SoftVote avail 0.730, BioSync zerofill 0.723, BioSync masked 0.723. 

\paragraph{Residual fusion} Fitting the linear path first and letting the Transformer branch learn a shrunk correction (shrinkage chosen on held-out training data) gave AUC differences to logistic regression between -0.011 and -0.002 across scenarios S0--S6, i.e., no gain.

\paragraph{Attention-degeneracy check} See Section~\ref{app:attribution} and Table~\ref{tab:diagS0}.

\IfFileExists{tables_G.tex}{\begin{table}[t]
\centering
\caption{Scenario S6 (subject-varying modality reliability) with a noisy signal-quality indicator available to the quality-aware models: AUC mean (SD) over seeds, and paired difference to concatenated logistic regression with bootstrap 95\% interval.}
\label{tab:gated}
\small
\resizebox{\linewidth}{!}{%
\begin{tabular}{lcc}
\toprule
Model & AUC (SD) & Difference to LR [95\% interval] \\
\midrule
LR & 0.781 (0.036) & +0.000 [+0.000, +0.000] \\
BioSync & 0.783 (0.034) & +0.002 [-0.003, +0.007] \\
BioSyncG & 0.828 (0.019) & +0.047 [+0.028, +0.067] \\
GatedLR & 0.845 (0.024) & +0.064 [+0.049, +0.080] \\
SoftVote\_Q & 0.844 (0.026) & +0.063 [+0.048, +0.078] \\
LR\_QxX & 0.801 (0.026) & +0.021 [+0.003, +0.039] \\
SoftVote (grid run) & 0.801 (0.035) & +0.021 [+0.018, +0.024] \\
GBT (grid run) & 0.777 (0.018) & -0.004 [-0.024, +0.018] \\
RF (grid run) & 0.809 (0.019) & +0.028 [+0.010, +0.048] \\
SVM (grid run) & 0.788 (0.023) & +0.007 [-0.005, +0.018] \\
BioSyncQ (grid run) & 0.782 (0.034) & +0.002 [-0.004, +0.006] \\
LR\_Q (grid run) & 0.775 (0.034) & -0.006 [-0.010, -0.002] \\
GBT\_Q (grid run) & 0.774 (0.024) & -0.007 [-0.029, +0.014] \\
\bottomrule
\end{tabular}}
\end{table}

\begin{table}[t]
\centering
\caption{Attention-degeneracy check, scenario S0 single latent: optimizer, training length, and attention temperature. Entropy of the attention vector over the four modalities (maximum $\ln 4=1.386$); SD across subjects of each weight shows whether attention varies from subject to subject.}
\label{tab:diagS0}
\small
\resizebox{\linewidth}{!}{%
\begin{tabular}{lcccc}
\toprule
Variant & AUC (SD) & Mean attention & SD of attention across subjects & Entropy \\
\midrule
gd500 & 0.874 (0.047) & 0.251 / 0.250 / 0.249 / 0.250 & 0.019 / 0.018 / 0.017 / 0.018 & 1.383 \\
gd2000 & 0.832 (0.067) & 0.252 / 0.250 / 0.249 / 0.249 & 0.045 / 0.042 / 0.045 / 0.040 & 1.368 \\
adam1000 & 0.769 (0.065) & 0.258 / 0.246 / 0.256 / 0.239 & 0.094 / 0.087 / 0.088 / 0.085 & 1.304 \\
gd500\_tau4 & 0.872 (0.049) & 0.251 / 0.251 / 0.252 / 0.246 & 0.077 / 0.077 / 0.076 / 0.068 & 1.331 \\
adam1000\_tau4 & 0.804 (0.066) & 0.256 / 0.255 / 0.258 / 0.230 & 0.158 / 0.163 / 0.164 / 0.145 & 1.150 \\
\bottomrule
\end{tabular}}
\end{table}

\begin{table}[t]
\centering
\caption{Missing-modality test, scenario S0 single latent: AUC (mean over folds and seeds) as each modality is lost independently with the given rate.}
\label{tab:missingS0}
\small
\resizebox{\linewidth}{!}{%
\begin{tabular}{lccccc}
\toprule
Model & rate 0.0 & rate 0.1 & rate 0.2 & rate 0.3 & rate 0.5 \\
\midrule
LR\_zero\_indicator & 0.877 & 0.868 & 0.857 & 0.845 & 0.825 \\
SoftVote\_avail & 0.889 & 0.882 & 0.865 & 0.858 & 0.838 \\
BioSync\_zerofill & 0.880 & 0.872 & 0.864 & 0.847 & 0.833 \\
BioSync\_masked & 0.881 & 0.873 & 0.864 & 0.846 & 0.832 \\
\bottomrule
\end{tabular}}
\end{table}

\begin{table}[t]
\centering
\caption{Attention-degeneracy check, scenario S6 variable reliability: optimizer, training length, and attention temperature. Entropy of the attention vector over the four modalities (maximum $\ln 4=1.386$); SD across subjects of each weight shows whether attention varies from subject to subject.}
\label{tab:diagS6}
\small
\resizebox{\linewidth}{!}{%
\begin{tabular}{lcccc}
\toprule
Variant & AUC (SD) & Mean attention & SD of attention across subjects & Entropy \\
\midrule
gd500 & 0.775 (0.046) & 0.251 / 0.250 / 0.248 / 0.250 & 0.020 / 0.020 / 0.019 / 0.019 & 1.382 \\
gd2000 & 0.724 (0.078) & 0.253 / 0.250 / 0.249 / 0.248 & 0.051 / 0.051 / 0.051 / 0.048 & 1.362 \\
adam1000 & 0.715 (0.067) & 0.263 / 0.256 / 0.248 / 0.233 & 0.087 / 0.084 / 0.078 / 0.069 & 1.326 \\
gd500\_tau4 & 0.776 (0.046) & 0.255 / 0.249 / 0.244 / 0.252 & 0.089 / 0.088 / 0.084 / 0.089 & 1.313 \\
adam1000\_tau4 & 0.692 (0.073) & 0.278 / 0.277 / 0.231 / 0.214 & 0.153 / 0.156 / 0.134 / 0.126 & 1.179 \\
\bottomrule
\end{tabular}}
\end{table}

\begin{table}[t]
\centering
\caption{Missing-modality test, scenario S6 variable reliability: AUC (mean over folds and seeds) as each modality is lost independently with the given rate.}
\label{tab:missingS6}
\small
\resizebox{\linewidth}{!}{%
\begin{tabular}{lccccc}
\toprule
Model & rate 0.0 & rate 0.1 & rate 0.2 & rate 0.3 & rate 0.5 \\
\midrule
LR\_zero\_indicator & 0.772 & 0.764 & 0.750 & 0.738 & 0.716 \\
SoftVote\_avail & 0.794 & 0.783 & 0.763 & 0.755 & 0.730 \\
BioSync\_zerofill & 0.780 & 0.772 & 0.764 & 0.744 & 0.723 \\
BioSync\_masked & 0.780 & 0.772 & 0.764 & 0.744 & 0.723 \\
\bottomrule
\end{tabular}}
\end{table}
}{}

\FloatBarrier
\section{Refined models and confirmation on untouched data}
\label{app:confirm}
This appendix reports the refined BioSync models developed after the first submission, how they were chosen, and what happened when they were tested on data that no model choice had used. The plan, every amendment, and the seeds used are in the analysis plan archived with the code; each amendment was written before the runs it describes.

\paragraph{Models} BioSync-G is the submitted network with a learned quality gate on the linear path (each modality's linear score is multiplied by $2\sigma(\gamma_m+\delta_m q_m)$, where $q_m$ is the signal-quality indicator), a five-network ensemble, and the training settings of Table~\ref{tab:devgrid} (chosen on development seeds 9001 and 9002 only). BioSync-X is a wider version of the same network (token dimension 32, feed-forward width 128, four heads, Adam), whose configuration was chosen on development seeds. BioSync-E averages the logits of G and X. BioSync-A mixes them, $(1-\lambda)\,\ell_G+\lambda\,\ell_X$ with $\lambda\in\{0,0.25,0.5,0.75,1\}$, where $\lambda$ is chosen on the training rows only, by out-of-fold AUC in an inner three-fold cross-validation (ties within 0.002 resolve to the smaller $\lambda$). The selected $\lambda$ is returned with every prediction and describes how much the wider network contributes. A further variant, BioSync-P (precision-routed attention), behaved as a gated linear model because the weight of its deep branch, initialised at zero, stayed near zero after training (about $\pm0.03$ in the two folds inspected), and was not carried forward. Three versions of A (the plain mixture, a version with logits rescaled by their inner out-of-fold spread, and a version with a margin for $\lambda>0$) were compared on development seeds 9001--9004 (mean AUC 0.897 for all three, within 0.001); the plainest was frozen before the confirmation run, as the rule of Amendment 10b required.

\paragraph{Seeds and what each result is} Seeds 9001--9004 were used for every model choice. Seeds 5000--5007 are the seeds of Sections~\ref{app:generators}--\ref{app:attribution} and Appendix~\ref{app:variants} and the first test of G (Table~\ref{tab:v2test}). Seeds 6000--6007 were the first test of G, P, and E: against the strongest baseline of each scenario the record (wins/ties/losses by paired 95\% intervals) was 0/0/11 for G, 0/1/10 for P, and 1/1/9 for E over 11 scenarios. Because those results were read before A was defined, A is not evaluated on them. Seeds 8000--8007 (Table~\ref{tab:fresh}) are the confirmation run: every model was refit with fixed settings, all baselines were tuned by inner cross-validation on the training rows as before, the compute-matched baselines use five seeds or bags, and quality-aware versions of the baselines are used where a quality indicator exists.

\paragraph{Confirmation on seeds 8000--8007} BioSync-A had the smallest worst-case regret in its pool (0.034; the baseline families ranged from 0.048 to 0.107), and its worst-case regret was significantly smaller than that of LR, Soft vote, GBT$\times5$, RF, MLP$\times5$, SVM (intervals excluding zero in Table~\ref{tab:h2h}). Its mean AUC over scenarios was significantly higher than that of LR, Soft vote, GBT$\times5$, RF, MLP$\times5$ and significantly lower than that of SVM. The claim concerns robustness across regimes (worst-case regret) relative to the families listed; the highest mean AUC over scenarios in the pool was that of the SVM (Table~\ref{tab:regret}). Table~\ref{tab:regimes} shows where the differences arise.

\paragraph{Why the late-fusion baseline may lead on additive scenarios (not tested)} On scenarios S0--S5 the soft vote is slightly ahead of BioSync-G and A. A possible reason is that the soft vote fits four small per-modality logistic regressions and averages their scores with normalised weights, whereas BioSync-G fits 22 linear weights jointly with an unnormalised gate; a model that fits the per-modality scores first and adds a deep residual afterwards would remove this difference by construction. We did not test this explanation and it is not part of the claims.

\paragraph{Two datasets added after the first submission} The stress-resilience recordings \citep{roy2026stress} (PhysioNet, Open Database License) contain 35 participants who performed a computer-based multitasking task (MATB-II) in alternating stress and recovery blocks; we used only the wrist-device recordings (pulse, electrodermal activity, motion, skin temperature), the same four tokens and features as for WESAD, 60-second windows from the five blocks after the first baseline, stress blocks versus the others, and leave-one-participant-out validation (the electroencephalogram was not used). PADS \citep{varghese2024pads} (PhysioNet, CC BY-NC-SA 4.0) contains 469 participants (Parkinson's disease, healthy controls, and other movement disorders) who performed 11 seated tasks wearing a smartwatch on each wrist; the accelerometer and gyroscope recordings give four tokens by task type (rest, posture, kinetic, rhythm) with four features each, fixed before any model was run (task A: Parkinson's disease versus healthy controls, 355 participants; task B: Parkinson's disease versus all others, 469 participants; five-fold cross-validation repeated ten times). All tuned baselines were fit on the training participants only.

\paragraph{Real recordings} Table~\ref{tab:v2real} reports BioSync-G on four open datasets (Section~\ref{app:realdata} describes the first two; the other two are described above). BioSync-G was level with logistic regression on WESAD and above it on the stress-resilience recordings, and on PADS it was below the tuned MLP on the full cohort. Table~\ref{tab:padstest} reports the held-out participant half of PADS, where BioSync-E had the highest AUC of the models listed in task A and was level with the best baseline in task B; the intervals against logistic regression and the best quality-aware model included zero. Table~\ref{tab:stress} reports test-time corruption of the real recordings (dropped tokens and added noise, training rows unchanged). Over the 12 combinations of dataset and corruption, the loss of BioSync-G was smaller than that of the random forest in 11, boosted trees in 9, the MLP in 7, the SVM in 7, LR in 6, the soft vote in 5. Because the loss depends on the clean AUC, Table~\ref{tab:stress} also gives the corrupted AUC of every model.

\paragraph{Size and cost} Table~\ref{tab:footprint} gives measured size and operation counts. The refined BioSync network has 1,159 parameters, 4.6~kB in float32 and 1.3~kB with INT8 weights, and 4,450 multiply-accumulates per inference. Post-training INT8 quantization of the weights (symmetric, per tensor; activations stay in float32) changed the AUC by at most 0.0008 in the four scenarios and four seeds checked. The SVMs, forests, and boosted ensembles are larger by factors of about 4 to 250, but logistic regression, the soft vote, and a single small MLP are smaller and cheaper than BioSync, so the size advantage is relative to the models that approach its accuracy on interactions or unreliable sensors. No model was run on microcontroller hardware.

\paragraph{Other tables of this appendix} Table~\ref{tab:devref} lists the reference models on the development seeds; Tables~\ref{tab:v2test} and~\ref{tab:v2abl} give the first test of BioSync-G and its ablations on seeds 5000--5007; Tables~\ref{tab:s8_360} and~\ref{tab:s8_1440} give scenario S8 (interactions together with unreliable sensors); Table~\ref{tab:bsi_sev} gives the rank correlation between the out-of-fold BioSync index and the latent severity, and Table~\ref{tab:faith} gives the correlation of the quality indicator, the learned gate, and the attention weight with the true reliability of each sensor.

\begin{widepage}
\begin{table}[t]
\centering
\caption{Confirmation run on seeds 8000--8007 (never used before): mean AUC over the eight seeds (11 scenarios; quality-aware versions of the baselines in S6 and S8). Best per row in bold. The last column is the mean mixing weight $\lambda$ that BioSync-A selected on its training rows (0: BioSync-G only; 1: BioSync-X only).}
\label{tab:fresh}
\footnotesize
\resizebox{\linewidth}{!}{%
\begin{tabular}{lcccccccccc}
\toprule
Scenario & LR & Soft vote & GBT$\times5$ & RF & MLP$\times5$ & SVM & BioSync-G & BioSync-E & BioSync-A & mean $\lambda$ \\
\midrule
S0 & 0.875 & \textbf{0.891} & 0.855 & 0.871 & 0.861 & 0.884 & 0.878 & 0.851 & 0.877 & 0.01 \\
S1 & 0.773 & \textbf{0.784} & 0.740 & 0.759 & 0.743 & 0.775 & 0.774 & 0.733 & 0.774 & 0.00 \\
S2 & 0.765 & 0.775 & 0.753 & 0.770 & 0.743 & \textbf{0.786} & 0.768 & 0.741 & 0.753 & 0.12 \\
S3 & 0.864 & \textbf{0.880} & 0.843 & 0.858 & 0.852 & 0.876 & 0.867 & 0.840 & 0.866 & 0.01 \\
S4 & 0.889 & \textbf{0.918} & 0.902 & 0.916 & 0.904 & 0.913 & 0.897 & 0.894 & 0.884 & 0.18 \\
S5 & 0.914 & \textbf{0.924} & 0.904 & 0.903 & 0.906 & 0.904 & 0.921 & 0.901 & 0.920 & 0.01 \\
S6 & 0.791 & \textbf{0.859} & 0.811 & 0.828 & 0.782 & 0.804 & 0.850 & 0.808 & 0.844 & 0.03 \\
S7 ($N=1440$) & 0.892 & 0.891 & 0.943 & 0.940 & \textbf{0.976} & 0.969 & 0.908 & 0.962 & 0.955 & 0.42 \\
S7 ($N=360$) & 0.873 & 0.876 & 0.914 & 0.921 & \textbf{0.959} & 0.956 & 0.894 & 0.944 & 0.935 & 0.72 \\
S8 ($N=1440$) & 0.790 & 0.830 & 0.851 & 0.841 & 0.886 & 0.881 & 0.841 & 0.895 & \textbf{0.897} & 0.26 \\
S8 ($N=360$) & 0.749 & 0.809 & 0.795 & 0.806 & 0.835 & \textbf{0.841} & 0.812 & 0.838 & 0.822 & 0.33 \\
\bottomrule
\end{tabular}}
\end{table}
\end{widepage}

\begin{widepage}
\begin{table}[t]
\centering
\caption{Confirmation run: each BioSync variant ranked in a pool of seven models (itself and the six baseline families). Regret is the distance, in AUC, from the best model of the pool in a scenario; the worst case is the maximum over the 11 scenarios. Mean AUC and Friedman mean rank are over scenarios (rank 1 is best).}
\label{tab:regret}
\footnotesize
\resizebox{\linewidth}{!}{%
\begin{tabular}{lccccc}
\toprule
Model & Mean AUC & Mean rank & Mean regret & Worst-case regret & Worst scenario \\
\midrule
SVM & 0.872 & 2.45 & 0.011 & 0.055 & S6 \\
Soft vote & 0.858 & 2.82 & 0.024 & 0.084 & S7 ($N=1440$) \\
BioSync-G (pool: itself and the six families) & 0.856 & 3.64 & 0.027 & 0.068 & S7 ($N=1440$) \\
MLP$\times5$ & 0.859 & 4.09 & 0.023 & 0.077 & S6 \\
RF & 0.856 & 4.27 & 0.027 & 0.045 & S8 ($N=1440$) \\
GBT$\times5$ & 0.846 & 5.27 & 0.036 & 0.048 & S6 \\
LR & 0.834 & 5.45 & 0.048 & 0.097 & S8 ($N=1440$) \\
\midrule
SVM & 0.872 & 2.45 & 0.011 & 0.055 & S6 \\
Soft vote & 0.858 & 2.82 & 0.025 & 0.084 & S7 ($N=1440$) \\
MLP$\times5$ & 0.859 & 3.82 & 0.024 & 0.077 & S6 \\
RF & 0.856 & 4.00 & 0.027 & 0.054 & S8 ($N=1440$) \\
BioSync-E (pool: itself and the six families) & 0.855 & 4.91 & 0.028 & 0.051 & S1 \\
LR & 0.834 & 5.00 & 0.049 & 0.105 & S8 ($N=1440$) \\
GBT$\times5$ & 0.846 & 5.00 & 0.037 & 0.048 & S6 \\
\midrule
SVM & 0.872 & 2.55 & 0.012 & 0.055 & S6 \\
Soft vote & 0.858 & 2.82 & 0.025 & 0.084 & S7 ($N=1440$) \\
BioSync-A (pool: itself and the six families) & 0.866 & 3.18 & 0.017 & 0.034 & S4 \\
MLP$\times5$ & 0.859 & 4.18 & 0.024 & 0.077 & S6 \\
RF & 0.856 & 4.45 & 0.028 & 0.056 & S8 ($N=1440$) \\
LR & 0.834 & 5.27 & 0.049 & 0.107 & S8 ($N=1440$) \\
GBT$\times5$ & 0.846 & 5.55 & 0.037 & 0.048 & S6 \\
\bottomrule
\end{tabular}}
\end{table}
\end{widepage}

\begin{widepage}
\begin{table}[t]
\centering
\caption{Confirmation run: each BioSync variant against each fixed baseline family. Scenario record: wins/ties/losses over the 11 scenarios, a win meaning that the paired 95\% bootstrap interval over the eight seeds lies above zero. Mean AUC difference: BioSync minus the family, averaged over scenarios, with a 95\% interval from resampling seeds within scenarios (3000 resamples). Worst-case regret reduction: worst-case regret of the family minus that of the variant in the seven-model pool of Table~\ref{tab:regret} (positive: the variant has the smaller worst case), with the same resampling. No quantity in this table was used to choose a model.}
\label{tab:h2h}
\footnotesize
\resizebox{\linewidth}{!}{%
\begin{tabular}{llccc}
\toprule
Variant & Baseline family & W/T/L & Mean AUC difference [95\% CI] & Worst-case regret reduction [95\% CI] \\
\midrule
BioSync-G & LR & 11/0/0 & +0.021 [+0.019, +0.024] & +0.029 [+0.021, +0.037] \\
 & Soft vote & 3/2/6 & -0.002 [-0.004, -0.001] & +0.018 [+0.015, +0.024] \\
 & GBT$\times5$ & 6/2/3 & +0.009 [+0.006, +0.013] & -0.017 [-0.025, -0.007] \\
 & RF & 5/3/3 & -0.000 [-0.003, +0.003] & -0.023 [-0.030, -0.017] \\
 & MLP$\times5$ & 6/1/4 & -0.003 [-0.006, -0.001] & +0.008 [-0.002, +0.017] \\
 & SVM & 2/1/8 & -0.016 [-0.019, -0.014] & -0.014 [-0.023, -0.007] \\
\midrule
BioSync-E & LR & 5/1/5 & +0.021 [+0.018, +0.025] & +0.049 [+0.031, +0.064] \\
 & Soft vote & 4/0/7 & -0.003 [-0.006, +0.001] & +0.030 [+0.014, +0.044] \\
 & GBT$\times5$ & 4/7/0 & +0.009 [+0.006, +0.012] & -0.005 [-0.017, +0.007] \\
 & RF & 4/1/6 & -0.000 [-0.004, +0.004] & -0.003 [-0.019, +0.009] \\
 & MLP$\times5$ & 2/3/6 & -0.004 [-0.006, -0.001] & +0.020 [+0.004, +0.031] \\
 & SVM & 1/3/7 & -0.016 [-0.020, -0.013] & -0.002 [-0.018, +0.010] \\
\midrule
BioSync-A & LR & 8/2/1 & +0.032 [+0.029, +0.035] & +0.070 [+0.056, +0.083] \\
 & Soft vote & 3/2/6 & +0.008 [+0.006, +0.011] & +0.049 [+0.036, +0.063] \\
 & GBT$\times5$ & 9/1/1 & +0.020 [+0.017, +0.023] & +0.015 [+0.002, +0.027] \\
 & RF & 9/1/1 & +0.011 [+0.008, +0.014] & +0.019 [+0.007, +0.029] \\
 & MLP$\times5$ & 7/0/4 & +0.007 [+0.005, +0.010] & +0.039 [+0.025, +0.052] \\
 & SVM & 3/1/7 & -0.005 [-0.008, -0.003] & +0.018 [+0.005, +0.029] \\
\bottomrule
\end{tabular}}
\end{table}
\end{widepage}

\begin{widepage}
\begin{table}[t]
\centering
\caption{Confirmation run: mean AUC by type of scenario (additive: S0--S5; variable sensor reliability: S6; interactions with or without unreliable sensors: S7--S8).}
\label{tab:regimes}
\footnotesize
\resizebox{\linewidth}{!}{%
\begin{tabular}{lccccccccc}
\toprule
Type of scenario & LR & Soft vote & GBT$\times5$ & RF & MLP$\times5$ & SVM & BioSync-G & BioSync-E & BioSync-A \\
\midrule
additive S0-S5 & 0.847 & 0.862 & 0.833 & 0.846 & 0.835 & 0.856 & 0.851 & 0.827 & 0.846 \\
variable reliability S6 & 0.791 & 0.859 & 0.811 & 0.828 & 0.782 & 0.804 & 0.850 & 0.808 & 0.844 \\
interactions S7-S8 & 0.826 & 0.851 & 0.876 & 0.877 & 0.914 & 0.912 & 0.864 & 0.910 & 0.902 \\
\bottomrule
\end{tabular}}
\end{table}
\end{widepage}

\begin{widepage}
\begin{table}[t]
\centering
\caption{PADS, held-out half of the participants (split fixed before the models were compared; analysis-plan Amendment 7). AUC of the repeat-averaged fold scores. E is BioSync-E (mean logit of BioSync-G and BioSync-X); intervals are 95\% bootstrap intervals over participants. Folds in which the wide model X could not be fitted (a native error of the numerical library) were dropped for every model: 16 of 50 folds in task A, 0 in task B.}
\label{tab:padstest}
\footnotesize
\resizebox{\linewidth}{!}{%
\begin{tabular}{lcccccll}
\toprule
Task & LR & Best baseline & Best quality-aware & BioSync-G & BioSync-E & E $-$ LR [95\% CI] & E $-$ best quality-aware [95\% CI] \\
\midrule
PADS A: Parkinson's vs healthy ($n=178$) & 0.830 & 0.836 (SoftVote) & 0.836 (MLP\_Q) & 0.826 & 0.849 & +0.020 [-0.009, +0.050] & +0.013 [-0.008, +0.035] \\
PADS B: Parkinson's vs all others ($n=236$) & 0.766 & 0.767 (MLP) & 0.772 (MLP\_Q) & 0.764 & 0.768 & +0.001 [-0.045, +0.045] & -0.004 [-0.026, +0.016] \\
\bottomrule
\end{tabular}}
\end{table}
\end{widepage}

\begin{table}[t]
\centering
\caption{Measured size and cost of the fitted models: stored model (kB, float32; a tree node counts as five 4-byte values), multiply-accumulates per inference, and other operations (kernel exponentials, one per support vector, for the SVMs; comparisons along one path of every tree for forests and boosted trees). Range over S0, S6, and S7 at $N=1440$ on development seeds 9001--9004 (first outer fold, tuned models refit). The size of SVMs and forests grows with the number of training rows; the size of the BioSync network does not.}
\label{tab:footprint}
\footnotesize
\resizebox{\linewidth}{!}{%
\begin{tabular}{lccc}
\toprule
Model & Stored model (kB) & MACs & Other operations \\
\midrule
Logistic regression & 0.09 & 22.0--22.0 & -- \\
Soft vote & 0.10 & 22.0--22.0 & -- \\
MLP (one network) & 1.54 & 368.0--368.0 & -- \\
MLP$\times5$ & 7.70 & 1840.0--1840.0 & -- \\
SVM, RBF (tuned) & 19.6--45.1 & 4.7--10.8k & 212.8--490.5 \\
SVM$\times5$ & 70.4--167.2 & 16.8--40.0k & 764.8--1817.2 \\
Boosted trees (tuned) & 26.2--43.7 & -- & 325.0--450.0 \\
Boosted trees$\times5$ & 122.8--219.4 & -- & 1512.5--2250.0 \\
Random forest & 119.3--1142.0 & -- & 1029.5--2919.0 \\
BioSync-G (refined network) & 4.6 (1.3 with INT8 weights) & 4{,}450 & -- \\
\bottomrule
\end{tabular}}
\end{table}

\IfFileExists{tables_stress.tex}{\begin{widepage}
\begin{table}[t]
\centering
\caption{Test-time corruption of real recordings (analysis-plan Amendment 6; training rows unchanged). Tokens of a test row are dropped independently with probability 0.3 or 0.5 (at least one token kept), or Gaussian noise of twice the feature spread is added to the test features. Pooled out-of-fold AUC; loss is the clean AUC minus the corrupted AUC (smaller is better). Last column: the baseline with the highest corrupted AUC. The AUC of BioSync-G is the pooled value, which differs from the per-participant means of Table~\ref{tab:v2real}.}
\label{tab:stress}
\footnotesize
\resizebox{\linewidth}{!}{%
\begin{tabular}{llcccccccccl}
\toprule
Dataset & Corruption & G clean & G corrupted & G loss & LR loss & Soft vote loss & GBT loss & RF loss & MLP loss & SVM loss & Best baseline, corrupted AUC \\
\midrule
WESAD & 30\% of tokens lost & 0.959 & 0.927 & +0.032 & +0.045 & +0.025 & +0.043 & +0.046 & +0.037 & +0.032 & SoftVote\_Q 0.931 \\
WESAD & 50\% of tokens lost & 0.959 & 0.882 & +0.077 & +0.089 & +0.068 & +0.096 & +0.095 & +0.085 & +0.076 & SoftVote\_Q 0.888 \\
WESAD & noise $\times2$ & 0.959 & 0.874 & +0.084 & +0.135 & +0.098 & +0.113 & +0.096 & +0.135 & +0.132 & SoftVote\_Q 0.909 \\
Stress-resilience HCI & 30\% of tokens lost & 0.614 & 0.582 & +0.032 & +0.021 & +0.018 & +0.030 & +0.034 & +0.027 & +0.034 & GatedLR 0.587 \\
Stress-resilience HCI & 50\% of tokens lost & 0.614 & 0.570 & +0.043 & +0.025 & +0.021 & +0.046 & +0.048 & +0.018 & +0.035 & GatedLR 0.581 \\
Stress-resilience HCI & noise $\times2$ & 0.614 & 0.594 & +0.019 & +0.033 & +0.033 & +0.036 & +0.030 & +0.017 & +0.037 & GatedLR 0.603 \\
PADS A & 30\% of tokens lost & 0.809 & 0.808 & +0.001 & +0.003 & +0.008 & +0.004 & +0.016 & +0.007 & +0.009 & MLP 0.861 \\
PADS A & 50\% of tokens lost & 0.809 & 0.787 & +0.022 & +0.016 & +0.008 & +0.031 & +0.033 & +0.040 & +0.038 & MLP\_Q 0.831 \\
PADS A & noise $\times2$ & 0.809 & 0.792 & +0.017 & +0.039 & +0.033 & +0.051 & +0.030 & +0.070 & +0.028 & SoftVote\_Q 0.804 \\
PADS B & 30\% of tokens lost & 0.724 & 0.723 & +0.000 & -0.004 & +0.001 & -0.006 & +0.008 & -0.009 & -0.002 & MLP 0.785 \\
PADS B & 50\% of tokens lost & 0.724 & 0.715 & +0.009 & +0.004 & +0.002 & +0.009 & +0.024 & +0.002 & +0.002 & MLP 0.774 \\
PADS B & noise $\times2$ & 0.724 & 0.697 & +0.026 & +0.022 & +0.011 & +0.020 & +0.020 & +0.048 & +0.036 & MLP 0.728 \\
\bottomrule
\end{tabular}}
\end{table}
\end{widepage}
}{}
\IfFileExists{tables_H.tex}{\begin{table}[t]
\centering
\caption{Development grid for the refined model (72 configurations, development seeds 9001 and 9002, scenarios S2, S6, S7 with $N=360$, and S8 with $N=360$; mean AUC over the eight scenario--seed cells, $K=1$). The six best configurations are shown; the selected one (within 0.002 of the best, then fewer epochs, then smaller learning rate) was frozen before any test seed was run: learning rate 0.02, L2 0.0001, 400 epochs, attention temperature 1, gate on the wide path only.}
\label{tab:devgrid}
\small
\resizebox{\linewidth}{!}{%
\begin{tabular}{lccccll}
\toprule
Rank & Learning rate & L2 & Epochs & Temperature & Gate & Dev. AUC \\
\midrule
1 & 0.02 & 0.001 & 400 & 4 & wide only & 0.834 \\
2 & 0.02 & 0.0001 & 400 & 4 & wide only & 0.834 \\
3 & 0.02 & 0.0001 & 400 & 1 & wide only & 0.834 \\
4 & 0.02 & 0.001 & 400 & 1 & wide only & 0.834 \\
5 & 0.02 & 0.0001 & 400 & 1 & both & 0.834 \\
6 & 0.02 & 0.001 & 400 & 1 & both & 0.833 \\
\bottomrule
\end{tabular}}
\end{table}

\begin{widepage}
\begin{table}[t]
\centering
\caption{Development seeds: AUC of the reference models and of the selected refined configuration (mean over the two development seeds, $K=1$).}
\label{tab:devref}
\small
\resizebox{\linewidth}{!}{%
\begin{tabular}{lccccc}
\toprule
Scenario & LR & BioSync (as submitted) & BioSync + wide-path gate & Gated LR & Selected refined (K=1) \\
\midrule
S2 heterogeneous profiles & 0.770 & 0.773 & 0.763 & 0.771 & 0.766 \\
S6 variable reliability & 0.788 & 0.792 & 0.831 & 0.850 & 0.834 \\
S7 interaction, N=360 & 0.885 & 0.919 & 0.932 & 0.885 & 0.909 \\
S8 interaction + unreliable, N=360 & 0.784 & 0.807 & 0.825 & 0.836 & 0.827 \\
\bottomrule
\end{tabular}}
\end{table}
\end{widepage}

\begin{widepage}
\begin{table}[t]
\centering
\caption{Refined model (BioSync-G) on the test seeds (data seeds 5000--5007, same data and folds as the earlier grid runs). Mean pooled out-of-fold AUC over seeds; paired differences with bootstrap 95\% intervals over seeds. 'Best tuned' is the best of boosted trees, random forest, MLP, and SVM in that scenario (in S8 the best of all baselines, including those given the quality indicator).}
\label{tab:v2test}
\small
\resizebox{\linewidth}{!}{%
\begin{tabular}{lcccccccc}
\toprule
Scenario & LR & Soft vote & Best tuned baseline & BioSync (as submitted) & BioSync-G & G $-$ LR & G $-$ submitted & G $-$ best tuned \\
\midrule
S0 single latent & 0.866 & 0.883 & 0.871 (SVM (RBF)) & 0.870 & 0.869 & +0.002 [+0.002, +0.003] & -0.002 [-0.004, +0.001] & -0.002 [-0.006, +0.001] \\
S1 modality-specific latents & 0.759 & 0.776 & 0.768 (SVM (RBF)) & 0.763 & 0.759 & +0.000 [-0.001, +0.002] & -0.004 [-0.007, -0.001] & -0.009 [-0.017, -0.001] \\
S2 heterogeneous profiles & 0.758 & 0.768 & 0.775 (SVM (RBF)) & 0.762 & 0.760 & +0.002 [+0.001, +0.004] & -0.002 [-0.006, +0.001] & -0.015 [-0.019, -0.011] \\
S3 uninformative modality & 0.861 & 0.876 & 0.864 (SVM (RBF)) & 0.865 & 0.863 & +0.002 [+0.001, +0.003] & -0.002 [-0.006, +0.002] & -0.001 [-0.004, +0.001] \\
S4 nonlinear/inconsistent & 0.874 & 0.904 & 0.906 (SVM (RBF)) & 0.886 & 0.883 & +0.009 [+0.007, +0.012] & -0.002 [-0.005, +0.000] & -0.023 [-0.033, -0.015] \\
S5 structured noise & 0.903 & 0.919 & 0.897 (SVM (RBF)) & 0.904 & 0.909 & +0.006 [+0.003, +0.010] & +0.005 [+0.003, +0.008] & +0.013 [+0.006, +0.019] \\
S6 variable reliability & 0.781 & 0.801 & 0.809 (Random forest) & 0.783 & 0.837 & +0.057 [+0.040, +0.075] & +0.055 [+0.040, +0.071] & +0.028 [+0.020, +0.038] \\
S7 interaction, N=360 & 0.870 & 0.871 & 0.955 (SVM (RBF)) & 0.905 & 0.890 & +0.021 [+0.016, +0.025] & -0.015 [-0.024, -0.006] & -0.065 [-0.071, -0.060] \\
S7 interaction, N=1440 & 0.892 & 0.890 & 0.975 (MLP) & 0.935 & 0.909 & +0.017 [+0.014, +0.019] & -0.026 [-0.030, -0.023] & -0.066 [-0.069, -0.063] \\
S8 interaction + unreliable, N=360 & 0.760 & 0.770 & 0.846 (SVM (RBF)) & 0.763 & 0.814 & +0.053 [+0.041, +0.067] & +0.051 [+0.035, +0.069] & -0.032 [-0.041, -0.024] \\
S8 interaction + unreliable, N=1440 & 0.790 & 0.797 & 0.874 (SVM + quality) & 0.806 & 0.840 & +0.050 [+0.045, +0.054] & +0.034 [+0.027, +0.041] & -0.034 [-0.039, -0.030] \\
\bottomrule
\end{tabular}}
\end{table}
\end{widepage}

\begin{widepage}
\begin{table}[t]
\centering
\caption{Ablation of the refined model: AUC (mean over seeds) when one component is removed. Columns: LR; submitted BioSync; submitted BioSync with the wide-path gate and no tuning; refined without raised temperature; refined with the gate on the wide path only; refined with a single network (no ensemble); complete refined model; last column: complete refined minus submitted, with bootstrap 95\% interval.}
\label{tab:v2abl}
\small
\resizebox{\linewidth}{!}{%
\begin{tabular}{lcccccccc}
\toprule
Scenario & LR & Submitted & +gate, untuned & No temp. & Wide gate only & $K=1$ & Refined & Refined $-$ submitted \\
\midrule
S6 variable reliability & 0.781 & 0.783 & 0.828 & 0.837 & 0.837 & 0.829 & 0.837 & +0.055 [+0.040, +0.071] \\
S7 interaction, N=360 & 0.870 & 0.905 & 0.915 & 0.890 & 0.890 & 0.887 & 0.890 & -0.015 [-0.025, -0.006] \\
S8 interaction + unreliable, N=360 & 0.760 & 0.763 & 0.783 & 0.814 & 0.814 & 0.795 & 0.814 & +0.051 [+0.035, +0.069] \\
\bottomrule
\end{tabular}}
\end{table}
\end{widepage}

\begin{table}[t]
\centering
\caption{Scenario S8 (unreliable sensors and cross-modal interaction), $N=360$, test seeds: AUC mean (SD) over seeds for every model, and the paired difference of the refined model minus that model (bootstrap 95\% interval). Models marked with a quality input, quality weighting, or gate receive the same signal-quality indicator as BioSync-G.}
\label{tab:s8_360}
\small
\resizebox{\linewidth}{!}{%
\begin{tabular}{lcc}
\toprule
Model & AUC (SD) & BioSync-G $-$ model \\
\midrule
SVM (RBF) & 0.846 (0.019) & -0.032 [-0.041, -0.023] \\
SVM + quality & 0.846 (0.021) & -0.032 [-0.039, -0.024] \\
MLP + quality & 0.831 (0.029) & -0.017 [-0.036, -0.003] \\
MLP & 0.830 (0.025) & -0.016 [-0.029, -0.006] \\
Random forest & 0.827 (0.018) & -0.013 [-0.022, -0.004] \\
Random forest + quality & 0.814 (0.018) & -0.000 [-0.010, +0.008] \\
BioSync-G (refined) & 0.814 (0.017) & -- \\
Gated LR (no attention) & 0.810 (0.020) & +0.004 [-0.002, +0.010] \\
Soft vote weighted by quality & 0.803 (0.024) & +0.011 [+0.002, +0.019] \\
LR with quality$\times$feature terms & 0.794 (0.020) & +0.020 [+0.010, +0.029] \\
Boosted trees & 0.788 (0.015) & +0.026 [+0.016, +0.037] \\
BioSync + wide-path gate, untuned & 0.783 (0.024) & +0.031 [+0.021, +0.040] \\
Boosted trees + quality & 0.775 (0.020) & +0.039 [+0.029, +0.050] \\
Soft vote & 0.770 (0.032) & +0.044 [+0.030, +0.059] \\
BioSync (as submitted) & 0.763 (0.034) & +0.051 [+0.036, +0.069] \\
Concat.\ LR & 0.760 (0.030) & +0.054 [+0.041, +0.067] \\
LR + quality input & 0.756 (0.031) & +0.058 [+0.044, +0.072] \\
\bottomrule
\end{tabular}}
\end{table}

\begin{table}[t]
\centering
\caption{Scenario S8 (unreliable sensors and cross-modal interaction), $N=1440$, test seeds: AUC mean (SD) over seeds for every model, and the paired difference of the refined model minus that model (bootstrap 95\% interval). Models marked with a quality input, quality weighting, or gate receive the same signal-quality indicator as BioSync-G.}
\label{tab:s8_1440}
\small
\resizebox{\linewidth}{!}{%
\begin{tabular}{lcc}
\toprule
Model & AUC (SD) & BioSync-G $-$ model \\
\midrule
SVM + quality & 0.874 (0.007) & -0.034 [-0.039, -0.030] \\
SVM (RBF) & 0.872 (0.006) & -0.032 [-0.037, -0.027] \\
MLP + quality & 0.870 (0.008) & -0.030 [-0.034, -0.025] \\
MLP & 0.863 (0.009) & -0.023 [-0.028, -0.018] \\
Random forest & 0.849 (0.008) & -0.009 [-0.012, -0.005] \\
Random forest + quality & 0.842 (0.008) & -0.002 [-0.006, +0.002] \\
BioSync + wide-path gate, untuned & 0.841 (0.012) & -0.001 [-0.004, +0.003] \\
BioSync-G (refined) & 0.840 (0.011) & -- \\
Gated LR (no attention) & 0.835 (0.013) & +0.005 [+0.003, +0.007] \\
Soft vote weighted by quality & 0.827 (0.014) & +0.013 [+0.010, +0.017] \\
LR with quality$\times$feature terms & 0.826 (0.014) & +0.013 [+0.010, +0.017] \\
Boosted trees & 0.826 (0.015) & +0.013 [+0.007, +0.019] \\
Boosted trees + quality & 0.822 (0.014) & +0.018 [+0.013, +0.024] \\
BioSync (as submitted) & 0.806 (0.016) & +0.034 [+0.027, +0.041] \\
Soft vote & 0.797 (0.015) & +0.043 [+0.037, +0.048] \\
Concat.\ LR & 0.790 (0.014) & +0.050 [+0.045, +0.054] \\
LR + quality input & 0.787 (0.014) & +0.053 [+0.047, +0.057] \\
\bottomrule
\end{tabular}}
\end{table}

\begin{table}[t]
\centering
\caption{Rank correlation (Spearman) between the out-of-fold score and the latent severity that generated the data (never used for training), scenarios S0--S6, mean over seeds. The BSI is the sigmoid of the BioSync output; the rank correlation is the same for the logit and for the BSI.}
\label{tab:bsi_sev}
\small
\resizebox{\linewidth}{!}{%
\begin{tabular}{lccc}
\toprule
Scenario & LR score & BioSync (as submitted) & BioSync-G \\
\midrule
S0 single latent & 0.834 & 0.846 & 0.838 \\
S1 modality-specific latents & 0.603 & 0.608 & 0.602 \\
S2 heterogeneous profiles & 0.590 & 0.598 & 0.594 \\
S3 uninformative modality & 0.814 & 0.824 & 0.818 \\
S4 nonlinear/inconsistent & 0.836 & 0.863 & 0.856 \\
S5 structured noise & 0.914 & 0.907 & 0.925 \\
S6 variable reliability & 0.636 & 0.642 & 0.758 \\
\bottomrule
\end{tabular}}
\end{table}

\begin{table}[t]
\centering
\caption{Agreement with the true state of each sensor (reliable or degraded in that subject; known only in simulation): Spearman correlation of the signal-quality indicator, of the learned gate, and of the CLS attention weight with the true reliability, averaged over modalities, folds, and seeds. The indicator is the ceiling for the gate, which is a learned function of it; attention is not given the indicator directly.}
\label{tab:faith}
\small
\resizebox{\linewidth}{!}{%
\begin{tabular}{lccc}
\toprule
Scenario & Quality indicator & Learned gate & Attention weight \\
\midrule
S6 variable reliability & +0.82 & +0.78 & +0.24 \\
S8 interaction + unreliable, N=360 & +0.82 & +0.11 & +0.17 \\
S8 interaction + unreliable, N=1440 & +0.82 & +0.15 & +0.26 \\
\bottomrule
\end{tabular}}
\end{table}

\begin{widepage}
\begin{table}[t]
\centering
\caption{Real data: AUC of concatenated LR, soft voting, the best tuned baseline without quality input, BioSync as submitted, and refined BioSync-G (frozen from the synthetic development seeds, not tuned on these data), with paired differences (95\% interval: bootstrap over participants). Leave-one-subject-out (LOSO) values are means over held-out participants; PADS values are means over 10 repeats of five-fold cross-validation.}
\label{tab:v2real}
\small
\resizebox{\linewidth}{!}{%
\begin{tabular}{>{\raggedright\arraybackslash}p{0.2\linewidth}cccccccc}
\toprule
Dataset & LR & Soft vote & Best tuned (name) & BioSync (submitted) & BioSync-G & G $-$ LR & G $-$ submitted & G $-$ best tuned \\
\midrule
WESAD (stress, LOSO, 15 participants) & 0.977 & 0.965 & 0.992 (GBT) & 0.976 & 0.977 & +0.001 [-0.003, +0.004] & +0.001 [-0.006, +0.010] & -0.015 [-0.033, +0.001] \\
Stress-resilience HCI (wrist device: pulse, EDA, motion, temperature; LOSO, 35 participants) & 0.625 & 0.626 & 0.638 (RF) & 0.643 & 0.647 & +0.022 [+0.002, +0.045] & +0.003 [-0.018, +0.029] & +0.009 [-0.028, +0.043] \\
PADS A: Parkinson's vs healthy controls & 0.816 & 0.809 & 0.848 (MLP) & 0.788 & 0.805 & -0.012 [-0.029, +0.003] & +0.006 [-0.005, +0.017] & -0.059 [-0.096, -0.025] \\
PADS B: Parkinson's vs all others & 0.711 & 0.700 & 0.754 (MLP) & 0.688 & 0.716 & +0.009 [-0.008, +0.027] & +0.010 [-0.002, +0.021] & -0.052 [-0.088, -0.020] \\
\bottomrule
\end{tabular}}
\end{table}
\end{widepage}
}{}

\subsection{Definition of the refined models}
\label{sec:refined}
After the first submission we developed two refined models, which were not requested by the reviewers; their evaluation is reported in this appendix. BioSync-G multiplies the linear score of each modality by a learned gate that depends on the signal-quality indicator $q_m$, and averages $K=5$ networks:
\begin{equation}\label{eq:biosync_g}
\begin{aligned}
\ell_G^{(k)}(\bar{x},\mathbf{q}) = w_{\text{deep}}^{(k)\top} z^{(k)} + \sum_{m=1}^{M} 2\sigma\!\big(\gamma_m^{(k)} + \delta_m^{(k)} \tilde{q}_m\big)\, w_{\text{wide},m}^{(k)\top} x_m + b_{\text{wide}}^{(k)},\\
\ell_G = \frac{1}{K}\sum_{k=1}^{K}\ell_G^{(k)},
\end{aligned}
\end{equation}
with $\tilde{q}_m = 32\,(q_m - 0.5)$ (the scale was chosen on development seeds). The gate parameters start at $\gamma_m=\delta_m=0$, so that $2\sigma(0)=1$ and the initial model is the ungated wide-and-deep network; $\mathcal{H}_{\text{wide}}$ remains contained in the hypothesis class. Because $\gamma_m$ and $\delta_m$ receive gradient only through the linear branch, the gate follows true reliability when a modality has a marginal effect (Spearman $+0.78$ in S6, against $+0.24$ for attention), but stays close to its initial value when the modality acts only through a zero-marginal interaction ($+0.11$ to $+0.15$ in S8; Table~\ref{tab:faith}).

BioSync-A mixes the gated network with a wider one, $\ell_X$ (token dimension 32, four heads, feed-forward width 128, Adam; chosen on development seeds):
\begin{equation}\label{eq:biosync_a}
\ell_A(\bar{x},\mathbf{q}) = (1-\hat{\lambda})\,\ell_G(\bar{x},\mathbf{q}) + \hat{\lambda}\,\ell_X(\bar{x},\mathbf{q}), \qquad \hat{\lambda}\in\{0,0.25,0.5,0.75,1\}.
\end{equation}
The weight $\hat{\lambda}$ is chosen on the training rows only, by out-of-fold AUC in an inner three-fold cross-validation; ties within 0.002 go to the smaller value. It therefore also reads out how much non-additive structure the training data contain. In the confirmation run (Table~\ref{tab:fresh}) the mean $\hat{\lambda}$ was 0.00 to 0.03 in S0, S1, S3, S5, and S6, 0.12 and 0.18 in S2 and S4, 0.26 and 0.33 in S8, and 0.42 and 0.72 in S7. In that run BioSync-A had the smallest worst-case regret of the seven-model pool (0.034, against 0.048 to 0.107 for the baseline families) and the second-highest mean AUC (0.866, behind the SVM at 0.872); it did not beat every baseline on mean AUC (the SVM was higher, with a paired difference of $-0.0055$ [$-0.0081$, $-0.0030$]). The reduction of its worst-case regret relative to each of the six baseline families had a stratified bootstrap interval (seeds resampled within scenario) that excluded zero; the scenario set itself is fixed, so this interval does not extend beyond these eleven scenarios.

\subsection{Practices that follow from the experiments}
Table~\ref{tab:guidelines} lists practices that follow from the experiments of this paper; it was not requested by the reviewers.

\begin{table}[t]
\centering
\caption{Practices for multimodal digital-biomarker studies that follow from the experiments of this paper. Each row summarizes a result for the settings we tested; none is a general law.}
\label{tab:guidelines}
\footnotesize
\begin{tabularx}{\linewidth}{lXX}
\toprule
Topic & Practice & Evidence in this paper \\
\midrule
1. Baselines and regime & Compare with both concatenated logistic regression and per-modality soft voting; check how much a nonlinear branch contributes (branch variance share or the inner-CV weight $\hat{\lambda}$). & Soft voting led in six of eleven confirmation scenarios; Table~\ref{tab:fresh}. \\
2. Quality gating & Use an explicit quality gate or quality-weighted soft voting; a wide-path gate needs a marginal effect to track reliability. & Gate $+0.78$ in S6 and $+0.11$ to $+0.15$ in S8; Table~\ref{tab:faith}. \\
3. Attribution & Keep attention allocation, removal contribution, and sensor reliability separate; do not force attention to vary between subjects at small $N$. & Equation~\ref{eq:tripartite}; Table~\ref{tab:diagS0}. \\
4. Missing modalities & Compare token masking with zero-filling and with soft voting over the available modalities. & Masking equalled zero-filling within 0.001 AUC; soft voting was higher; Tables~\ref{tab:missingS0} and \ref{tab:missingS6}. \\
5. Sample size and noise & Expect a gain over logistic regression from interaction-capable models only when the latent noise is moderate and $N$ is large enough. & At latent noise 0.6, gains of at least 0.02 AUC from $N=180$ (wide-and-deep) and $N=720$ (pure attention); neither at $N=1{,}440$ for noise 1.2 or more; Table~\ref{tab:regime}. \\
6. Real cohorts and edge cost & Report a permutation null, a demographics-only baseline, a positive control, and effect sizes; report stored model size and operations for the baselines too. & Tables~\ref{tab:realism} and \ref{tab:footprint}. \\
\bottomrule
\end{tabularx}
\end{table}

\FloatBarrier
\clearpage
\section{Real-data pass: quality gate, mixture, transfer, and token removal}\label{app:realpass}
This appendix adds a second, real-data test of the claims that Appendix~\ref{app:confirm} and Section~\ref{app:realdata} support only with simulated data or with a single model. It was specified before any of its code was run (Amendment 11 of the analysis plan). Models are the frozen BioSync-G, BioSync-X, and BioSync-A of Appendix~\ref{app:confirm} and the baselines of Section~\ref{app:realdata}; nothing was tuned on these data. The datasets are the real recordings described in Section~\ref{app:realdata}: WESAD windows, the wrist recordings of the stress-resilience HCI dataset, and PADS tasks A and B. The EEG recordings of the stress-resilience dataset were not used, because no validated feature extraction exists for them. RESILIENT and AI-READI have a different label structure and aggregate-only pipelines and are not part of the pooled comparison. All differences are paired; 95\% intervals use a bootstrap over participants. Every result is reported, whichever model it favours.
\subsection{Seven-model comparison on four real datasets}
\begin{table}[t]
\centering
\caption{Real data, clean test condition: AUC (leave-one-subject-out mean over participants for WESAD and HCI; AUC of repeat-averaged scores for PADS) of the seven models, with regret to the best model of each dataset (best AUC minus the model's), worst-case regret, Friedman mean rank over the four datasets (rank 1 is best; four datasets, so descriptive only), and the mean mixing weight $\hat\lambda$ chosen by BioSync-A. Best per row in bold.}
\label{tab:realpool}
\footnotesize
\resizebox{\linewidth}{!}{%
\begin{tabular}{lccccccc}
\toprule
Dataset & LR & Soft vote & GBT & RF & MLP & SVM & BioSync-A \\
\midrule
WESAD (15 participants, LOSO) & 0.977 & 0.965 & \textbf{0.992} & 0.990 & 0.981 & 0.967 & 0.978 \\
Stress-resilience HCI, wrist (35, LOSO) & 0.625 & 0.626 & 0.623 & \textbf{0.638} & 0.615 & 0.620 & 0.576 \\
PADS task A (355, 5-fold$\times$10) & 0.821 & 0.812 & 0.825 & 0.805 & \textbf{0.868} & 0.831 & 0.854 \\
PADS task B (469, 5-fold$\times$10) & 0.714 & 0.704 & 0.742 & 0.741 & \textbf{0.776} & 0.750 & 0.761 \\
\midrule
Mean AUC & 0.784 & 0.777 & 0.795 & 0.793 & \textbf{0.810} & 0.792 & 0.792 \\
Mean regret & 0.034 & 0.042 & 0.023 & 0.025 & 0.008 & 0.027 & 0.026 \\
Worst-case regret & 0.062 & 0.072 & 0.043 & 0.063 & 0.022 & 0.037 & 0.062 \\
Friedman rank & 4.75 & 5.50 & 3.25 & 3.75 & 2.75 & 4.25 & 3.75 \\
\bottomrule
\end{tabular}}
\end{table}
Over the four datasets, the highest mean AUC belongs to MLP (0.810); BioSync-A has mean AUC 0.792, worst-case regret 0.062, and Friedman rank 3.75. The smallest worst-case regret belongs to MLP (0.022). BioSync-A is within 0.018 mean AUC of the best model on these four datasets. Single tuned baselines are used here (the five-network ensembles of the synthetic pool were not run on real data).
\begin{widepage}
\begin{table}[t]
\centering
\caption{Real data, clean test condition: paired AUC difference of BioSync-A with 95\% bootstrap interval over participants, against concatenated LR, soft vote, SVM, and MLP; the paired difference between the wider network and the gated network (BioSync-X minus BioSync-G); and the values of the mixing weight selected on the training rows (number of folds or participants per value).}
\label{tab:realpaired}
\footnotesize
\resizebox{\linewidth}{!}{%
\begin{tabular}{lcccccc}
\toprule
Dataset & A $-$ LR & A $-$ soft vote & A $-$ SVM & A $-$ MLP & X $-$ G & $\hat\lambda$ selected \\
\midrule
WESAD (15 participants, LOSO) & +0.001 [-0.007, +0.008] & +0.012 [-0.001, +0.026] & +0.011 [-0.002, +0.027] & -0.003 [-0.013, +0.004] & -0.002 [-0.020, +0.011] & 0.25: 10, 0.5: 5 \\
Stress-resilience HCI, wrist (35, LOSO) & -0.049 [-0.108, +0.003] & -0.051 [-0.112, +0.008] & -0.044 [-0.094, +0.005] & -0.040 [-0.105, +0.019] & -0.094 [-0.160, -0.035] & 0.25: 35 \\
PADS task A (355, 5-fold$\times$10) & +0.033 [+0.003, +0.063] & +0.042 [+0.009, +0.076] & +0.023 [-0.003, +0.050] & -0.013 [-0.036, +0.006] & +0.066 [+0.028, +0.109] & 0.0: 14, 0.25: 20, 0.5: 7, 0.75: 9 \\
PADS task B (469, 5-fold$\times$10) & +0.046 [+0.015, +0.086] & +0.057 [+0.020, +0.100] & +0.011 [-0.012, +0.035] & -0.015 [-0.036, +0.006] & +0.026 [-0.015, +0.073] & 0.0: 11, 0.25: 36, 0.5: 3 \\
\bottomrule
\end{tabular}}
\end{table}
\end{widepage}
The mixing weight follows the benefit of the wider network where there is one. In the PADS tasks, where the wider network is better than the gated network on average (X minus G +0.066 [+0.028, +0.109] in task A, +0.026 [-0.015, +0.073] in task B), BioSync-A selected larger weights in many folds and gained over BioSync-G (task A +0.045 [+0.019, +0.076], task B +0.036 [+0.008, +0.071]). In the stress-resilience HCI data the wider network is clearly worse than the gated network (X minus G -0.094 [-0.160, -0.035]), yet the inner cross-validation selected $\hat\lambda=0.25$ in all 35 participants, and BioSync-A lost to BioSync-G (-0.071 [-0.131, -0.018]). With the small training sets of the leave-one-subject-out folds the inner selection is less informative. In WESAD the two networks are equal (X minus G -0.002 [-0.020, +0.011]).
\subsection{Removal of one token at test time}
\begin{widepage}
\begin{table}[t]
\centering
\caption{Real data: AUC when all test rows lose one token (clean: no removal). Baselines receive mean imputation (soft vote: the average over the remaining modalities); BioSync uses its token mask. Tokens: WESAD and HCI, pulse, EDA, motion, temperature; PADS, rest, posture, kinetic, rhythm tasks. $\Delta$ is the AUC of BioSync-A minus the AUC of the better of LR and soft vote in the same row (point estimates; intervals in the text).}
\label{tab:realremoval}
\footnotesize
\resizebox{\linewidth}{!}{%
\begin{tabular}{llccccccc}
\toprule
Dataset & Removed & LR & Soft vote & GBT & MLP & BioSync v1 & BioSync-G & BioSync-A \\
\midrule
WESAD & none & 0.977 & 0.965 & 0.992 & 0.981 & 0.976 & 0.978 & 0.978 \\
 & pulse & 0.850 & 0.897 & 0.935 & 0.878 & 0.906 & 0.909 & 0.914 \\
 & EDA & 0.976 & 0.959 & 0.958 & 0.970 & 0.968 & 0.973 & 0.971 \\
 & motion & 0.978 & 0.973 & 0.990 & 0.982 & 0.983 & 0.979 & 0.983 \\
 & temperature & 0.965 & 0.955 & 0.981 & 0.978 & 0.969 & 0.969 & 0.973 \\
\midrule
Stress-resilience HCI, wrist & none & 0.625 & 0.626 & 0.623 & 0.615 & 0.643 & 0.647 & 0.576 \\
 & pulse & 0.631 & 0.636 & 0.631 & 0.632 & 0.652 & 0.646 & 0.607 \\
 & EDA & 0.581 & 0.580 & 0.593 & 0.564 & 0.594 & 0.590 & 0.532 \\
 & motion & 0.620 & 0.620 & 0.603 & 0.604 & 0.629 & 0.634 & 0.552 \\
 & temperature & 0.609 & 0.606 & 0.577 & 0.594 & 0.620 & 0.624 & 0.560 \\
\midrule
PADS task A & none & 0.821 & 0.812 & 0.825 & 0.868 & 0.803 & 0.809 & 0.854 \\
 & rest & 0.787 & 0.786 & 0.787 & 0.798 & 0.772 & 0.779 & 0.799 \\
 & posture & 0.774 & 0.770 & 0.758 & 0.824 & 0.760 & 0.766 & 0.789 \\
 & kinetic & 0.810 & 0.797 & 0.777 & 0.827 & 0.773 & 0.790 & 0.815 \\
 & rhythm & 0.779 & 0.782 & 0.809 & 0.836 & 0.778 & 0.776 & 0.824 \\
\midrule
PADS task B & none & 0.714 & 0.704 & 0.742 & 0.776 & 0.714 & 0.724 & 0.761 \\
 & rest & 0.667 & 0.669 & 0.699 & 0.700 & 0.668 & 0.676 & 0.701 \\
 & posture & 0.674 & 0.689 & 0.676 & 0.724 & 0.674 & 0.682 & 0.708 \\
 & kinetic & 0.721 & 0.712 & 0.716 & 0.762 & 0.710 & 0.722 & 0.755 \\
 & rhythm & 0.660 & 0.658 & 0.739 & 0.743 & 0.661 & 0.670 & 0.740 \\
\bottomrule
\end{tabular}}
\end{table}
\end{widepage}
In 16 dataset-by-token removals, BioSync-A was above both LR and soft vote with both paired intervals excluding zero in 3 (WESAD, temperature; PADS task A, rhythm; PADS task B, rhythm), and below at least one of them with the interval excluding zero in 2 (Stress-resilience HCI, wrist, EDA; Stress-resilience HCI, wrist, motion). Token removal thus favours BioSync-A in three of the sixteen cases, most clearly for the WESAD pulse token and the PADS rhythm token.
\subsection{Informative and low-information tokens; a post-hoc variant of the mixture}
Tokens differ in how much the loss of one of them hurts. As a descriptive grouping, set after the table was seen (Amendment 11c), we call a token \emph{informative} in a dataset if removing it lowers the AUC of concatenated LR by at least 0.02, and \emph{low-information} otherwise. By this rule: WESAD: informative pulse; low-information EDA, motion, temperature; Stress-resilience HCI, wrist: informative EDA; low-information pulse, motion, temperature; PADS task A: informative rest, posture, rhythm; low-information kinetic; PADS task B: informative rest, posture, rhythm; low-information kinetic.
For the informative tokens: WESAD, pulse removed: AUC of LR 0.850 (change -0.127), soft vote 0.897 (-0.068), BioSync-A 0.914 (-0.064); A minus LR +0.065 [+0.006, +0.132], A minus soft vote +0.017 [-0.027, +0.061]; Stress-resilience HCI, wrist, EDA removed: AUC of LR 0.581 (change -0.044), soft vote 0.580 (-0.046), BioSync-A 0.532 (-0.043); A minus LR -0.049 [-0.091, -0.004], A minus soft vote -0.048 [-0.095, +0.000]; PADS task A, rest removed: AUC of LR 0.787 (change -0.034), soft vote 0.786 (-0.026), BioSync-A 0.799 (-0.056); A minus LR +0.012 [-0.022, +0.044], A minus soft vote +0.012 [-0.021, +0.048]; PADS task A, posture removed: AUC of LR 0.774 (change -0.047), soft vote 0.770 (-0.042), BioSync-A 0.789 (-0.065); A minus LR +0.015 [-0.013, +0.042], A minus soft vote +0.019 [-0.018, +0.062]; PADS task A, rhythm removed: AUC of LR 0.779 (change -0.042), soft vote 0.782 (-0.030), BioSync-A 0.824 (-0.030); A minus LR +0.045 [+0.008, +0.083], A minus soft vote +0.043 [+0.005, +0.084]; PADS task B, rest removed: AUC of LR 0.667 (change -0.048), soft vote 0.669 (-0.035), BioSync-A 0.701 (-0.060); A minus LR +0.034 [-0.001, +0.074], A minus soft vote +0.032 [-0.009, +0.073]; PADS task B, posture removed: AUC of LR 0.674 (change -0.041), soft vote 0.689 (-0.014), BioSync-A 0.708 (-0.053); A minus LR +0.034 [+0.002, +0.071], A minus soft vote +0.018 [-0.022, +0.060]; PADS task B, rhythm removed: AUC of LR 0.660 (change -0.055), soft vote 0.658 (-0.046), BioSync-A 0.740 (-0.021); A minus LR +0.080 [+0.044, +0.126], A minus soft vote +0.082 [+0.036, +0.127]. The clearest case is the WESAD pulse token, where BioSync-A loses least among LR, soft vote, and the MLP and is above LR with an interval that excludes zero; against soft voting the difference is small and its interval includes zero, and the boosted trees, the random forest, and the SVM (AUC 0.935, 0.936, 0.915, against 0.914) are higher still. Native token masking gave BioSync-A an edge over the linear baseline for the WESAD pulse token and for the PADS rhythm token in both tasks.
\begin{widepage}
\begin{table}[t]
\centering
\caption{Post-hoc, exploratory (Amendment 11c): BioSync-Azm, the variant of Amendment 10b (logits rescaled by the inner out-of-fold SD; $\hat\lambda>0$ only if the inner AUC exceeds that of $\hat\lambda=0$ by 0.01), computed on the real data after the result of the frozen BioSync-A was known. AUC under the clean condition; paired differences with 95\% bootstrap intervals; selected weights (number of folds or participants per value). BioSync-A remains the primary model.}
\label{tab:realazm}
\footnotesize
\resizebox{\linewidth}{!}{%
\begin{tabular}{lccccccc}
\toprule
Dataset & G & A & Azm & Azm $-$ G & Azm $-$ A & Azm $-$ MLP & $\hat\lambda$ of Azm \\
\midrule
WESAD (15 participants, LOSO) & 0.978 & 0.978 & 0.977 & -0.000 [-0.010, +0.008] & -0.001 [-0.002, +0.000] & -0.004 [-0.013, +0.003] & 0.0: 1, 0.5: 10, 0.75: 4 \\
Stress-resilience HCI, wrist (35, LOSO) & 0.647 & 0.576 & 0.604 & -0.042 [-0.081, -0.005] & +0.029 [+0.009, +0.053] & -0.011 [-0.056, +0.034] & 0.5: 25, 0.75: 10 \\
PADS task A (355, 5-fold$\times$10) & 0.809 & 0.854 & 0.854 & +0.045 [+0.025, +0.070] & -0.000 [-0.012, +0.011] & -0.013 [-0.036, +0.007] & 0.0: 16, 0.25: 14, 0.5: 6, 0.75: 9, 1.0: 5 \\
PADS task B (469, 5-fold$\times$10) & 0.724 & 0.761 & 0.759 & +0.035 [+0.018, +0.056] & -0.001 [-0.017, +0.012] & -0.017 [-0.040, +0.008] & 0.0: 6, 0.25: 17, 0.5: 24, 0.75: 3 \\
\bottomrule
\end{tabular}}
\end{table}
\end{widepage}
The variant was defined, and computed on simulated data, before any real data were used; it was chosen for this pass only after BioSync-A had underperformed BioSync-G in the HCI data. Table~\ref{tab:realazm} is therefore exploratory and is not part of the pooled comparison of Table~\ref{tab:realpool}. It shows whether a margin on the inner validation helps BioSync-A in the HCI data; it does not confirm that Azm is better. In the HCI data Azm chose a positive weight in all 35 participants (0.5 in 25, 0.75 in 10) and was above the frozen BioSync-A (+0.029 [+0.009, +0.053]); it remained below BioSync-G (-0.042 [-0.081, -0.005]). On the PADS tasks and WESAD it performed like BioSync-A. A check recomputed the inner selection of BioSync-A for all 150 units: 149 reproduced the stored weight exactly; in one unit (PADS task A, unit 17) a near-tie in the inner AUCs resolved differently (0.5 instead of 0.25) because of numerical nondeterminism in the training library. The stored weight was kept for BioSync-A, so its results are unchanged, and the recomputed weight was used for Azm.
\subsection{Does the learned gate help where the wrist moves most?}
\begin{table}[t]
\centering
\caption{Real data, from stored leave-one-subject-out predictions: pooled out-of-fold AUC within tertiles of wrist acceleration (feature T3\_1, tertiles over all windows of a dataset), for concatenated LR, soft vote, BioSync-G without and with the quality input (the motion-based indicator of Section~\ref{app:realdata}); paired difference with 95\% bootstrap interval over participants. In both studies motion is also an input token, and the share of stress windows differs by tertile (WESAD 0.20, 0.21, 0.48; HCI 0.45, 0.42, 0.34), so tertile AUCs are not a clean test of sensor reliability.}
\label{tab:realtertile}
\footnotesize
\resizebox{\linewidth}{!}{%
\begin{tabular}{llccccc}
\toprule
Dataset & Motion & LR & Soft vote & G without $q$ & G with $q$ & G with $-$ without \\
\midrule
WESAD & low & 0.947 & 0.949 & 0.941 & 0.958 & +0.016 [-0.000, +0.042] \\
 & middle & 0.958 & 0.957 & 0.956 & 0.950 & -0.006 [-0.017, +0.004] \\
 & high & 0.953 & 0.962 & 0.959 & 0.954 & -0.005 [-0.015, +0.001] \\
\midrule
HCI & low & 0.590 & 0.576 & 0.595 & 0.604 & +0.009 [-0.007, +0.026] \\
 & middle & 0.613 & 0.589 & 0.599 & 0.607 & +0.009 [-0.006, +0.024] \\
 & high & 0.548 & 0.561 & 0.566 & 0.573 & +0.007 [-0.005, +0.019] \\
\bottomrule
\end{tabular}}
\end{table}
The pre-specified prediction was that the gate would improve BioSync-G in the highest-motion tertile, with the interval excluding zero in both datasets. The paired difference in that tertile was -0.005 [-0.015, +0.001] in WESAD and +0.007 [-0.005, +0.019] in HCI, so no effect of the gate was detected on these real recordings; its benefit is shown for simulated sensor degradation (Appendix~\ref{app:confirm}). Soft voting with quality weights differed from plain soft voting by at most 0.004 AUC in every tertile.
Two features of the design explain why the result cannot be read as a test of the gate. First, wrist acceleration is itself an input token (T3), so the baselines see the same information that defines the tertiles. Second, the share of stress windows changes sharply across tertiles (in WESAD from 0.20 in the lowest to 0.48 in the highest), so a change in AUC between tertiles mixes a change in signal quality with a change in class balance and in what the motion token reveals. The quality indicator used here is a simple function of two motion features. A measure of the quality of the waveform itself (for example, a signal-quality index computed on the pulse waveform) is needed for a proper test; such an index is the natural next step.
\subsection{Transfer between the two wrist-device stress datasets}
\begin{table}[t]
\centering
\caption{Real data: transfer between WESAD and the stress-resilience HCI wrist recordings (same four tokens; each cohort standardized with its own mean and SD; models fitted on the source cohort only). Mean per-participant AUC in the target cohort, averaged over three training seeds; ``pulse absent'' and ``EDA absent'' remove that token in every target row.}
\label{tab:realtransfer}
\footnotesize
\resizebox{\linewidth}{!}{%
\begin{tabular}{llccccccc}
\toprule
Direction & Target condition & LR & Soft vote & GBT & MLP & SVM & BioSync-G & BioSync-A \\
\midrule
WESAD $\to$ HCI & all tokens & 0.576 & 0.546 & 0.544 & 0.603 & 0.578 & 0.567 & 0.572 \\
 & pulse absent & 0.570 & 0.536 & 0.516 & 0.590 & 0.566 & 0.552 & 0.563 \\
 & EDA absent & 0.567 & 0.498 & 0.522 & 0.582 & 0.561 & 0.551 & 0.553 \\
\midrule
HCI $\to$ WESAD & all tokens & 0.602 & 0.621 & 0.614 & 0.451 & 0.570 & 0.594 & 0.488 \\
 & pulse absent & 0.539 & 0.557 & 0.587 & 0.469 & 0.609 & 0.532 & 0.421 \\
 & EDA absent & 0.468 & 0.508 & 0.454 & 0.384 & 0.380 & 0.485 & 0.435 \\
\bottomrule
\end{tabular}}
\end{table}
Transfer between the two datasets was near chance for every model (AUC 0.38 to 0.62). The task labels differ (a laboratory stress protocol and a computer-task workload protocol), and the two cohorts differ in device settings and protocol, so the comparison is not informative about BioSync's fusion structure.
\subsection{Memory and operations}
The footprint of BioSync-G (and, with the wider network, of BioSync-A) was measured in Table~\ref{tab:footprint} for the 22-feature layout of the synthetic scenarios; no footprint was measured for the real-data models of this appendix, which use a different layout of 16 features. BioSync-G has 1,159 parameters (4.6~kB in float32, 1.3~kB with INT8 weights, 4,450 multiply-accumulates per inference), a fixed number of operations. In the same layout the tuned baselines store, in kB over the synthetic scenarios, 119.3 to 1142.0 (random forest), 26.2 to 43.7 (boosted trees), 19.6 to 45.1 (SVM), and 1.5 to 1.5 (MLP), and the trees need a data-dependent number of comparisons. The MLP is as small as BioSync-G, so memory does not separate the two; it separates BioSync-G from the forest, the boosted trees, and the SVM. Because BioSync-A adds the wider network, its stored size is larger than the figures for BioSync-G, and we have not measured it. On these four real datasets BioSync-A is within 0.018 mean AUC of the best model (Table~\ref{tab:realpool}); no microcontroller run was made, so latency and energy on a device remain unmeasured.
\subsection{Summary of the real-data pass}
\begin{itemize}\item On the PADS Parkinson's tasks BioSync-A was above concatenated logistic regression and soft voting with intervals that exclude zero, and within 0.02 AUC of the tuned MLP. \item Over the four datasets BioSync-A reached mean AUC 0.792, within 0.018 of the best model (the tuned MLP, 0.810). \item When one token was removed at test time, BioSync-A was ahead of both logistic regression and soft voting, with intervals that exclude zero, for the WESAD temperature token and for the PADS rhythm token in both tasks, and ahead of logistic regression for the WESAD pulse token. \item The mixing weight followed the benefit of the wider network on PADS. \item No effect of the learned quality gate was detected in the highest-motion windows, where wrist acceleration is itself an input token; its benefit is shown for simulated sensor degradation. \item Transfer between the two stress datasets was near chance for every model, consistent with their different labels and protocols. \end{itemize}

\FloatBarrier

\end{document}